\documentclass{article} 
\usepackage{colm2024_conference}

\usepackage{booktabs}
\usepackage{graphicx}
\usepackage{enumitem}
\usepackage{wrapfig}
\usepackage{algorithm}
\usepackage{algpseudocode}
\usepackage{natbib}
\usepackage{makecell}
\usepackage{booktabs}
\usepackage{bbm}
\usepackage{array}
\usepackage{amsmath} 
\usepackage{amssymb}
\usepackage{amsfonts}
\usepackage{multirow}
\usepackage{verbatim}
\usepackage{caption}
\usepackage{longtable}
\usepackage{supertabular}
\usepackage{hyperref}
\usepackage{CJKutf8}
\usepackage[utf8]{inputenc} 
\usepackage[T1]{fontenc} 
\usepackage[french,vietnamese,mongolian,greek,english]{babel}
\usepackage{pifont}
\usepackage{afterpage}
\usepackage{enumitem}
\usepackage{tablefootnote}
\usepackage{xspace}
\usepackage{textcomp}
\usepackage{makecell}
\usepackage{lscape} 
\usepackage{siunitx}
\usepackage{listings}
\usepackage{xcolor}
\usepackage{adjustbox}
\definecolor{dt}{gray}{0.7}
\definecolor{tongyi-purple}{RGB}{97,92,237}
\colorlet{tongyi-purple-alpha}{tongyi-purple!38}

\newcommand{\qwenvl}{{Qwen3-VL}\xspace}
\newcommand{\InstructBig}{{Qwen3-VL-235B-A22B-Instruct}\xspace}
\newcommand{\ThinkBig}{{Qwen3-VL-235B-A22B-Thinking}\xspace}
\newcommand{\ModelBig}{{Qwen3-VL-235B-A22B}\xspace}

\usepackage{pifont}       
\usepackage{bbding}       
\usepackage{fontawesome}

\usepackage{scrextend}

\usepackage{tgpagella}
\usepackage{latexsym}
\usepackage[T1]{fontenc}
\usepackage[utf8]{inputenc}
\usepackage{microtype}
\definecolor{mydarkblue}{rgb}{0,0.08,0.45}
\definecolor{citecolor}{HTML}{0071BC}
\usepackage{url}            
\usepackage{nicefrac}       
\usepackage{changepage}
\usepackage{xargs}          
\usepackage{wrapfig,lipsum,booktabs}
\usepackage{longtable}
\usepackage{subcaption}
\usepackage{endnotes}

\usepackage{pgfplots}
\usetikzlibrary{pgfplots.groupplots}
\pgfplotsset{compat=1.3}
\usepackage{tikz}
\usetikzlibrary{patterns}

\usepackage[most]{tcolorbox}
\usepackage{fvextra}
\usepackage{graphicx}
\usepackage[capitalize,noabbrev]{cleveref}
\crefname{section}{Section}{\S\S}
\Crefname{section}{Section}{\S\S}
\crefname{table}{Table}{Tables}
\crefname{figure}{Figure}{Figures}
\crefname{algorithm}{Algorithm}{}
\crefname{equation}{eq.}{}
\crefname{appendix}{Appendix}{}
\crefformat{section}{Section #2#1#3}
\usepackage{multicol}
\usepackage{fancyvrb} 
\usepackage{tcolorbox}
\newsavebox{\myverbcontent}
\usepackage{titlesec}
\titleformat*{\section}{\large\bfseries}

\usepackage{nicematrix} 
\usepackage{arydshln}

\makeatletter
\DeclareRobustCommand\onedot{\futurelet\@let@token\@onedot}
\def\@onedot{\ifx\@let@token.\else.\null\fi\xspace}

\title{Qwen3-VL Technical Report}

\author{
\bf Qwen Team}

\begin{document}

\maketitle

\begin{abstract} 
We introduce Qwen3-VL, the most capable vision–language model in the Qwen series to date, achieving superior performance across a broad range of multimodal benchmarks. It natively supports interleaved contexts of up to 256K tokens, seamlessly integrating text, images, and video. The model family includes both dense (2B/4B/8B/32B) and mixture-of-experts (30B-A3B/235B-A22B) variants to accommodate diverse latency–quality trade-offs.
Qwen3-VL delivers three core pillars:
(i) markedly stronger pure-text understanding, surpassing comparable text-only backbones in several cases;
(ii) robust long-context comprehension with a native 256K-token window for both text and interleaved multimodal inputs, enabling faithful retention, retrieval, and cross-referencing across long documents and videos; and
(iii) advanced multimodal reasoning across single-image, multi-image, and video tasks, demonstrating leading performance on comprehensive evaluations such as MMMU and visual-math benchmarks (e.g., MathVista and MathVision).
Architecturally, we introduce three key upgrades:
(i) an enhanced interleaved-MRoPE for stronger spatial–temporal modeling across images and video;
(ii) DeepStack integration, which effectively leverages multi-level ViT features to tighten vision–language alignment; and
(iii) text-based time alignment for video, evolving from T-RoPE to explicit textual timestamp alignment for more precise temporal grounding.
To balance text-only and multimodal learning objectives, we apply square-root reweighting, which boosts multimodal performance without compromising text capabilities. We extend pretraining to a context length of 256K tokens and bifurcate post-training into non-thinking and thinking variants to address distinct application requirements.
Furthermore, we allocate additional compute resources to the post-training phase to further enhance model performance.
Under comparable token budgets and latency constraints, Qwen3-VL achieves superior performance in both dense and Mixture-of-Experts (MoE) architectures. We envision Qwen3-VL serving as a foundational engine for image-grounded reasoning, agentic decision-making, and multimodal code intelligence in real-world workflows.
\end{abstract}
\begin{figure*}[ht]
\centering
\includegraphics[width= 1\linewidth]{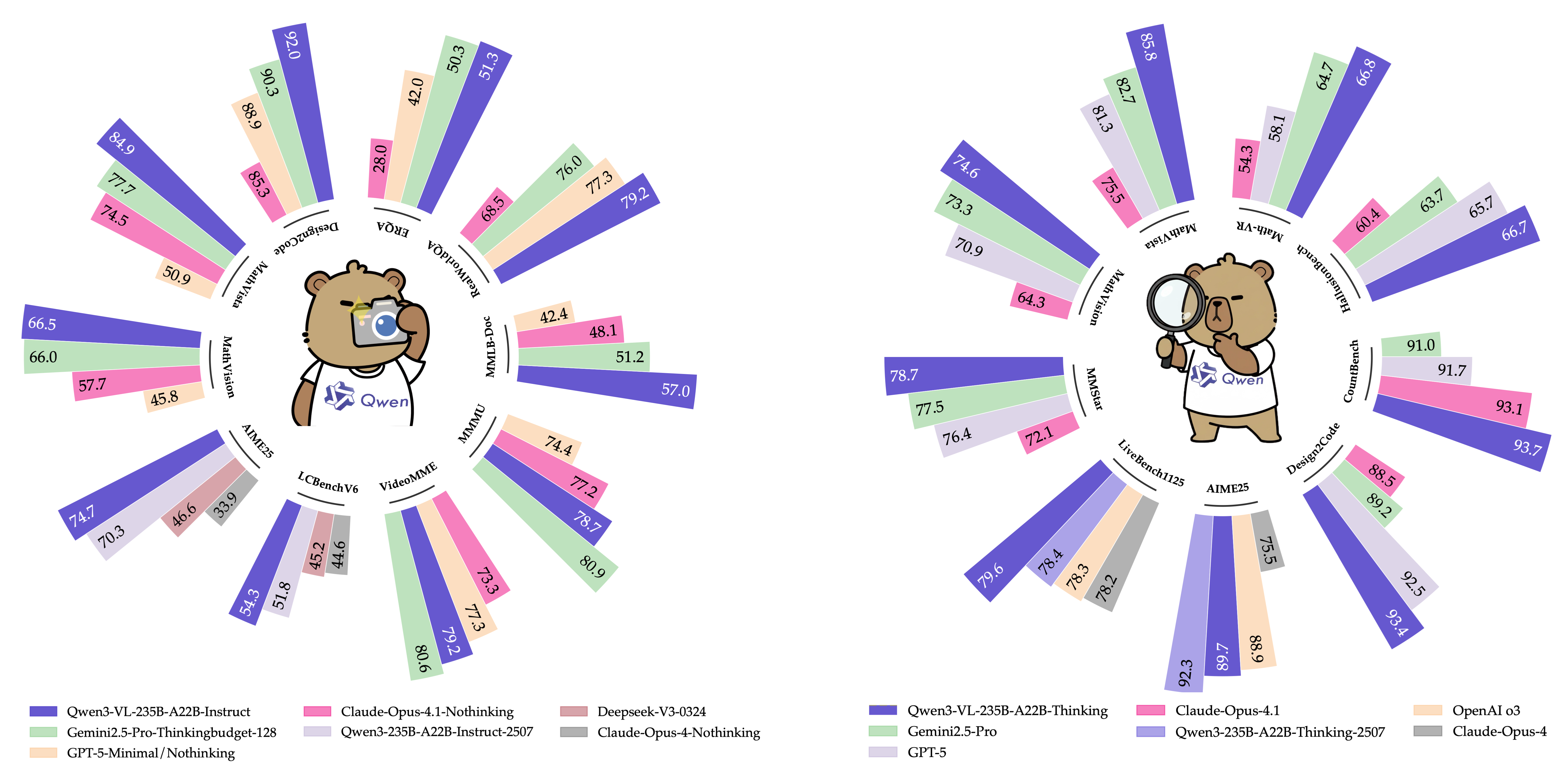}
\end{figure*}

\newpage
\section{Introduction}
Vision–language models (VLMs) have achieved substantive progress in recent years, evolving from foundational visual perception to advanced multimodal reasoning across images and video. The rapid advancement of VLMs has given rise to a rapidly expanding landscape of downstream applications—such as long-context understanding, STEM reasoning, GUI comprehension and interaction, and agentic workflows. Crucially, these advances must not erode the underlying large language model’s (LLM’s) linguistic proficiency; multimodal models are expected to match or surpass their text-only counterparts on language benchmarks.

In this report, we present Qwen3-VL and its advances in both general-purpose and advanced applications. Built on the Qwen3 series~\citep{qwen3}, we instantiate four dense models (2B/4B/8B/32B) and two mixture-of-experts (MoE) models (30B-A3B / 235B-A22B), each trained with a context window of up to 256K tokens to enable long-context understanding. By optimizing the training corpus and training strategy, we preserve the underlying LLM’s language proficiency during vision–language (VL) training, thereby substantially improving overall capability. We release both non-thinking and thinking variants; the latter demonstrates significantly stronger multimodal reasoning capabilities, achieving superior performance on complex reasoning tasks.

We first introduce the architectural improvements, which span three components:
1) Enhanced positional encoding. In Qwen2.5-VL, we used MRoPE as a unified positional encoding scheme for text and vision. We observed that chunking the embedding dimensions into temporal (t), horizontal (h), and vertical (w) groups induces an imbalanced frequency spectrum and hampers long-video understanding. We therefore adopt an interleaved MRoPE that distributes t, h, and w uniformly across low- and high-frequency bands, yielding more faithful positional representations.
2) DeepStack for cross-layer fusion. To strengthen vision–language alignment, we incorporate the pioneering DeepStack~\citep{meng2024deepstack} mechanism. Visual tokens from different layers of the vision encoder are routed to corresponding LLM layers via lightweight residual connections, enhancing multi-level fusion without introducing extra context length.
3) Explicit video timestamps. We replace the absolute-time alignment via positional encoding used in Qwen2.5-VL with explicit timestamp tokens to mark frame groups, providing a simpler and more direct temporal representation.
In addition, on the optimization side, we move from a per-sample loss to a square-root-normalized per-token loss, which better balances the contributions of text and multimodal data during training.

To build a more capable and robust vision–language foundation model, we overhauled our training data in terms of quality, diversity, and structure. Key upgrades include enhanced caption supervision, expanded omni-recognition and OCR coverage, normalized grounding with 3D/spatial reasoning, and new corpora for code, long documents, and temporally grounded video. We further infused chain-of-thought reasoning and high-quality, diverse GUI-agent interaction data to bridge perception, reasoning, and action. Together, these innovations enable stronger multimodal understanding, precise grounding, and tool-augmented intelligence.

Our training pipeline consists of two stages: pretraining and post-training.
Pretraining proceeds in four phases: a warm-up alignment phase that updates only the merger (vision–language projection) layers while keeping the rest of the model frozen, followed by full-parameter training with progressively larger context windows at 8K, 32K, and 256K sequence lengths.
Post-training comprises three phases:
(i) supervised fine-tuning on long chain-of-thought data,
(ii) knowledge distillation from stronger teacher models, and
(iii) reinforcement learning.

The above innovations equip Qwen3-VL with strong capabilities not only as a robust vision–language foundation model but also as a flexible platform for real-world multimodal intelligence—seamlessly integrating perception, reasoning, and action across diverse application domains. In the following sections, we present the model architecture, training framework, and extensive evaluations that demonstrate its consistent and competitive performance on text, vision, and multimodal reasoning benchmarks.

\begin{figure*}[t]
\centering
\includegraphics[width= 1\linewidth]{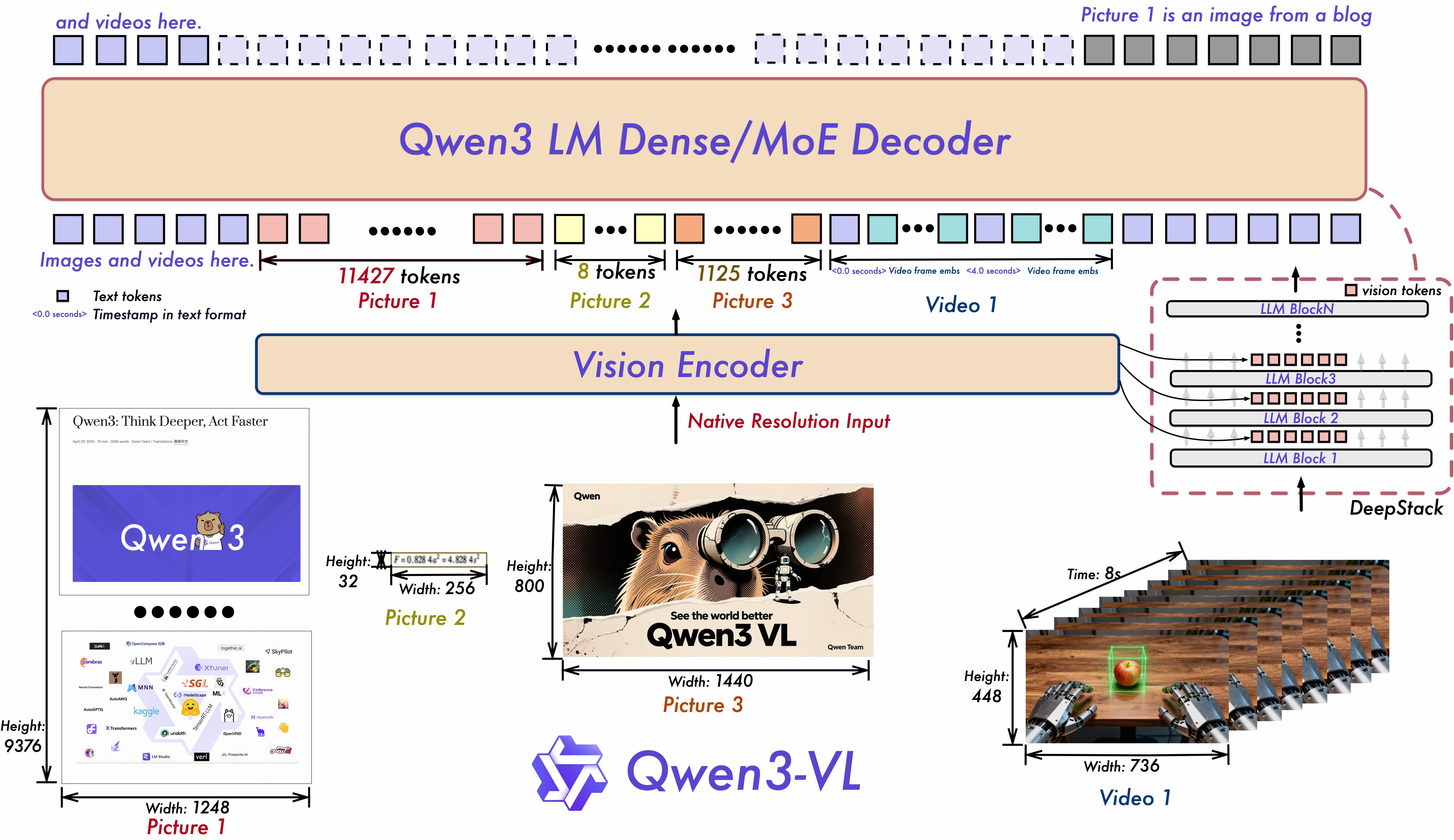}
   \caption{
   The Qwen3-VL framework integrates a vision encoder and a language model decoder to process multimodal inputs, including text, images, and video. The vision encoder is specifically designed to handle dynamic, native-resolution visual inputs, mapping them to visual tokens of variable length. To enhance perceptual capability and preserve rich visual information, we incorporate the pioneering DeepStack mechanism, which injects visual tokens from multiple layers of the vision encoder into corresponding layers of the LLM. Furthermore, we adopt Interleaved MRoPE to encode positional information for multimodal inputs with a balanced frequency spectrum, and introduce text-based timestamp tokens to more effectively capture the temporal structure of video sequences.
}
\label{fig:arc}
\end{figure*}

\section{Model Architecture}
Following Qwen2.5-VL~\citep{qwen2.5vl}, Qwen3-VL adopts a three-module architecture comprising a vision encoder, an MLP-based vision–language merger, and a large language model (LLM). Figure~\ref{fig:arc} depicts the detailed model structure.

\textbf{Large Language Model}: 
Qwen3-VL is instantiated in three dense variants (Qwen3-VL-2B/4B/8B/32B) and two MoE variants (Qwen3-VL-30B-A3B, Qwen3-VL-235B-A22B), all built upon Qwen3 backbones. The flagship model, Qwen3-VL-235B-A22B, has 235B total parameters with 22B activated per token. It outperforms most VLMs across a broad set of multimodal tasks and surpasses its text-only counterpart on the majority of language benchmarks.

\textbf{Vision Encoder}: 
We utilize the SigLIP-2 architecture~\citep{tschannen2025siglip} as our vision encoder and continue training it with dynamic input resolutions, initialized from official pretrained checkpoints. To accommodate dynamic resolutions effectively, we employ 2D-RoPE and interpolate absolute position embeddings based on input size, following the methodology of CoMP~\citep{chen2025comp}.
Specifically, we default to the SigLIP2-SO-400M variant and use SigLIP2-Large (300M) for small-scale LLMs (2B and 4B).

\textbf{MLP-based Vision-Language Merger}: 
As in Qwen2.5-VL, we use a two-layer MLP to compress $2\times2$ visual features from the vision encoder into a single visual token, aligned with the LLM’s hidden dimension. Additionally, we deploy specialized mergers to support the DeepStack mechanism~\citep{meng2024deepstack}, the details of which are fully described in~\cref{sec:deepstack}.

\subsection{Interleaved MRoPE}
Qwen2-VL~\citep{Qwen2-VL} introduced MRoPE to model positional information for multimodal inputs. In its original formulation, the embedding dimensions are partitioned into temporal (t), horizontal (h), and vertical (w) subspaces, each assigned distinct rotary frequencies. This results in an imbalanced frequency spectrum, which subsequent studies have shown to degrade performance on long-video understanding benchmarks.
To address this, we redesign the frequency allocation by interleaving the t, h, and w components across the embedding dimensions~\citep{interleave-mrope}. This ensures that each spatial–temporal axis is uniformly represented across both low- and high-frequency bands. The resulting balanced spectrum mitigates the original spectral bias and significantly improves long-range positional modeling for video.

\subsection{DeepStack}
\label{sec:deepstack}
We draw inspiration from DeepStack~\citep{meng2024deepstack} and inject visual tokens into multiple layers of the LLM. Unlike the original DeepStack approach, which stacks tokens from multi-scale visual inputs, we extend DeepStack to extract visual tokens from intermediate layers of the Vision Transformer (ViT). This design preserves rich visual information, ranging from low- to high-level representations.

Specifically, as illustrated in~\cref{fig:arc}, we select features from three distinct levels of the vision encoder. Subsequently, dedicated vision–language merger modules project these multi-level features into visual tokens, which are then added directly to the corresponding hidden states of the first three LLM layers.

\subsection{Video Timestamp}
In Qwen2.5-VL, a time-synchronized variant of MRoPE is employed to endow the model with temporal awareness. However, we identify two key limitations of this approach:
(1) By tying temporal position IDs directly to absolute time, the method produces excessively large and sparse temporal position ids for long videos, degrading the model’s ability to understand long temporal contexts.
(2) Effective learning under this scheme requires extensive and uniformly distributed sampling across various frame rates (fps), significantly increasing the cost of training data construction.

To address these issues, we adopt a textual token–based time encoding strategy~\citep{chen2024timemarker}, wherein each video temporal patch is prefixed with a timestamp expressed as a formatted text string—e.g., \texttt{<3.0 seconds>}.
Furthermore, during training, we generate timestamps in both seconds and HMS (hours:minutes:seconds) formats to ensure the model learns to interpret diverse timecode representations. Although this approach incurs a modest increase in context length, it enables the model to perceive temporal information more effectively and precisely, thereby facilitating time-aware video tasks such as video grounding and dense captioning.

\section{Pre-Training}

\subsection{Training Recipe}
We first enhance the vision encoder by conducting continuous training with dynamic resolutions based on the pre-trained SigLIP-2 model. The overall Qwen3-VL model adopts a three-module architecture, comprising this vision encoder, an MLP-based vision–language merger, and a Qwen3 large language model (LLM) backbone. Building on this architecture, our pre-training methodology is systematically structured into four distinct stages, designed to progressively build capabilities from basic alignment to long-context understanding. An overview of these stages is presented in Table~\ref{tab:training_setup}.
\begin{table}[htbp]
\centering
\caption{Training setup and hyperparameters across different stages for Qwen3-VL.}
\label{tab:training_setup}
\begin{tabular}{@{}cllccc@{}}
\toprule
\textbf{Stage} & \textbf{Objective} & \textbf{Training} & \textbf{Token Budget} & \textbf{Sequence Length} \\ \midrule
S0 & Vision-Language Alignment & Merger & 67B & 8,192 \\
S1 & Multimodal Pre-Training & All & \textasciitilde 1T & 8,192 \\
S2 & Long-Context Pre-Training & All & \textasciitilde 1T & 32,768 \\
S3 & Ultra-Long-Context Adaptation & All & 100B & 262,144 \\ \bottomrule
\end{tabular}
\end{table}

\textbf{Stage 0: Vision-Language Alignment.} The initial stage (S0) focuses on efficiently bridging the modality gap between the vision encoder and the LLM. Crucially, only the parameters of the MLP merger are trained during this phase, while both the vision encoder and the LLM backbone remain frozen. We utilize a curated dataset of approximately 67B tokens, consisting of high-quality image-caption pairs, visual knowledge collections, and optical character recognition (OCR) data. All training is conducted with a sequence length of 8,192. This alignment-first approach establishes a solid foundation for cross-modal understanding before proceeding to full-parameter training.

\textbf{Stage 1: Multimodal Pre-Training.} Following the initial alignment, Stage 1 (S1) transitions to full-parameter Multimodal Pre-Training. In this phase, we unfreeze all model components—the vision encoder, the merger, and the LLM—for joint end-to-end training. The model is trained on a massive and diverse dataset of approximately 1 trillion (1T) tokens. To maintain the LLM's strong language abilities, the data mixture is composed of vision-language (VL) data and text-only data. The VL portion is rich and varied, adding interleaved image-text documents, visual grounding tasks, visual question answering (VQA), data from STEM domains, and a small amount of video data to introduce temporal understanding. The sequence length remains at 8,192.

\textbf{Stage 2: Long-Context Pre-Training.} Stage 2 (S2) aims to significantly extend the model's contextual processing abilities. A key change in this stage is the quadrupling of the sequence length to 32,768, while all model parameters continue to be trainable. Training is conducted on a dataset of approximately 1T tokens, with an adjusted data mixture to support long-context tasks. The proportion of text-only data is increased to bolster long-form text comprehension, while the remaining VL data incorporates a significantly larger volume of video and agent-oriented instruction-following data. This stage is critical for enabling the model to process and reason over longer videos and complex, multi-step tasks.

\textbf{Stage 3: Ultra-Long-Context Adaptation.} The final stage (S3) is a specialized phase designed to push the model's context window to its operational limits. Here, we dramatically increase the sequence length to 262,144. The model is trained on a more focused 100B token dataset specifically curated for this purpose. The data is also composed of text-only data and VL data, with a strong emphasis on long-video and long-document understanding tasks. This final adaptation solidifies Qwen3-VL's proficiency in processing and analyzing extremely long sequential inputs, a key capability for applications like comprehensive document analysis and lengthy video summarization.

\subsection{Pre-Training Data}
\subsubsection{Image Caption and Interleaved Text-Image Data}
To build a robust foundation model for general-purpose vision–language understanding, we significantly expand and refine two core data modalities: image–caption pairs and interleaved text–image sequences. Our strategy emphasizes high-quality, diverse, and semantically rich multimodal grounding, supported by purpose-built models and rigorous filtering pipelines.

\textbf{Image Caption Data}:
We curate a large-scale corpus of contemporary, predominantly Chinese–English multilingual image–text pairs from web sources and apply a multi-stage refinement pipeline centered on a specialized Qwen2.5-VL-32B model fine-tuned for recaptioning. This model leverages the original raw text associated with each image to generate more comprehensive, fluent, and fine-grained captions—enriching descriptions of visual elements (e.g., object attributes, spatial layouts, and contextual semantics) while simultaneously improving the linguistic quality and informativeness of the textual component.

Deduplication is performed exclusively on the recaptioned text using semantic similarity metrics, ensuring removal of redundant samples without sacrificing visual diversity. To further enhance coverage of underrepresented concepts, we apply clustering~\citep{johnson2019billion,douze2024faiss,diao2025climb} over visual embeddings to identify sparse regions in the data distribution and perform targeted augmentation. The result is a high-fidelity caption dataset that balances scale, diversity, and descriptive granularity.

\textbf{Interleaved Text-Image Data}:
We collect diverse real-world multimodal documents sourced from recent Chinese and English websites~\citep{laurenccon2023obelics,zhu2023multimodal,li2024omnicorpus}. All documents undergo domain classification~\citep{wettig2025organize} using a lightweight Qwen-based scorer fine-tuned for fine-grained domain identification. Based on validation experiments across domains, we systematically exclude harmful or low-value categories—such as advertisements, promotional content, and clickbait—using the same efficient scorer to filter out undesirable samples.

For book-scale interleaved data, we employ a fine-tuned Qwen2.5-VL-7B model to perform high-accuracy multimodal parsing, precisely extracting and aligning text with embedded figures, diagrams, and photographs. To enable ultra-long context modeling, we construct a specialized subset by merging consecutive pages into sequences of up to 256K tokens, preserving natural page order and multimodal coherence. During preprocessing, we enforce strict quality controls:
(i) pure-text or low-alignment segments are removed;
(ii) for ultra-long book sequences, we require a minimum page count and a minimum image-to-text ratio to ensure meaningful visual–textual interaction throughout the context.
This yields a clean, diverse, and layout-aware interleaved corpus optimized for both grounded understanding and long-range multimodal reasoning.

\subsubsection{Knowledge}
World knowledge is essential for multimodal large language models (MLLMs) to achieve robust visual understanding, grounded reasoning, and entity-aware generation across diverse downstream tasks. To equip Qwen3-VL with a comprehensive grasp of both real-world and fictional concepts, we construct a large-scale pretraining dataset centered on well-defined entities spanning more than a dozen semantic categories—including animals, plants, landmarks, food, and everyday objects such as vehicles, electronics, and clothing.

Real-world entities follow a long-tailed distribution: prominent concepts appear frequently with high-quality annotations, while the majority are rare. To address this imbalance, we adopt an importance-based sampling strategy. High-prominence entities are sampled more heavily to ensure a sufficient learning signal, while low-prominence entities are included in smaller proportions to maintain broad coverage without overwhelming the training process. This approach effectively balances data quality, utility, and diversity.

All retained samples undergo a multi-stage refinement pipeline. In addition to standard filtering for noise and misalignment, we replace original or sparse captions—such as generic alt-text—with richer, LLM-generated descriptions. These enhanced captions not only identify the main entity but also describe its visual attributes, surrounding context, spatial layout, and interactions with other objects or people, thereby providing a more complete and grounded textual representation.

Together, these efforts yield a knowledge-rich, context-aware, and discrimination-focused training signal that significantly enhances Qwen3-VL’s ability to recognize, reason about, and accurately describe visual concepts in real-world scenarios.

\subsubsection{OCR, Document Parsing and Long Document Understanding}

\textbf{OCR:} 
To enhance OCR performance on real-world images, we curate a dataset of 30 million in-house collected samples using a coarse-to-fine pipeline. This pipeline refines OCR annotations by integrating pseudo-labels from OCR-specialized models with refinements from Qwen2.5-VL—without any human annotation.
Expanding beyond the 10 languages supported by Qwen2.5-VL (excluding Chinese and English), we incorporate an additional 29 languages, synthesizing approximately 30 million high-quality multilingual OCR samples and curating over 1 million internal real-world multilingual images.

 \textbf{Document Parsing:}
 For document parsing, we collect 3 million PDFs from Common Crawl, evenly distributed across 10 document types (300K samples each), along with 4 million internal documents. An in-house layout model first predicts the reading order and bounding boxes for textual and non-textual regions; Qwen2.5-VL-72B then performs region-specific recognition. The outputs are reassembled into position-aware, layout-aligned parsing data.

To ensure robust parsing across heterogeneous formats, we design a unified annotation framework supporting two representations:
\begin{itemize}[itemindent=0pt, labelsep=4pt, leftmargin=*]
    \item QwenVL-HTML, which includes fine-grained, element-level bounding boxes;
    \item QwenVL-Markdown, where only images and tables are localized, with tables encoded in LaTeX.
\end{itemize}
We construct a large-scale synthetic HTML corpus with precise annotations and systematically convert it to Markdown format. To further improve model generalization, we generate pseudo-labels on extensive collections of real documents and filter them for quality. The final training set combines synthetic and high-quality pseudo-labeled data to enhance both scalability and robustness.

\textbf{Long Document Understanding:} 
To enhance the model’s ability to understand multi-page PDFs—often spanning dozens of pages—we leverage a large-scale corpus of long-document data.
First, we synthesize long-document parsing sequences by merging single-page document samples. In each sequence, multiple page images are placed at the beginning, followed by their corresponding text derived from OCR or HTML parsing.
Second, we construct long-document visual question answering (VQA) data. Specifically, we sample high-quality multi-page PDFs and generate a diverse set of VQA examples that require the model to reason across multiple pages and heterogeneous document elements—such as charts, tables, figures, and body text.
We carefully balance the distribution of question types and ensure that supporting evidence draws from a wide range of modalities and layout components, thereby promoting robust, grounded, and multi-hop reasoning over extended contexts.

\subsubsection{Grounding and Counting}
Visual grounding is a fundamental capability for multimodal models, enabling them to accurately identify, interpret, and localize a wide spectrum of visual targets from specific objects to arbitrary image regions. In Qwen3-VL, we systematically enhance grounding proficiency and support two grounding modalities: bounding boxes and points. These representations allow for precise and flexible interpretation of image content across diverse scenarios and downstream tasks. In addition, we extend the grounding capacity of the model to support counting, enabling quantitative reasoning about visual entities. In the following, we briefly describe the data construction pipelines for grounding and counting.

\textbf{Box-based Grounding}:
We begin by aggregating widely used open-source datasets, including COCO~\citep{lin2014microsoft}, Objects365~\citep{shao2019objects365}, OpenImages~\citep{kuznetsova2020open}, and RefCOCO/+/g ~\citep{refcoco, refcocog}. To further enrich data diversity, we developed an automated synthesis pipeline that generates high-quality object annotations across a broad range of scenarios. This pipeline operates in three stages:
(i) object candidates are extracted from unlabeled images using Qwen2.5-VL;
(ii) these candidates are localized and annotated using both open-vocabulary detectors (specifically, Grounding DINO~\citep{grounding_dino}) and Qwen2.5-VL; and
(iii) the resulting annotations undergo quality assessment, with low-confidence or inaccurate ones systematically filtered out.
Through this approach, we constructed a large-scale, highly diverse box-based grounding dataset spanning a wide variety of visual contexts and object categories.

\textbf{Point-based Grounding}: 
To ensure robust point-based grounding, we curated a comprehensive dataset combining publicly available and synthetically generated pointing annotations. It integrates three sources:
(i) public pointing and counting annotations from PixMo~\citep{deitke2024molmo};
(ii) object grounding data derived from public object detection and instance segmentation benchmarks; and
(iii) high-precision pointing annotations generated by a dedicated synthesis pipeline designed to target fine-grained image details.

\textbf{Counting}: Building upon the grounding data, we curated a high-quality subset to form the basis of our counting dataset, which includes three distinct task formulations: direct counting, box-based counting, and point-based counting. Collectively, these three task types constitute a comprehensive counting dataset.

Different from Qwen2.5-VL, we adopt a normalized coordinate system scaled to the range $[0,1000]$ in this version. This design improves robustness to variations in image resolution and aspect ratio across diverse inputs, while also simplifying post-processing and enhancing the usability of predicted coordinates in downstream applications.

\subsubsection{Spatial Understanding and 3D Recognition}
To facilitate sophisticated interaction with the physical world, Qwen3-VL is designed with a deep understanding of spatial context. This enables the model to interpret spatial relationships, infer object affordances, and perform action planning and embodied reasoning. It can also estimate the 3D spatial positions of objects from a single monocular image. To support these capabilities, we created two comprehensive datasets focused on Spatial Understanding and 3D Grounding.

\textbf{Spatial Understanding}. 
Beyond localizing objects, Qwen3-VL is trained to reason about spatial relationships, object affordances, and feasible actions in 2D scenes—capabilities essential for embodied AI and interactive applications. To this end, we construct a specialized dataset that goes beyond standard grounding by incorporating:
(i) relational annotations (e.g., “the cup to the left of the laptop”),
(ii) affordance labels (e.g., “graspable”, “pressable”, “sittable”), and
(iii) action-conditioned queries that require planning (e.g., “What should I move first to reach the book behind the monitor?”).
These samples are derived from both curated real-world scenes and synthetically generated layouts, with natural language queries automatically generated via templated and LLM-based methods to ensure diversity and complexity. Critically, all spatial references are expressed relative to other objects or scene frames, rather than absolute coordinates, encouraging robust relational reasoning.
This training enables Qwen3-VL to not only answer “where” questions but also “how” and “what can be done” — forming a foundation for agentic interaction with visual environments.

\textbf{3D Grounding}.
To further enhance the model’s ability to understand the physical world from images, we constructed a specialized pretraining dataset for 3D visual grounding. We sourced data from public collections of diverse indoor and outdoor scenes and reformulated it into a visual question-answering format. Each sample consists of:
1) a single-view camera image,
2) a natural language referring expression, and
3) the corresponding 9-DoF 3D bounding box annotations in a structured JSON format, specifying the object’s spatial position and semantic label.
As the 3D bounding boxes are derived from multiple sensors and data sources, they exhibit varying camera intrinsic parameters and inherent noise. To this end, we filter out heavily occluded and inaccurate labels and follow Omni3D~\citep{brazil2023omni3d} to unify all data into a virtual camera coordinate system.
We also synthesized a large corpus of descriptive captions to create rich textual queries for 3D grounding. These descriptions go beyond naming the object’s category to include detailed attributes, layout arrangements, spatial location, visual affordances, and interactions with surrounding objects—yielding more fine-grained and grounded referring expressions.

\subsubsection{Code}
We enhance the Qwen3-VL series with dedicated coding capabilities by incorporating two categories of code-related data into the training corpus, enabling the model to read, write, and reason about programs in both text-only and visually grounded contexts.

\paragraph{Text-Only Coding.} 
We reuse the extensive code corpus from the Qwen3 and Qwen3-Coder series. This large-scale dataset spans a wide range of programming languages and domains—including software development, algorithmic problem solving, mathematical reasoning, and agent-oriented tasks—and establishes the model’s foundational understanding of code syntax, algorithmic logic, and general-purpose program generation.

\paragraph{Multimodal Coding.} To address tasks requiring both visual understanding and code generation, we curate data for a diverse suite of multimodal coding tasks. This dataset, sourced from both open-source datasets and internal synthesis pipelines, teaches the model to jointly understand visual inputs and generate functional code. The data covers several key tasks, including: converting UI screenshots into responsive HTML/CSS; generating editable SVG codes from images~\citep{likaixin2025iconstack}; solving visual programming challenges~\citep{li2024mmcode}; answering multimodal coding questions (e.g., StackOverflow posts with images); and transcribing visual representations (such as flowcharts, diagrams, and \LaTeX{} equations) into their respective code or markup. This novel data mixture enables Qwen3-VL to act as a bridge between visual perception and executable logic.

\subsubsection{Video}

The video comprehension capabilities of Qwen3-VL have been substantially advanced, enabling robust modeling of temporal dynamics across frames, fine-grained perception of spatial relationships, and coherent summarization of ultra-long video sequences.
This enhancement is underpinned by a data processing pipeline featuring two principal innovations:

\textbf{Temporal-Aware Video Understanding}. 
(i) Dense Caption Synthesis: For long video sequences, we employ a short-to-long caption synthesis strategy to generate holistic, timestamp-interleaved, and temporally coherent story-level descriptions. Leveraging in-house captioning models, we further produce fine-grained annotations that jointly capture event-level temporal summaries and segment-specific visual details.
(ii) Spatio-Temporal Video Grounding: We curate and synthesize large-scale video data annotated at the levels of objects, actions, and persons to strengthen the model’s spatio-temporal grounding capabilities, thereby improving its capacity for fine-grained video understanding.

\textbf{Video Data Balancing and Sampling}. 
(i) Source Balancing: To ensure data balance and diversity, we assemble a large-scale dataset encompassing various video sources, including instructional content, cinematic films, egocentric recordings, etc. Dataset balance is achieved through systematic curation guided by metadata such as video titles, duration, and categorical labels.
(ii) Length-Adaptive Sampling: During pre-training stages, we dynamically adjust sampling parameters, such as frames per second (fps) and the maximum number of frames, according to different sequence length constraints. 
This adaptive strategy mitigates information loss associated with suboptimal sampling practices (e.g., overly sparse frame selection or excessively low spatial resolution), thus preserving visual details and optimizing training efficacy.

\subsubsection{Science, Technology, Engineering, and Mathematics (STEM)}

Multimodal reasoning lies at the heart of Qwen3-VL, with STEM reasoning constituting its most essential part. Our philosophy follows a divide-and-conquer strategy: we first develop fine-grained visual perception and robust linguistic reasoning capabilities independently, and then integrate them in a synergistic manner to achieve effective multimodal reasoning.

\textbf{Visual Perception Data}. We develop a dedicated synthetic data generation pipeline that constructs geometric diagrams through programmatic (code-based) rendering. Using this pipeline, we generate: (i) 1 million point-grounding samples, such as intersection points, corners, and centers of gravity; and (ii) 2 million perception-oriented visual question answering pairs targeting fine-grained visual understanding of diagrams. To obtain high-fidelity textual descriptions, we further implement a two-stage captioning framework: an initial generation phase followed by rigorous model-based verification. Both stages employ ensembles of specialized models to ensure accuracy and descriptive granularity. This process yields a comprehensive dataset of 6 million richly annotated diagram captions spanning diverse STEM disciplines. 

\textbf{Multi-modal Reasoning Data}. The majority of our multi-modal reasoning data consists of over 60 million K–12 and undergraduate-level exercises, meticulously curated through a rigorous cleaning and reformulation pipeline. During quality filtering, we discard low-quality items, including those with corrupted images, irrelevant content, or incomplete or incorrect answers. During the reformulation stage, we translate exercises between Chinese and English and standardize the format of answers—such as step-by-step solution lists, mathematical expressions, and symbolic notations—to ensure consistency and uniform presentation.  Regarding long CoT problem-solving data, we synthesize over 12 million multimodal reasoning samples paired with images. To ensure the continuity and richness of the reasoning process, we utilize the original rollouts generated by a strong reasoning model. To guarantee data reliability and applicability, each sample’s reasoning trajectory undergoes rigorous validation—combining rule-based checks and model-based verification—and any instances containing ambiguous answers or code-switching are explicitly filtered out. Furthermore, to enhance reasoning quality, we retain only challenging problems via rejection sampling. 

\textbf{Linguistic Reasoning Data}. In addition to multimodal reasoning data, we also incorporate reasoning data from Qwen3, as multimodal reasoning capabilities are largely derived from linguistic reasoning competence.

\subsubsection{Agent}
\textbf{GUI:} 
To endow Qwen3-VL with agentic capability for autonomous interaction with graphical user interfaces (GUIs), we curate and synthesize large-scale, cross-platform data spanning desktop, mobile, and web environments~\citep{ye2025mobile, wang2025opencua, lu2025videoagenttrekcomputerusepretraining}. 
For GUI interface perception, we leverage metadata, parsing tools, and human annotations to construct tasks such as element description, dense captioning, and dense grounding, enabling robust understanding of diverse user interfaces. For agentic capability, we assemble multi-step task trajectories via a self-evolving trajectory-production framework, complemented by targeted human audits; we also carefully design and augment Chain-of-Thought rationales to strengthen planning, decision-making, and reflective self-correction during real-world execution.

\textbf{Function Calling:} For general function calling capabilities with multimodal contexts, we build a multimodal function calling trajectory synthesis pipeline. We first instruct capable models with images to generate user queries and their corresponding function definitions. We then sample model function calls with rationales and synthesize the function responses. This process is repeated until the user's query is judged to be solved. Between each step, trajectories can be filtered out due to formatting errors. Such a pipeline enables us to construct large-scale multimodal function-calling trajectories from vast images, without the need to implement executable functions.

\textbf{Search:} Among the general function calling capabilities, we regard the ability to perform searches as key to facilitating knowledge integration for long-tail entities in real-world scenarios. In this case, we collect multimodal factual lookup trajectories with online image search and text search tools, encouraging the model to perform searches for unfamiliar entities. By doing so, the model learns to gather information from the web to generate more accurate responses.

\section{Post-Training}

\subsection{Training Recipe}

Our post-training pipeline is a three-stage process designed to refine the model's instruction-following capabilities, bolster its reasoning abilities, and align it with human preferences. The specific data and methods for each stage are detailed in the subsequent sections.

\textbf{Supervised Fine-Tuning (SFT).} The first stage imparts instruction-following abilities and activates latent reasoning skills. This is conducted in two phases: an initial phase at a 32k context length, followed by an extension to a 256k context window that focuses on long-document and long-video data. To cater to different needs, we bifurcate the training data into standard formats for non-thinking models and Chain-of-Thought (CoT) formats for thinking models, the latter of which explicitly models the reasoning process.

\textbf{Strong-to-Weak Distillation.} The second stage employs knowledge distillation, where a powerful teacher model transfers its capabilities to our student models. Crucially, we perform this distillation using \textit{text-only} data to fine-tune the LLM backbone. This method proves highly effective, yielding significant improvements in reasoning abilities across both text-centric and multimodal tasks.

\textbf{Reinforcement Learning (RL).} The final stage utilizes RL to further enhance model performance and alignment. This phase is divided into Reasoning RL and General RL. We apply large-scale reinforcement learning across a comprehensive set of text and multimodal domains, including but not limited to math, OCR, grounding, and instruction-following, to improve finer-grained capabilities.

\subsection{Cold Start Data}
\subsubsection{SFT Data}

Our principal objective is to endow the model with the capacity to address a wide spectrum of real-world scenarios. Building upon the foundational capabilities of Qwen2.5-VL, which is proficient in approximately eight core domains and 30 fine-grained subcategories, we have strategically expanded its functional scope. This expansion was achieved by integrating insights from community feedback, academic literature, and practical applications, facilitating the introduction of novel capabilities. These include, but are not limited to, spatial reasoning for embodied intelligence, image-grounded reasoning for fine-grained visual understanding, spatio-temporal grounding in videos for robust object tracking, and the comprehension of long-context technical documents spanning hundreds of pages. Guided by these target tasks and grounded in authentic use cases, we systematically curated the SFT dataset through the meticulous selection and synthesis of samples from open-source datasets and web resources. This targeted data engineering effort has been instrumental in establishing Qwen3-VL as a more comprehensive and robust multimodal foundation model.

This dataset comprises approximately 1,200,000 samples, strategically composed to foster robust multimodal capabilities. This collection is partitioned into unimodal and multimodal data, with one-third consisting of text-only entries and the remaining two-thirds comprising image-text and video-text pairs. The integration of multimodal content is specifically designed to enable the model to interpret complex, real-world scenarios. To ensure global relevance, the dataset extends beyond its primary Chinese and English corpora to include a diverse set of multilingual samples, thereby broadening its linguistic coverage. Furthermore, it simulates realistic conversational dynamics by incorporating both single-turn and multi-turn dialogues contextualized within various visual settings, from single-image to multi-image sequences. Crucially, the dataset also features interleaved image-text examples engineered to support advanced agentic behaviors, such as tool-augmented image search and visually-grounded reasoning. This heterogeneous data composition ensures comprehensive coverage and enhances the dataset's representativeness for training generalizable and sophisticated multimodal agents.

Given Qwen3-VL's native support for a 256K token context length, we employ a staged training strategy to optimize for computational efficiency. This strategy comprises two phases: an initial one-epoch training phase with a sequence length of 32K tokens, followed by a second epoch at the full 256K token length. During this latter stage, the model is trained on a curriculum that interleaves long-context inputs with data sampled at the 32K token length. The long-context inputs include materials such as hundreds of pages of technical documents, entire textbooks, and videos up to two hours in duration.

The quality of training data is a critical determinant of the performance of vision-language models. Datasets derived from open-source and synthetic origins are often plagued by substantial variability and noise, including redundant, irrelevant, or low-quality samples. To mitigate these deficiencies, the implementation of a rigorous data filtering protocol is indispensable. Accordingly, our data curation process incorporates a two-phase filtering pipeline: Query Filtering and Response Filtering.

\textbf{Query Filtering.} In this initial phase, we leverage Qwen2.5-VL to identify and discard queries that are not readily verifiable. Queries with ambiguous instructions are minimally revised to enhance clarity while preserving the original semantic intent. Furthermore, web-sourced queries lacking substantive content are systematically eliminated. Crucially, all remaining queries undergo a final assessment of their complexity and contextual relevance, ensuring only appropriately challenging and pertinent samples are retained for the next stage.

\textbf{Response Filtering.} This phase integrates two complementary strategies:

\begin{itemize}[itemindent=0pt, labelsep=4pt, leftmargin=*]

\item \textbf{Rule-Based Filtering:} A set of predefined heuristics is applied to eliminate responses exhibiting qualitative deficiencies, such as repetition, incompleteness, or improper formatting. To maintain semantic relevance and uphold ethical principles, we also discard any query-response pairs that are off-topic or possess the potential to generate harmful content.

\item \textbf{Model-Based Filtering:} The dataset is further refined by employing reward models derived from the Qwen2.5-VL series. These models conduct a multi-dimensional evaluation of multimodal question-answering pairs. Specifically: (a) answers are scored against a range of criteria, including correctness, completeness, clarity, and helpfulness; (b) for vision-grounded tasks, the evaluation places special emphasis on verifying the accurate interpretation and utilization of visual information; and (c) this model-based approach enables the detection of subtle issues that typically elude rule-based methods, such as inappropriate language mixing or abrupt stylistic shifts.

\end{itemize}

This multi-dimensional filtering framework ensures that only data meeting stringent criteria for quality, reliability, and ethical integrity is advanced to the SFT phase.

\subsubsection{Long-CoT Cold Start Data}

The foundation of our thinking models is a meticulously curated Long Chain-of-Thought (CoT) cold start dataset, engineered to elicit and refine complex reasoning capabilities. This dataset is built upon a diverse collection of queries spanning both pure-text and multimodal data, maintaining an approximate 1:1 ratio between vision-language and text-only samples to ensure balanced skill development.

The multimodal component, while covering established domains such as visual question answering (VQA), optical character recognition (OCR), 2D/3D grounding, and video analysis, places a special emphasis on enriching tasks related to STEM and agentic workflows. This strategic focus is designed to push the model's performance on problems requiring sophisticated, multi-step inference. The pure-text portion closely mirrors the data used for Qwen3, featuring challenging problems in mathematics, code generation, logical reasoning, and general STEM.

To guarantee high quality and an appropriate level of difficulty, we implement a rigorous multi-stage filtering protocol.
\begin{itemize}[itemindent=0pt, labelsep=4pt, leftmargin=*]
\item \textbf{Difficulty Curation:} We selectively retain instances where baseline models exhibited low pass rates or generated longer, more detailed responses. This enriches the dataset with problems that are genuinely challenging for current models.
\item \textbf{Multimodal Necessity Filtering:} For vision-language mathematics problems, we introduce a critical filtering step: we discard any samples that our Qwen3-30B-\textit{nothink} model could solve correctly without access to the visual input. This ensures that the remaining instances genuinely necessitate multimodal understanding and are not solvable via textual cues alone.
\item \textbf{Response Quality Control:} Aligning with the methodology of Qwen3, we sanitize the generated responses. For queries with multiple candidate answers, we first remove those containing incorrect final results. Subsequently, we filter out responses exhibiting undesirable patterns, such as excessive repetition, improper language mixing, or answers that showed clear signs of guessing without sufficient reasoning steps.
\end{itemize}
This stringent curation process yields a high-quality, challenging dataset tailored for bootstrapping advanced multimodal reasoning.

\subsection{Strong-to-Weak Distillation}

We adopt the Strong-to-Weak Distillation pipeline as described in Qwen3 to further improve the performance of lightweight models. This distillation process consists of two main phases:
\begin{itemize}[itemindent=0pt, labelsep=4pt, leftmargin=*]

\item \textbf{Off-policy Distillation:} In the first phase, outputs generated by teacher models are combined to provide response distillation. This helps lightweight student models acquire fundamental reasoning abilities, establishing a strong foundation for subsequent on-policy training.

\item \textbf{On-policy Distillation:} In the second phase, the student model generates the responses based on the provided prompts. These on-policy sequences are then used for fine-tuning the student model. We align the logits predicted by the student and teacher by minimizing the KL divergence.

\end{itemize}

\subsection{Reinforcement Learning}
\subsubsection{Reasoning Reinforcement Learning}

We train models across a diverse set of text and multimodal tasks, including mathematics, coding, logical reasoning, visual grounding, and visual puzzles. Each task is designed so that solutions can be verified deterministically via rules or code executors. 

\paragraph{Data Preparation}
We curate training data from both open-source and proprietary sources and apply rigorous preprocessing and manual annotation to ensure high-quality RL queries. For multimodal queries, we use a preliminary checkpoint of our most advanced vision–language model (Qwen3-VL-235B-A22B) to sample 16 responses per query; any query for which all responses are incorrect is discarded. We then run preliminary RL experiments per task to identify and remove data sources with limited potential for improvement. This process yields approximately 30K RL queries covering a variety of text and multimodal tasks.
For training each model, we sample 16 responses for all queries and filter out easy queries whose pass rate exceeds 90\%. We shuffle and combine task-specific datasets to construct mixed-task batches, ensuring a consistent, predefined ratio of samples per task. The ratio is determined through extensive preliminary experiments.

\paragraph{Reward System}
We implement a unified reward framework that delivers precise feedback across all tasks. The system provides shared infrastructure—data preprocessing, utility functions, and a reward manager to integrate multiple reward types—while the core reward logic is implemented per task. We use task-specific format prompts to guide model outputs to the required formats and therefore do not rely on explicit format rewards. To mitigate code-switching, we apply a penalty when the response language differs from the prompt language.

\paragraph{RL Algorithm}
We employ SAPO~\citep{sapo}, a smooth and adaptive policy-gradient method, for RL training. SAPO delivers consistent improvements across diverse text and multimodal tasks and across different model sizes and architectures.

\subsubsection{General Reinforcement Learning}

The General Reinforcement Learning (RL) stage is designed to enhance the model's generalization capabilities and operational robustness. To this end, we employ a multi-task RL paradigm where the reward function is formulated based on a comprehensive set of tasks from the SFT phase, including VQA, image captioning, OCR, document parsing, grounding, and clock recognition. The reward mechanism is structured to optimize two principal dimensions of model performance:

\begin{itemize}[itemindent=0pt, labelsep=4pt, leftmargin=*]
\item \textbf{Instruction Following:} This dimension evaluates the model's adherence to explicit user directives. It assesses the ability to handle complex constraints on content, format, length, and structured outputs (e.g., JSON), ensuring the generated response precisely matches user requirements.

\item \textbf{Preference Alignment:} For open-ended or subjective queries, this dimension aligns the model's outputs with human preferences by optimizing for helpfulness, factual accuracy, and stylistic appropriateness. This fosters a more natural and engaging user interaction.

\end{itemize}

Furthermore, this stage acts as a corrective mechanism to unlearn strong but flawed knowledge priors ingrained during SFT. We address this by introducing specialized, verifiable tasks designed to trigger these specific errors, such as counter-intuitive object counting and complex clock time recognition. This targeted intervention is designed to supplant erroneous priors with factual knowledge.

Another critical objective is to mitigate inferior behaviors like inappropriate language mixing, excessive repetition, and formatting errors. However, the low prevalence of these issues makes general RL a sample-inefficient correction strategy. To overcome this, we curate a dedicated dataset at this stage. This dataset isolates prompts known to elicit such undesirable behaviors. This focused training enables the application of targeted, high-frequency penalties, effectively suppressing these residual errors.

Feedback for the RL process is delivered via a hybrid reward system that combines two complementary approaches:

\begin{itemize}[itemindent=0pt, labelsep=4pt, leftmargin=*]
\item \textbf{Rule-Based Rewards:} This approach provides unambiguous, high-precision feedback for tasks with verifiable ground truths, such as format adherence and instruction following. By using well-defined heuristics, this method offers a robust mechanism for assessing correctness and effectively mitigates reward hacking, where a model might exploit ambiguities in a learned reward function.

\item \textbf{Model-Based Rewards:} This method employs Qwen2.5-VL-72B-Instruct or Qwen3 as sophisticated judgers. The judge models evaluate each generated response against a ground-truth reference, scoring its quality across multiple axes. This approach offers superior flexibility for assessing nuanced or open-ended tasks where strict, rule-based matching is inadequate. It is particularly effective at minimizing false negatives that would otherwise penalize valid responses with unconventional formatting or phrasing.

\end{itemize}

\subsection{Thinking with Images}

Inspired by the great prior works on "thinking with images" ~\citep{wu2025mmsearch, jin2025search, zheng2025deepeyes, lai2025mini}, we endow Qwen3-VL with similar agentic capabilities through a two-stage training paradigm.

In the first stage, we synthesize a cold-start agentic dataset comprising approximately 10k grounding examples—primarily simple two-turn visual question answering tasks such as attribute detection. We then perform supervised fine-tuning (SFT) on Qwen2.5-VL-32B to emulate the behavior of a visual agent: \textit{think} → \textit{act} → \textit{analyze feedback} → \textit{answer}. To further enhance its reasoning abilities, we apply multi-turn, tool-integrated reinforcement learning (RL). 

In the second stage, we distill the trained Qwen2.5-VL-32B visual agents from the first stage to generate a larger, more diverse dataset of approximately 120k multi-turn agentic interactions spanning a broader range of visual tasks. We then apply a similar cold-start SFT and tool-integrated RL pipeline (now using both distilled and synthesized data) for the post-training of Qwen3-VL.

The multi-turn, tool-integrated RL procedure is nearly identical across both stages, differing only in the underlying data. During RL, we employ three complementary reward signals to encourage robust, tool-mediated reasoning:

\begin{itemize}[itemindent=0pt, labelsep=4pt, leftmargin=*]
    \item \textbf{Answer Accuracy Reward} leverages Qwen3-32B to measure whether the final answer is correct.
    \item \textbf{Multi-Turn Reasoning Reward} leverages Qwen2.5-VL-72B to evaluate whether the assistant correctly interprets tool or environment feedback and arrives at the answer through coherent, step-by-step reasoning.
    \item \textbf{Tool-Calling Reward} encourages appropriate tool usage by comparing the actual number of tool calls to an expert-estimated target. This target is determined offline by Qwen2.5-VL-72B based on task complexity.
\end{itemize}

Early experiments reveal a tendency for models to degenerate into making only a single tool call to hack the first two rewards, regardless of task demands. To mitigate this, we explicitly incorporate the tool-calling reward to promote adaptive tool exploration aligned with task complexity.

\subsection{Infrastructure}



We train the Qwen3-VL series models on Alibaba Cloud's PAI-Lingjun AI Computing Service, which provides the high-performance computing power required for compute-intensive scenarios such as AI and high-performance computing.

During the pretraining phase, the system employs a hybrid parallelism strategy built upon the Megatron-LM framework, integrating Tensor Parallelism (TP), Pipeline Parallelism (PP), Context Parallelism (CP), Expert Parallelism (EP), and ZeRO-1 Data Parallelism (DP). This configuration achieves a fine-grained balance among model scale, computational load, and communication overhead, enabling high hardware utilization and sustaining both high throughput and low communication latency—even at scales of up to 10,000 GPUs.

For local deployment and performance evaluation, we adopt deployment strategies based on either vLLM or SGLang. vLLM utilizes PagedAttention to enable memory-efficient management and high-throughput inference, while SGLang excels at structured generation and handling complex prompts. Together, these backends provide efficient inference and evaluation with stable, efficient, and flexible model inference capabilities.

\section{Evaluation}
\subsection{General Visual Question Answering}
To comprehensively assess the general visual question answering (VQA) capabilities of the Qwen3-VL series, we conduct extensive evaluations on a diverse set of benchmarks, including MMBench-V1.1~\citep{MMBench}, RealWorldQA~\citep{realworldqa2024}, MMStar~\citep{chen2024we}, and SimpleVQA~\citep{cheng2025simplevqa}. As detailed in~\cref{tab:results_flagship},~\cref{tab:results_medium} and~\cref{tab:results_small}, the Qwen3-VL family demonstrates robust and highly competitive performance across a wide spectrum of model sizes, from 2B to 235B parameters.

In the comparison of thinking mode, \ThinkBig achieves the highest score of 78.7 on MMStar. Gemini-2.5-Pro's~\citep{comanici2025gemini} Thinking mode delivers the best overall performance, but \ThinkBig is not far behind. In the non-reasoning mode comparison, \InstructBig obtains the highest scores on MMBench and RealWorldQA, with 89.3/88.9 and 79.2, respectively. 

In the experiments with medium-sized models, Qwen3-VL-32B-Thinking achieves the highest scores on MMBench and RealWorldQA, with 89.5/89.5 and 79.4, respectively. Notably, Qwen3-VL-32B-Instruct even outperforms the Thinking variant on RealWorldQA, scoring 79.0.

The scalability of the Qwen3-VL series is evident in the strong performance of our smaller models. Specifically, the largest model, Qwen3-VL-8B, achieves the highest performance across all five benchmarks. For example, on MMBench-EN, the score in "thinking" mode increases from 79.9 for the 2B model to 85.3 for the 8B model. A similar upward trend is observed on other benchmarks, such as MMStar, where the score rises from 68.1 (2B, thinking) to 75.3 (8B, thinking).

\subsection{Multimodal Reasoning}
We evaluate the Qwen3-VL series on a wide range of multimodal reasoning benchmarks, primarily focusing on STEM-related tasks and visual puzzles, including MMMU~\citep{yue2024mmmu}, MMMU-Pro~\citep{yue2024mmmupro}, MathVision~\citep{wang2024measuring},  MathVision-Wild\textsubscript{\scriptsize photo} (hereafter MathVision\textsubscript{\scriptsize WP})~, MathVista~\citep{lu2023mathvista}, We-Math~\citep{qiao2024we}, MathVerse~\citep{zhang2024mathverse}, DynaMath~\citep{zou2024dynamath}, Math-VR~\citep{duan2025codeplot}, LogicVista~\citep{xiao2024logicvista}, VisualPuzzles~\citep{song2025visualpuzzles},  VLM are Blind~\citep{rahmanzadehgervi2025visionlanguagemodelsblind}, ZeroBench~(Main/Subtasks)~\citep{roberts2025zerobenchimpossiblevisualbenchmark}, and VisuLogic~\citep{xu2025visulogicbenchmarkevaluatingvisual}. As shown in \cref{tab:results_flagship}, the flagship Qwen3-VL model demonstrates outstanding performance across both “non-thinking” and “thinking” models. Notably, \InstructBig achieves the best reported results among non-thinking or low-thinking-budget models on multiple benchmarks, including MathVista\textsubscript{mini}, MathVision, MathVerse\textsubscript{mini}, DynaMath, ZeroBench, VLMsAreBlind, VisuLogic, and VisualPuzzles\textsubscript{Direct}. While, \ThinkBig achieves state-of-the-art results on MathVista\textsubscript{mini}, MathVision, MathVerse\textsubscript{mini}, ZeroBench, LogicVista, and VisuLogic. 

Among medium-sized models, as shown in~\cref{tab:results_medium}, Qwen3-VL-32B demonstrates significant advantages, consistently outperforming Gemini-2.5-Flash and GPT-5-mini. Compared to the previous-generation Qwen2.5-VL-72B model, the medium-sized Qwen3-VL model has already surpassed it on reasoning tasks. This highlights significant progress in VLMs. Additionally, our newly introduced Qwen3-VL-30B-A3B MoE model also delivers competitive results.

Among small-sized models, we compare Qwen3-VL-2B/4B/8B against GPT-5-Nano, with results presented in Table~\ref{tab:results_small}. The 8B variant maintains a clear advantage overall, while the 4B model achieves the highest scores on DynaMath and VisuLogic. Notably, even the smallest 2B model exhibits strong reasoning capabilities. 

\afterpage{
\clearpage
\begin{table}[htp]
\centering
\small
\caption{\textbf{Performance of Qwen3-VL-235B-A22B and top-tier models on visual benchmarks.} The highest scores of the reasoning and non-reasoning models are shown in \textbf{bold} and \underline{underlined}, respectively. Results marked with an ${*}$ are sourced from the technical report. 
${+}$ denotes results with tool use. }
\label{tab:results_flagship}
\renewcommand{\arraystretch}{0.99} 
\begin{adjustbox}{max width=1\textwidth}
\begin{NiceTabular}{c|c|w{c}{1cm}w{c}{1cm}|w{c}{1cm}w{c}{1cm}|w{c}{1cm}w{c}{1cm}|w{c}{1cm}w{c}{1cm}}
\toprule
\multirow[t]{2}{*}{}&\multirow[c]{2}{*}{\textbf{Benchmark}} &
\Block{1-2}{\textbf{Qwen3-VL} \\ \textbf{235B-A22B}} & & 
\Block{1-2}{\textbf{Gemini}\\\textbf{2.5 Pro}} & & 
\Block{1-2}{\textbf{OpenAI}\\\textbf{GPT-5}} & & 
\Block{1-2}{\textbf{Claude}\\\textbf{Opus 4.1}} & \\
 & &  {\scriptsize thinking} & {\scriptsize instruct} & {\scriptsize thinking} & {\scriptsize budget-128} & {\scriptsize high} & {\scriptsize minimal}& {\scriptsize thinking} & {\scriptsize non-thinking}  \\ 
\midrule
\multirow{12}{*}{\shortstack{STEM \\ Puzzle}} 
&MMMU & 80.6 & 78.7 & $81.7^*$ & \underline{80.9} & $\mathbf{84.2^*}$  & $74.4^*$ & 78.4  & 77.2 \\ 
&MMMU-Pro & 69.3 & 68.1  & $68.8^*$ & \underline{71.2} & $\mathbf{78.4^*}$  & $62.7^*$ & 64.8  & 60.7 \\  
&MathVista\textsubscript{\scriptsize mini}& \textbf{85.8} & \underline{84.9} & $82.7^*$ & 77.7 & 81.3  & 50.9 & 75.5  & 74.5 \\
&MathVision & \textbf{74.6} & \underline{66.5}  & $73.3^*$ & 66.0 & 70.9  & 45.8 & 64.3  & 57.7 \\
&MathVision\textsubscript{\scriptsize WP} & \textbf{63.8} & \underline{57.0}  & 63.2 & 56.9 & 62.8  & 40.1 & 54.0  & 46.4 \\
&We-Math & 74.8 & 67.5  & \textbf{80.6} & \underline{74.5} & 73.8  & 51.8 & 65.2  & 60.2 \\
&MathVerse\textsubscript{\scriptsize mini} & \textbf{85.0} & \underline{72.5}  & 82.9 & 65.9 & 84.1  & 43.0 & 70.6  & 68.1 \\
&DynaMath & 82.8 & \underline{79.4} & 80.0 & 78.5 & \textbf{85.4}  & 74.0 & 75.1  & 72.0 \\
&Math-VR & \textbf{66.8} & \underline{65.0} & 64.7* & 54.3 & 58.1  & 21.7 & 54.3  & 38.0 \\
&ZeroBench & \textbf{4} & \underline{2} & 3 & 1 & 2  & \underline{2} & 3  & 1 \\
&VlmsAreBlind& 79.5 & \underline{80.4} & \textbf{86.1} & 78.5 & 80.5  & 53.4 & 77.8  & 72.2 \\
&LogicVista& \textbf{72.2} & 65.8  & 72.0 & \underline{68.7} & 71.8  & 46.3 & 67.3  & 63.5 \\
&VisuLogic& \textbf{34.4} & \underline{29.9}  & 31.6 & 26.9 & 28.5  & 27.2 & 27.9  & 27.2 \\
&VisualPuzzles & 57.2 & 54.7  & \textbf{60.9} & \underline{56.9} & 57.3  & 47.9 & 48.8  & 47.6 \\
\midrule
\multirow{5}{*}{General VQA}
&MMBench-EN & 88.8 & \underline{89.3} & $\mathbf{90.1^*}$ & 88.4 & 83.8 & 81.3 & 79.4 & 83.0 \\ 
&MMBench-CN & 88.6 & \underline{88.9} & $\mathbf{89.7^*}$ & 86.4 & 83.5 & 79.9 & 84.9 & 74.3 \\ 
&RealWorldQA & 81.3 & \underline{79.2} & $78.0^*$ & 76.0 & \textbf{82.8} & 77.3 & 69.9 & 68.5 \\ 
&MMStar & \textbf{78.7} & 78.4 & $77.5^*$ & \underline{78.5} & 76.4 & 65.2 & 72.1 & 71.0 \\ 
&SimpleVQA & 61.3 & 63.0 & \textbf{65.4} & \underline{66.9} & 61.8 & 56.7 & 56.7 & 55.7 \\
\midrule
\multirow{3}{*}{Alignment}
&HallusionBench & \textbf{66.7} & \underline{63.2} & 63.7$^*$ & 60.9 & 65.7 & 53.7 & 60.4  & 55.1 \\  
&MM-MT-Bench & \textbf{8.5} & \underline{8.5} & 8.4$^*$ & 7.6 & 7.6  & 7.5 & 7.8  & 7.9 \\ 
&MIA-Bench & \textbf{92.7} & 91.3 & 92.3 & 91.3 & 92.4  & \underline{92.6} & 91.2  & 90.0 \\
\midrule
\multirow{13}{*}{\shortstack{Document \\ Understanding}}
&DocVQA$_{test}$ & \textbf{96.5} & \underline{97.1} & 92.6 & 94.0 & 91.5 & 89.6 & 92.5 & 89.2 \\  
&InfoVQA$_{test}$ & \textbf{89.5} & \underline{89.2} & 84.2 & 82.9 & 79.0 & 69.9 & 69.4 & 60.9 \\
&AI2D$_{\text{w. M.}}$ & 89.2 & 89.7 & \textbf{90.9} & \underline{90.0} & 89.7 & 84.1 & 86.4 & 84.4 \\
&ChartQA$_{test}$ & \textbf{90.3} & \underline{90.3} & 83.3 & 62.6 & 59.7 & 59.1 & 86.2 & 83.9 \\
&OCRBench & \textbf{875} & \underline{920} & 866 & 872 & 810 & 787 & 764 & 750 \\
&OCRBench\_v2\textsubscript{\scriptsize en} & \textbf{66.8} & \underline{67.1} & 54.3 & 55.2 & 53.0 & 48.2 & 48.4 & 47.2 \\
&OCRBench\_v2\textsubscript{\scriptsize zh} & \textbf{63.5} & \underline{61.8} & 48.5 & 53.1 & 43.2 & 37.7 & 43.7 & 38.0 \\
&CC-OCR & \textbf{81.5} & \underline{82.2} & 77.2 & 76.8 & 68.3 & 66.1 & 69.1 & 66.0 \\
&OmniDocBench\textsubscript{\scriptsize en} & \textbf{0.155} & \underline{0.143} & 0.347 & 0.206 & 0.356 & 0.174 & 0.194 & - \\
&OmniDocBench\textsubscript{\scriptsize zh}& \textbf{0.207} & \underline{0.207} & 0.238 & 0.249 & 0.472 & 0.389 & 0.293 & - \\
&CharXiv(DQ) & 90.5 & \underline{89.4} & \textbf{94.4} & 87.8 & 89.2 & 79.5 & 88.5 & 87.8 \\
&CharXiv(RQ) & 66.1 & 62.1 & 67.9 & \underline{62.9} & $\mathbf{81.1^*}$ & 57.8 & 63.6 & 60.2 \\
&MMLongBench\textsubscript{\scriptsize Doc} & \textbf{56.2} & \underline{57.0} & 55.6 & 51.2 & 51.5 & 42.4 & 54.5 & 48.1 \\
\midrule
\multirow{6}{*}{\makecell{2D/3D \\ Grounding}}
&RefCOCO-avg & \textbf{92.1} & \underline{91.9} &  $74.6^*$ & - & 66.8 & - &  -   &  -   \\  
&CountBench  & \textbf{93.7} & \underline{93.0} & $91.0^*$ & 91.0 & 91.7 & 87.8 & 93.1 & 91.9 \\
&ODinW-13 & \textbf{43.2} & \underline{48.6} & $33.7^*$ & 34.5 & - & - & - & - \\
&ARKitScenes & \textbf{53.7} & \underline{56.9} & - & - & - & - &  -   &  -   \\
&Hypersim& \textbf{11.0} & \underline{13.0} & - & - & - & - &  -   &  -   \\
&SUNRGBD & \textbf{34.9} & \underline{39.4} & 29.7 & - & - & - &  -   &  -   \\
\midrule
\multirow{6}{*}{\shortstack{Embodied/Spatial \\  Understanding}}
&ERQA & 52.5 & \underline{51.3} & 55.3 & 50.3 & \textbf{65.7}$^*$ & $42.0^*$ &  34.8   &  28.0   \\  
&VSI-Bench  & \textbf{60.0} & \underline{62.7} & - & - & - & - & - & - \\
&EmbSpatialBench & \textbf{84.3} & \underline{83.1} & 79.1 & 73.3 & 82.9 & 75.1 & 69.2 & 66.0 \\
&RefSpatialBench & \textbf{69.9} & \underline{65.5} & 36.5 & 35.6 & 23.8 & 23.1 &  -   &  -   \\
&RoboSpatialHome & \textbf{73.9} & \underline{69.4} & 47.5 & 49.2 & 53.5 & 43.6 &  -   &  -   \\
\midrule
\multirow{2}{*}{\shortstack{Multi-Image}}
&BLINK & 67.1 & \underline{70.7} &  $70.6^*$ & 70.0 & \textbf{71.0} & 62.8 &  64.1   &  62.9   \\  
&MUIRBENCH & \textbf{80.1} & 73.0 &  77.2 & \underline{74.0} & 77.5 & 66.5 &  -   &  -   \\  
\midrule
\multirow{6}{*}{\shortstack{Video \\ Understanding}}
&MVBench                        & 75.2 & \underline{76.5} & 69.9 & 65.8 & \textbf{75.3} & 64.6 & 61.4 & 59.0 \\
&Video-MME$_{\text{w/o sub.}}$  & 79.0 & 79.2 & \textbf{85.1} & \underline{80.6} & 84.7 & 77.3 & 75.6 & 73.3 \\
&MLVU$_{\text{M-Avg}}$          & 83.8 & \underline{84.3} & 85.6 & 81.2 & \textbf{86.2} & 78.3 & 73.5 & 71.2 \\
&LVBench                        & 63.6 & 67.7 & \textbf{73.0} & \underline{69.0} &  -   &  -   &  -   &  -   \\
&Charades-STA$_{\text{mIoU}}$   & \textbf{63.5} & \underline{64.8} &  -   &   -  &  -   &  -   &  -   &  -   \\
&VideoMMMU                      & 80.0 & 74.7 & $83.6^*$ & \underline{79.4} & \textbf{84.6}$^*$ & $61.6^*$ & 76.2 & 70.1 \\
&MMVU                      & 71.1 & 68.1 & \textbf{74.9} & \underline{72.2} & 73.0 & 68.1 & 66.4 & 61.4 \\
\midrule
\multirow{3}{*}{\shortstack{Perception \\ with Tool}} 
&V$^*$ & \textbf{85.9} &$\underline{93.7^{+}}$ & 83.8 & 72.7 & 72.8 & 56.7 &  -   &  -   \\
&HRBench4K  & 84.3 &$\underline{85.4^{+}}$ &  \textbf{87.3} & 84.8 & - & - & - & - \\
&HRBench8K & 76.6 &$\underline{82.4^{+}}$ & \textbf{85.4} & 80.1 & - & - & - & - \\
\midrule
\multirow{3}{*}{\shortstack{Multi-Modal \\ Coding}}
&Design2Code    & \textbf{93.4} & \underline{92.0} & 89.2 & 90.3 & 92.5 & 88.9 & 88.5 & 85.3 \\
&ChartMimic     & 78.4 & 80.5 & 83.9 & 79.9 & 62.1 & 41.4 & \textbf{85.2} & \underline{82.9} \\
&UniSVG         & 65.8 & 69.8 & 70.0 & 67.9 & 71.7 & \underline{74.5} & \textbf{73.0} & 72.5 \\
\midrule
\multirow{5}{*}{\shortstack{Multi-Modal \\ Agent}}
&ScreenSpot Pro  & \textbf{61.8} & \underline{62.0} &  - & - & - & - & - & - \\
&OSWorldG & \textbf{68.3} & \underline{66.7} & 45.2 & - & - & - & - & - \\
&AndroidWorld & \textbf{62.0} & \underline{63.7} & - & - & - & - & - & - \\
&OSWorld & \textbf{38.1} & 31.6 & - & - & - & - & - & \underline{44.4} \\
&WindowsAA &\textbf{32.1} & \underline{28.9}& - & - & - & - & - & - \\
\bottomrule
\end{NiceTabular}
\end{adjustbox}
\renewcommand{\arraystretch}{1.0} 
\end{table}
}
\afterpage{
\clearpage
\begin{table}[htp]
\centering
\caption{\textbf{Performance of medium-sized Qwen3-VL models and previous models on visual benchmarks.} The highest scores are shown in \textbf{bold}. Results marked with an ${*}$ are sourced from the technical report. ${+}$ denotes results with tool use. }
\label{tab:results_medium}
\renewcommand{\arraystretch}{0.98} 
\begin{adjustbox}{max width=1\textwidth}
\begin{NiceTabular}{c|c|w{c}{1cm}w{c}{1cm}|w{c}{1cm}w{c}{1cm}|w{c}{1cm}w{c}{1cm}|w{c}{1cm}w{c}{1cm}}
\toprule
\multirow[t]{2}{*}{}&\multirow[c]{2}{*}{\textbf{Benchmark}} &
\Block{1-2}{\textbf{Qwen3-VL} \\ \textbf{30B-A3B}} & & 
\Block{1-2}{\textbf{Qwen3-VL}\\\textbf{32B}} & & 
\Block{1-2}{\textbf{Gemini}\\\textbf{2.5 Flash}} & & 
\Block{1-2}{\textbf{GPT-5}\\\textbf{mini}} & \\
 & &  {\scriptsize thinking} & {\scriptsize instruct} & {\scriptsize thinking} & {\scriptsize instruct} & {\scriptsize thinking} & {\scriptsize non-thinking}& {\scriptsize high} & {\scriptsize minimal}  \\ 
\midrule
\multirow{12}{*}{\shortstack{STEM \\ Puzzle}}  
&MMMU  & 76.0 & 74.2 & 78.1 & 76.0& 77.7  & 76.3 & \textbf{79.0} & 67.9   \\ 
&MMMU-Pro  & 63.0 & 60.4 & \textbf{68.1} & 65.3 & 67.2  & 65.9 & 67.3 & 53.7  \\  
&MathVista\textsubscript{\scriptsize mini} & 81.9 & 80.1 & \textbf{85.9} & 83.8 & 79.4  & 75.3 & 79.1 & 59.6   \\
&MathVision  & 65.7 & 60.2 & 70.2 & 63.4 & 64.3  & 60.7 & \textbf{71.9} & 46.6   \\
&MathVision\textsubscript{\scriptsize WP} & \textbf{58.9} & 52.3 & 58.6 & 54.6 & 53.6  & 49.0 & 56.6  & 42.8 \\
&We-Math & 70.0 & 56.9 & \textbf{71.6} & 63.3 & 53.9  & 60.3 & 70.2 & 51.4    \\
&MathVerse\textsubscript{\scriptsize mini}  & 79.6 & 70.2 & \textbf{82.6} & 76.8 & 77.7  & 75.9 & 78.8 & 36.5   \\
&DynaMath& 80.1 & 73.4 & \textbf{82.0} & 76.7 & 75.9   & 69.7 & 81.4 & 71.3   \\
&Math-VR& 61.7 & 61.3 & \textbf{62.3} & 59.8 & 58.8 & 54.7 & 58.2 & 26.4   \\
&ZeroBench& 0 & 0 & 2 & 1 & 1  & \textbf{3} & \textbf{3} & 2   \\
&VlmsAreBlind& 72.5 & 67.5 & 85.1 & \textbf{87.0} & 77.5  & 75.9 & 75.8 & 62.0   \\
&LogicVista & 65.8 & 53.5 & 70.9 & 62.2& 67.3  & 60.0 & \textbf{71.4} & 50.8   \\
&VisuLogic & 26.6 & 23.0 & \textbf{32.4} & 29.7& 31.0  & 23.3 & 27.2 & 27.6   \\
&VisualPuzzles & 52.0 & 46.2 & 54.7 & 53.2  &41.4  & 45.0 & \textbf{59.3} & 48.2 \\
\midrule
\multirow{5}{*}{General VQA}
&MMBench-EN & 87.0 & 86.1 & \textbf{89.5} & 87.6 & 87.1 & 86.6 & 86.6 & 78.5 \\
&MMBench-CN & 85.9 & 85.3 & \textbf{89.4} & 87.7 & 87.3 & 86.0 & 84.0 & 76.3 \\ 
&RealWorldQA & 77.4 & 73.7 & 78.4 & \textbf{79.0} & 76.0 & 75.7 & \textbf{79.0} & 73.3 \\ 
&MMStar & 75.5 & 72.1 & \textbf{79.4} & 77.7 & 76.5 & 75.8 & 74.1 & 61.3 \\ 
&SimpleVQA & 54.3 & 52.7 & 55.4 & 56.9 & \textbf{63.2} & 59.2 & 56.8 & 50.3 \\
\midrule
\multirow{3}{*}{Alignment}
&HallusionBench & 66.0 & 61.5 & \textbf{67.4} & 63.8 & 63.5 & 59.1 & 63.2 & 55.9 \\  
&MM-MT-Bench & 7.9  & 8.0  & 8.3  & \textbf{8.4}  & 8.1  & 8.0  & 7.7  & 7.4 \\ 
&MIA-Bench & 91.6 & 91.2 & \textbf{92.3} & 91.8 & 91.1 & 90.6 & 92.0 & \textbf{92.3} \\
\midrule
\multirow{13}{*}{\shortstack{Document \\ Understanding}}
&DocVQA$_{test}$ & 95.5 & 95.0 & 96.1 & \textbf{96.9} & 92.8 & 93.0 & 90.5 & 90.6 \\  
&InfoVQA$_{test}$ & 85.6 & 81.8 & \textbf{89.2} & 87.0 & 82.5 & 81.7 & 77.6 & 72.8 \\
&AI2D$_{\text{w. M.}}$ & 86.9 & 85.0 & 88.9 & \textbf{89.5} & 88.7 & 87.7 & 88.2 & 82.9 \\
&ChartQA$_{test}$ & \textbf{89.4} & 86.8 & 89.0 & 88.5 & 60.6 & 69.0 & 57.5 & 57.8 \\
&OCRBench & 839  & \textbf{903}  & 855  & 895  & 853  & 864  & 821  & 807  \\
&OCRBench\_v2\textsubscript{\scriptsize en} & 62.6 & 63.2 & \textbf{68.4} & 67.4 & 52.2 & 50.6 & 52.6 & 45.7 \\
&OCRBench\_v2\textsubscript{\scriptsize zh} & 60.4 & 57.8 & \textbf{62.1} & 59.2 & 43.8 & 43.9 & 45.1 & 41.0 \\
&CC-OCR & 77.8 & \textbf{80.7} & 79.6 & 80.3 & 75.4 & 74.8 & 70.8 & 61.6 \\
&OmniDocBench\textsubscript{\scriptsize en} & 0.165 & 0.183 & \textbf{0.148} & 0.151 & 0.265 & 0.228 & 0.181 & 0.260 \\
&OmniDocBench\textsubscript{\scriptsize zh} & \textbf{0.233} & 0.253 & 0.236 & 0.239 & 0.245 & 0.305 & 0.316 & 0.425 \\
&CharXiv(DQ) & 86.9 & 85.5 & 90.2 & \textbf{90.5} & 90.1 & 85.5 & 89.4 & 78.6 \\
&CharXiv(RQ) & 56.6 & 48.9 & 65.2 & 62.8 & 61.7 & 60.1 & \textbf{68.6} & 48.9 \\
&MMLongBench\textsubscript{\scriptsize Doc} & 47.4 & 47.1 & 54.6 & \textbf{55.4} & 49.0 & 44.6 & 50.3 & 39.6 \\
\midrule
\multirow{6}{*}{\makecell{2D/3D \\ Grounding}}
&RefCOCO-avg & 89.3 & 89.7 & 91.1 & \textbf{91.9} & -    & -    & -    & -    \\  
&CountBench  & 90.0 & 89.8 & 94.1 & \textbf{94.9} & 86.0 & 83.7 & 91.0 & 84.1 \\
&ODinW-13 & 42.3 & \textbf{47.5} & 41.8 & 46.6 & -    & -    & -    & -    \\
&ARKitScenes & 55.6 & \textbf{56.1} & 46.1 & 55.6 & -    & -    & -    & -    \\
&Hypersim & 11.4 & 12.5 & 12.5 & \textbf{14.0} & -    & -    & -    & -    \\
&SUNRGBD & 34.6 & \textbf{38.1} & 33.9 & 37.0 & -    & -    & -    & -    \\
\midrule
\multirow{6}{*}{\shortstack{Embodied/Spatial \\  Understanding}}
&ERQA & 45.3 & 43.0 & 52.3 & 48.8 & -    & - & \textbf{54.0}    & 45.8 \\
&VSI-Bench  & 56.1 & \textbf{63.2} & 61.2 & 61.5 & -    & - & 31.5    & 30.5 \\
&EmbSpatialBench & 80.6 & 76.4 & \textbf{82.7} & 81.5 & -    & - & 80.7    & 72.1 \\
&RefSpatialBench & 54.2 & 53.1 & \textbf{67.2} & 61.4 & -    & -  & 9.0    & 4.0  \\
&RoboSpatialHome & 65.5 & 62.9 & \textbf{74.2} & 64.6 & -    & - & 54.3    & 44.6 \\

\midrule
\multirow{2}{*}{\shortstack{Multi-Image}}
&BLINK & 65.4 & 67.7 & \textbf{68.5} & 67.3 & 68.1 & 66.8 & -    & 56.7 \\
&MUIRBENCH & 77.6 & 62.9 & \textbf{80.3} & 72.8 & 72.7 & 67.5 & -    & 57.5 \\ 
\midrule
\multirow{6}{*}{\shortstack{Video \\ Understanding}}
&MVBench & 72.0 & 72.3 & \textbf{73.2} & 72.8 & -    & -    & -    & -    \\
&Video-MME$_{\text{w/o sub.}}$  & 73.3 & 74.5 & 77.3 & 76.6 & \textbf{79.6} & 75.6 & 78.9 & 71.0 \\
&MLVU$_{\text{M-Avg}}$ & 78.9 & 81.3 & 82.3 & 82.1 & 82.1 & 77.8 & \textbf{83.3} & 71.7 \\
&LVBench & 59.2 & 62.5 & 62.6 & 63.8 & \textbf{64.5}    & 62.2 & - & -    \\
&Charades-STA$_{\text{mIoU}}$   & 62.7 & \textbf{63.5} & 62.8 & 61.2 & -    & -    & -    & -    \\
&VideoMMMU & 75.0 & 68.7 & 79.0 & 71.9 & 73.9 & 65.2    & \textbf{82.5}* & 56.7 \\
&MMVU & 66.1 & 59.8 & 67.9 & 66.8 & \textbf{69.8} & 68.2 & \textbf{69.8} & 64.8 \\
\midrule
\multirow{3}{*}{\shortstack{Perception \\ with Tool}}
&V$^*$ & 81.2 & 89.5$^+$ & 84.8 & \textbf{91.1$^+$} & -    & -    & 78.6 & 63.9 \\  
&HRBench4K  & 77.8 & 82.5$^+$ & 82.1 & \textbf{84.6$^+$} & -    & -    & 78.6 & 66.3 \\
&HRBench8K & 71.3 & 79.3$^+$ & 74.8 & \textbf{81.6$^+$} & -    & -    & 74.4 & 60.9 \\
\midrule
\multirow{5}{*}{\shortstack{Multi-Modal \\ Agent}}
&ScreenSpot Pro  & 57.3 & \textbf{60.5} & 57.1 & 57.9 & -    & -    & -    & -    \\
&OSWorldG & 59.6 & 61.0 & 64.0 & \textbf{65.1} & -    & -    & -    & -    \\
&AndroidWorld & 55.0 & 54.3 & \textbf{63.7} & 57.3 & -    & -    & -    & -    \\
&OSWorld & 30.6 & 30.3 & \textbf{41.0} & 32.6 & -    & -    & -    & -    \\
&WindowsAA & 24.2 & 24.9 & \textbf{42.9} & 30.9 & -    & -    & -    & -    \\
\bottomrule
\end{NiceTabular}
\end{adjustbox}
\renewcommand{\arraystretch}{1.0} 
\end{table}
}
\afterpage{
\clearpage
\begin{table}[htp]
\centering
\caption{\textbf{Performance of small-sized Qwen3-VL models and GPT-5-nano on visual benchmarks.} }
\label{tab:results_small}
\begin{adjustbox}{max width=1\textwidth}
\begin{NiceTabular}{c|c|w{c}{1cm}w{c}{1cm}|w{c}{1cm}w{c}{1cm}|w{c}{1cm}w{c}{1cm}|w{c}{1cm}w{c}{1cm}}
\toprule
\multirow[t]{2}{*}{}&\multirow[c]{2}{*}{\textbf{Benchmark}} &
\Block{1-2}{\textbf{Qwen3-VL} \\ \textbf{2B}} & & 
\Block{1-2}{\textbf{Qwen3-VL}\\\textbf{4B}} & & 
\Block{1-2}{\textbf{Qwen3-VL}\\\textbf{8B}} & & 
\Block{1-2}{\textbf{OpenAI}\\\textbf{GPT-5 nano}} \\
 & &  {\scriptsize thinking} & {\scriptsize instruct} & {\scriptsize thinking} & {\scriptsize instruct} & {\scriptsize thinking} & {\scriptsize instruct}& {\scriptsize high} & {\scriptsize minimal}  \\ 
\midrule
\multirow{12}{*}{\shortstack{STEM \\ Puzzle}}  
&MMMU   & 61.4  & 53.4 & 70.8 & 67.4 & 74.1 & 69.6& 75.8  & 57.6   \\ 
&MMMU-Pro  & 42.5  & 36.5 & 57.0 & 53.2 & 60.4 & 55.9 & 57.2 & 36.5  \\  &MathVista\textsubscript{\scriptsize mini}& 73.6  & 61.3 & 79.5 & 73.7  & 81.4
 & 77.2  &  71.5  & 40.9   \\
&MathVision & 45.9  & 31.6 & 60.0 & 51.6 & 62.7 & 53.9 & 62.2  & 33.2    \\
&MathVision\textsubscript{\scriptsize WP} & 35.5 & 30.9 & 48.7 & 44.4 & 53.3  & 45.4 & 49.3  & 28.3 \\
&MathVerse\textsubscript{\scriptsize mini}  & 66.9  & 52.1 & 75.2 & 46.8 & 77.7 & 62.1 & 74.2  & 27.0  \\
&DynaMath& 66.7 & 54.2 &74.4 & 65.3 & 73.2 & 67.7  & 78.0 & 62.0  \\
&Math-VR& 37.7 & 20.7 & 58.1 & 52.3 & 59.0 & 53.4  & 49.7 & 25.0  \\
&ZeroBench& 0  & 0 & 0 & 0 & 2 & 1  & 1 & 1  \\
&VlmsAreBlind& 50.0  & 56.0 & 68.6 & 71.9 & 69.1 & 74.0 & 66.7 & 40.2  \\
&LogicVista & 50.0  & 35.8 & 61.1 & 53.2 & 65.1 & 55.3 & 59.7 & 40.5  \\
&VisuLogic& 25.4  & 11.5  & 30.2 & 19.0 & 27.5 & 22.5 & 24.5 & 24.0  \\
&VisualPuzzles &37.4  & 34.3 & 48.9 & 43.7 & 51.7 & 47.9 & 43.5 & 31.3 \\
\midrule
\multirow{5}{*}{General VQA}
&MMBench-EN & 79.9 & 78.4 & 84.6 & 83.9 & 85.3 & 84.5 & 78.4 & 50.8 \\
&MMBench-CN & 78.8 & 75.9 & 83.8 & 83.5 & 85.5 & 84.7 & 77.6 & 48.5 \\
&RealWorldQA & 69.5 & 63.9 & 73.2 & 70.9 & 73.5 & 71.5 & 71.8 & 60.7 \\
&MMStar & 68.1 & 58.3 & 73.2 & 69.8 & 75.3 & 70.9 & 68.6 & 41.3 \\
&SimpleVQA & 43.6 & 40.7 & 48.8 & 48.0 & 49.6 & 50.2 & 46.0 & 39.0 \\

\midrule
\multirow{3}{*}{Alignment}
&HallusionBench & 54.9 & 51.4 & 64.1 & 57.6 & 65.4 & 61.1 & 58.4 & 39.3 \\  
&MM-MT-Bench & 6.9 & 5.9 & 7.7 & 7.5 & 8.0 & 7.7 & 6.6  & 6.2 \\ 
&MIA-Bench & 85.6 & 83.6 & 91.0 & 89.7 & 91.5 & 91.1 & 89.9  & 89.6 \\
\midrule
\multirow{13}{*}{\shortstack{Document \\ Understanding}}
&DocVQA$_{test}$ & 92.9 & 93.3 & 94.2 & 95.3 & 95.3 & 96.1 & 88.2 & 78.3 \\  
&InfoVQA$_{test}$ & 77.1 & 72.4 & 83.0 & 80.3 & 86.0 & 83.1 & 68.6 & 49.2 \\
&AI2D$_{\text{w. M.}}$ & 80.4 & 76.9 & 84.9 & 84.1 & 84.9 & 85.7 & 81.9 & 65.7\\
&ChartQA$_{test}$ & 86.6 & 79.1 & 88.8 & 84.6 & 88.6 & 89.6 & 52.1 & 48.6 \\
&OCRBench & 792 & 858 & 808 & 881 & 819 & 896 & 753 & 701 \\
&OCRBench\_v2\textsubscript{\scriptsize en} & 56.4 & 56.3 & 61.8 & 63.7 & 63.9 & 65.4 & 48.1 & 37.9  \\
&OCRBench\_v2\textsubscript{\scriptsize zh} & 51.9 & 53.0 & 55.8 & 57.6 & 59.2 & 61.2 & 33.6 & 27.3  \\
&CC-OCR & 68.3 & 72.8 & 73.8 & 76.2 & 76.3 & 79.9 & 58.9 & 52.9 \\
&OmniDocBench\textsubscript{\scriptsize en} & 0.370 & 0.292 & 0.234 & 0.244 & 0.209 & 0.170 & 0.401 & 0.454 \\
&OmniDocBench\textsubscript{\scriptsize zh}& 0.447 & 0.348 & 0.297 & 0.285 & 0.253 & 0.264 & 0.518 & 0.568 \\
&CharXiv(DQ) & 70.1 & 62.3 & 83.9 & 76.2 & 85.9 & 83.0 & 82.0 & 64.4 \\
&CharXiv(RQ) & 37.1 & 26.8 & 50.3 & 39.7 & 53.0 & 46.4 & 50.1 & 31.7 \\
&MMLongBench\textsubscript{\scriptsize Doc} & 33.8 & 31.6 & 44.4 & 43.5 & 48.0 & 47.9 & 31.8  & 22.1\\
\midrule
\multirow{6}{*}{\makecell{2D/3D \\ Grounding}}
&RefCOCO-avg & 84.8 & 85.6 &  88.2 & 89.0 & 88.2 & 89.1 &  -   &  -   \\  
&CountBench  & 84.1 & 88.4 & 89.4 & 84.9 & 91.5 & 80.5 & 80.0 & 62.9 \\
&ODinW-13 & 36.0 & 43.4 & 39.4 & 48.2 & 39.8 & 44.7 & - & - \\
&ARKitScenes & 47.7 & 56.2 & 46.3 & 56.6 & 46.6 & 56.8 &  -   &  -   \\
&Hypersim& 11.2 & 12.0 & 11.9 & 12.2 & 12.0 & 12.7 &  -   &  -   \\
&SUNRGBD & 28.6 & 33.8 & 28.0 & 34.7 & 30.4 & 36.2 &  -   &  -   \\
\midrule
\multirow{5}{*}{\shortstack{Embodied/Spatial \\  Understanding}}
&ERQA & 41.8 & 28.3 & 47.3 & 41.3 & 46.8 & 45.8 & 45.8 & 37.8 \\
&VSI-Bench & 48.0 & 53.9 & 55.2 & 59.3 & 56.6 & 59.4 & 15.4 & 27.0 \\
&EmbSpatialBench & 75.9 & 69.2 & 80.7 & 79.6 & 81.1 & 78.5 & 74.2 & 50.7 \\
&RefSpatialBench & 28.9 & 30.3 & 45.3 & 46.6 & 44.6 & 54.2 & 12.6 & 2.5 \\
&RoboSpatialHome & 45.3 & 49.1 & 63.2 & 61.7 & 62.0 & 66.9 & 46.1 & 44.8 \\
\midrule
\multirow{2}{*}{\shortstack{Multi-Image}}
&BLINK & 57.2 & 53.8 & 63.4 & 65.8 & 64.7 & 69.1 & 58.3 & 42.2 \\
&MUIRBENCH & 68.1 & 47.4 & 75.0 & 63.8 & 76.8 & 64.4 & 65.7 & 45.7 \\
\midrule
\multirow{6}{*}{\shortstack{Video \\ Understanding}}
&MVBench                        & 64.5 & 61.7 & 69.3 & 68.9 & 69.0 & 68.7 &  -   &  -   \\
&Video-MME$_{\text{w/o sub.}}$  & 62.1 & 61.9 & 68.9 & 69.3 & 71.8 & 71.4 & 66.2 & 49.4 \\
&MLVU$_{\text{M-Avg}}$          & 69.2 & 68.3 & 75.7 & 75.3 & 75.1 & 78.1 & 69.2 & 52.6 \\
&LVBench                        & 47.6 & 47.4 & 53.5 & 56.2 & 55.8 & 58.0 &  -   &  -   \\
&Charades-STA$_{\text{mIoU}}$   & 56.9 & 54.5 & 59.0 & 55.5 & 59.9 & 56.0 &  -   &  -   \\
&VideoMMMU                      & 54.1 & 41.9 & 69.4 & 56.2 & 72.8 & 65.3 &  63.0   &  40.2   \\
&MMVU                      & 48.9 & 41.7 & 58.6 & 50.5 & 62.0 & 58.7 & 63.1 & 51.0 \\
\midrule
\multirow{3}{*}{\shortstack{Perception \\ with Tool}}
&V$^*$ & 69.1 & 75.9$^+$ & 74.9 & 88.0$^+$ & 77.5 & 90.1$^+$ &  -   &  -   \\  
&HRBench4K  & 69.4 & 72.6$^+$ &  73.5 & 81.3$^+$ & 72.4 & 82.3$^+$ & - & - \\
&HRBench8K & 62.6 & 68.9$^+$ & 67.1 & 74.4$^+$ & 68.1 & 78.0$^+$ & - & - \\
\midrule
\multirow{5}{*}{\shortstack{Multi-Modal \\ Agent}}
&ScreenSpot Pro  & 32.2 & 48.5 &  49.2 & 59.5 & 46.6 & 54.6 & - & - \\
&OSWorldG & 41.8 & 46.1 & 53.9 & 58.2 & 56.7 & 58.2 & - & - \\
&AndroidWorld & 46.1 & 36.4 & 52.0 & 45.3 & 50.0 & 47.6 & - & - \\
&OSWorld & 19.0 & 17.0 & 31.4 & 26.2 & 33.9 & 33.9 & - & - \\
&WindowsAA & - & - & 35.5 & 23.4 & 24.1 & 28.8 & - & - \\
\bottomrule
\end{NiceTabular}
\end{adjustbox}
\end{table}
}

\subsection{Alignment and Subjective Tasks}
The ability to follow complex user instructions and reduce potential image-level hallucinations is indispensable for current large vision language models (VLMs).
We assess our models on three representative benchmarks: MM-MT-Bench~\citep{mm_mt_bench}, HallusionBench~\citep{hallusion_bench} and MIA-Bench~\citep{mia-bench}. MM-MT-Bench is a multi-turn LLM-as-a-judge evaluation benchmark for testing multimodal instruction-tuned models. HallusionBench aims at diagnosing image-context reasoning and poses great challenges for current VLMs. MIA-Bench is a more comprehensive benchmark to evaluate models' reactions to users' complex instructions (e.g., creative writing with character limit and compositional instructions). 

As shown in Table~\ref{tab:results_flagship}, our flagship \ModelBig model consistently outperforms other closed-source models. On HallusionBench, our thinking version surpasses Gemini-2.5-pro~\citep{comanici2025gemini}, GPT-5~\citep{gpt5} and Claude opus 4.1~\citep{opus4_1} by 3.0, 1.0, and 6.3 points, respectively. On MIA-Bench, \ThinkBig achieves the overall best score across all the other models, showing our superior multimodal instruction following ability. We also investigate detailed subtask results of MIA-Bench: our model overtakes GPT-5-high-thinking version by 10.0 and 5.0 points in \textit{math} and \textit{textual} subtasks of MIA-Bench, respectively. The same trend can be observed on our smaller-sized models like Qwen3-VL-30B-A3B, and Qwen3-VL-32B, where they overtake other models with comparable sizes. Our 2B/4B/8B series also performs well and shows a negligible drop, especially on MIA-Bench.

\subsection{Text Recognition and Document Understanding}
We compare the Qwen3-VL series with other models of comparable size on document-related benchmarks, including OCR, document parsing, document question answering (QA), and document reasoning.

We evaluate our flagship model, \ModelBig, against state-of-the-art VLMs on the benchmarks listed in Table~\ref{tab:results_flagship}. On OCR-focused parsing benchmarks — including CC-OCR~\citep{yang2024ccocrcomprehensivechallengingocr} and OmniDocBench~\citep{ouyang2024omnidocbenchbenchmarkingdiversepdf} — as well as comprehensive OCR benchmarks such as OCRBench~\citep{Liu_2024_OCRBench} and OCRBench\_v2~\citep{fu2024ocrbenchv2improvedbenchmark}, the \InstructBig model establishes a new state of the art, marginally outperforming its “thinking” counterpart, \ThinkBig.
On OCR-related visual question answering (VQA) benchmarks that require both OCR capability and keyword search — such as DocVQA~\citep{docvqa}, InfoVQA~\citep{Mathew2021InfographicVQA}, AI2D~\citep{Kembhavi2016AI2D}, ChartQA~\citep{masry2022chartqa}, and the CharXiv~\citep{wang2024charxiv} description subset — both the Instruct and Thinking variants achieve comparable performance, demonstrating consistently strong results across these tasks.
Notably, on the reasoning subset of CharXiv — which demands deep chart comprehension and multi-step reasoning — the Thinking variant surpasses the Instruct version and ranks second only to GPT5-thinking and Gemini-2.5-Pro-Thinking.

Furthermore, among the smaller-sized variants in the Qwen3-VL series, both Qwen3-VL-30BA3B models and Qwen3-VL-32B models consistently outperform Gemini-2.5-Flash and GPT-5-mini across most evaluation metrics, as shown in~\cref{tab:results_medium}. Even the compact dense models — Qwen3-VL-8B, Qwen3-VL-4B, and Qwen3-VL-2B — demonstrate remarkably competitive performance on OCR parsing, visual question answering (VQA), and comprehensive benchmark suites, as detailed in~\cref{tab:results_small}. This highlights the exceptional efficiency and strong scalability of the Qwen3-VL architecture across model sizes.

In this version of the Qwen3-VL, we have placed particular emphasis on enhancing its ability to understand long documents. As reported in~\cref{tab:results_flagship}, in the comparison within the flagship models on the MMLongBench-Doc benchmark~\citep{ma2024mmlongbench}, our \ModelBig achieves overall accuracy of 57.0\%/56.2\% under the instruct/thinking settings, showcasing the SOTA performance on the long document understanding task.

\begin{figure}
    \centering
    \includegraphics[width=0.85\linewidth]{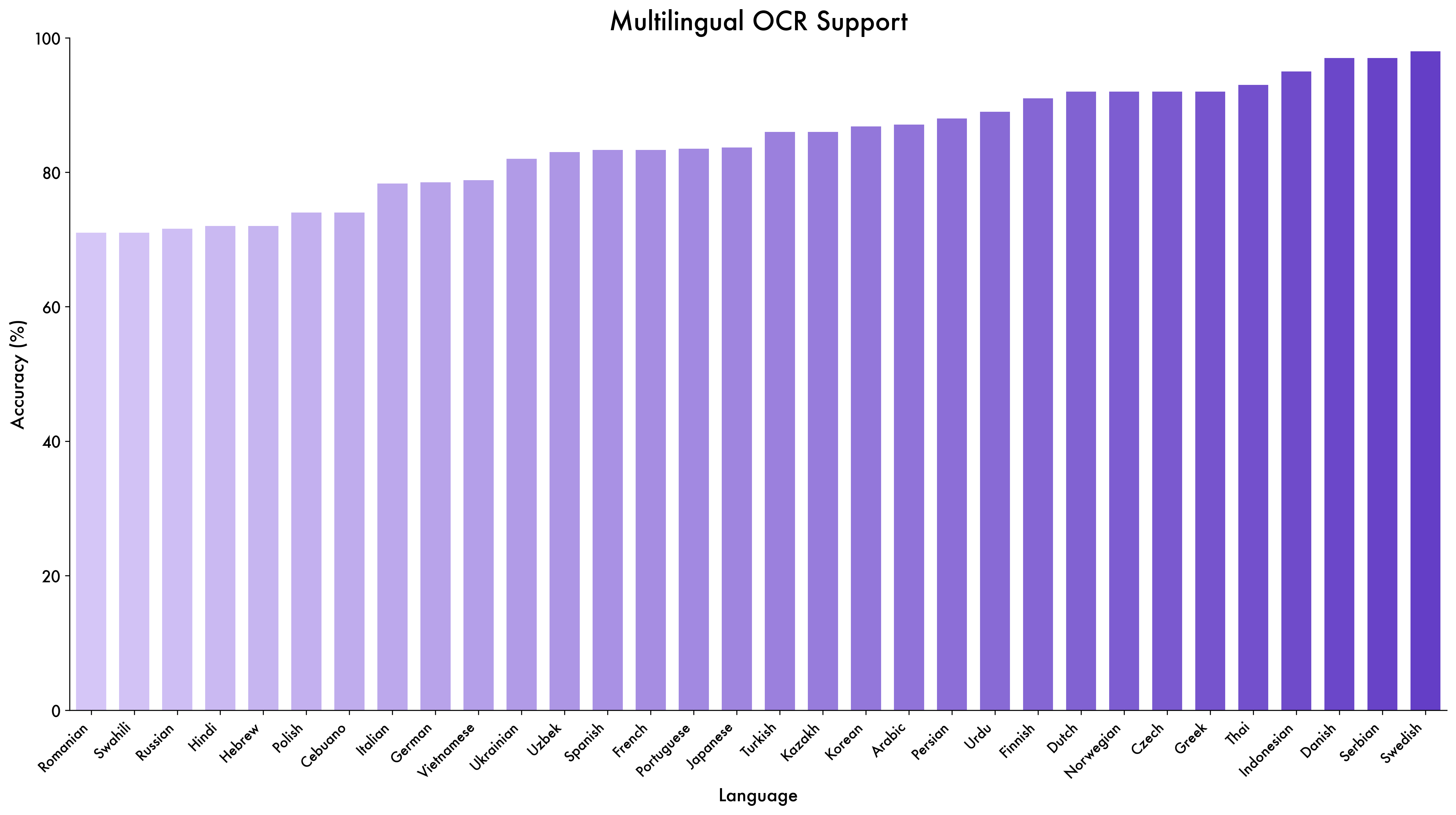}
    \caption{Multilingual OCR performance of our model on a self-built test set. The model achieves over 70\% accuracy on 32 out of 39 supported languages, demonstrating strong and usable multilingual capabilities.}
    \label{fig:multi_lan_ocr}
\end{figure}

Beyond its strong performance on established benchmarks, we have also made substantial strides in multilingual support. This represents a major expansion from the 10 non-English/Chinese languages supported by Qwen2.5-VL to 39 languages in Qwen3-VL. We assess this expanded capability on a newly constructed, in-house dataset. As illustrated in \cref{fig:multi_lan_ocr}, the model's accuracy surpasses 70\%—a threshold we consider practical for real-world usability—on 32 out of the 39 languages tested. This demonstrates that the strong OCR capabilities of Qwen3-VL are not confined to a handful of languages but extend across a broad and diverse linguistic spectrum.


\subsection{2D and 3D Grounding}

In this section, we conduct a comprehensive evaluation of the Qwen3-VL series on both 2D and 3D grounding-related benchmarks and compare the models with state-of-the-art models that possess similar capabilities.

We evaluate Qwen3-VL's 2D grounding capabilities on the referring expression comprehension benchmarks RefCOCO/+/g~\citep{refcoco, refcocog}, the open-vocabulary object detection benchmark ODinW-13~\citep{li2022grounded}, and the counting benchmark CountBench~\citep{paiss2023teaching}. For ODinW-13, we adopt mean Average Precision (mAP) as the evaluation metric by setting confidence scores to 1.0. To ensure comparability with conventional open-set object detection specialist models, we provide all dataset categories simultaneously within the prompt during evaluation. As shown in Table~\ref{tab:results_flagship}, our flagship model, Qwen3-VL-235B-A22B, demonstrates outstanding performance and achieves state-of-the-art (SOTA) results across 2D grounding and counting benchmarks. Notably, it achieves 48.6 mAP on ODinW-13, demonstrating strong performance in multi-target open-vocabulary object grounding. Detailed results for our smaller-scale variants, which also exhibit competitive performance in 2D visual grounding, are presented in Tables~\ref{tab:results_medium} and~\ref{tab:results_small}, respectively.

Moreover, in this version of Qwen3-VL, we enhance its spatial perception capabilities for 3D object localization. We evaluate the Qwen3-VL series against other models of comparable scale on Omni3D~\citep{brazil2023omni3d}, a comprehensive benchmark comprising datasets such as ARKitScenes~\citep{baruch2021arkitscenes}, Hypersim~\citep{roberts2021hypersim}, and SUN RGB-D~\citep{song2015sun}. We employ mean Average Precision (mAP) as our evaluation metric. Each input is an image-text pair consisting of the image and a textual prompt specifying the object category. To ensure a fair comparison with existing VLMs, we set the IoU threshold to 0.15 and report mAP@0.15 on the Omni3D test set, with detection confidence fixed at 1.0. As shown in Table~\ref{tab:results_flagship}, our flagship \ModelBig model consistently outperforms other closed-source models across multiple datasets. Specifically, on the SUN RGB-D dataset~\citep{song2015sun}, the \ThinkBig variants surpass the performance of Gemini-2.5-Pro by 5.2 points. Our smaller-scale variants (e.g., Qwen3-VL-30BA3B, -32B, -8B, -4B, -2B) also exhibit remarkably competitive performance in 3D object grounding, with detailed results provided in Tables~\ref{tab:results_medium} and~\ref{tab:results_small}, respectively.

\subsection{Fine-grained Perception}
We measure the models' fine-grained perception capabilities on three popular benchmarks. The Qwen3-VL series demonstrates a substantial leap in fine-grained visual understanding compared to its predecessor, Qwen2.5-VL-72B. Notably, Qwen3-VL-235B-A22B achieves the state-of-the-art performance across all three benchmarks when augmented with tools—reaching \textbf{93.7} on V*~\citep{Wu_2024_CVPR}, \underline{85.3} on HRBench-4k~\citep{hrbench}, and \underline{82.3} on HRBench-8k ~\citep{hrbench}. This consistent outperformance highlights the effectiveness of architectural refinements and training strategies introduced in Qwen3-VL, particularly in handling high-resolution inputs and subtle visual distinctions critical for fine-grained perception tasks. Second, and perhaps more surprisingly, the performance gains from integrating external tools consistently outweigh those from simply increasing model size. For example, within the Qwen3-VL family, the absolute improvement by adding tools is consistently $\sim$ 5 points across V*. These findings reinforce our conviction that scaling tool-integrated agentic learning in multimodality is a highly promising path forward.

\subsection{Multi-Image Understanding}
Beyond single-image grounded dialogue evaluation, advancing VLMs to handle multi-image understanding is of significant value. This task requires higher-level contextual analysis across diverse visual patterns, enabling more advanced recognition and reasoning capabilities. To this end, we nourish Qwen3-VL with comprehensive cross-image pattern learning techniques, including multi-image referring grounding, visual correspondence, and multi-hop reasoning. We evaluated Qwen3-VL on two prominent multi-image benchmarks: BLINK~\citep{fu2024blink} and MuirBench~\citep{wang2024muirbench}. As shown in~\cref{tab:results_flagship}, Qwen3-VL demonstrates overall superiority in multi-image understanding compared to other leading LVLMs. Specifically, \InstructBig achieves performance comparable to state-of-the-art models such as Gemini-2.5-pro, while \ThinkBig attains a remarkable leading score of 80.1 on MuirBench, surpassing all other models.

\subsection{Embodied and Spatial Understanding}
For embodied and spatial understanding, Qwen3-VL's performance is rigorously benchmarked against leading SOTA models using a challenging suite of benchmarks: ERQA~\citep{team2025gemini}, VSIBench~\citep{yang2025thinking}, EmbSpatial~\citep{du2024embspatial}, RefSpatial~\citep{zhou2025roborefer}, and RoboSpatialHome~\citep{song2025robospatial}. Across these benchmarks, the model showcases exceptional capabilities, rivaling the performance of top-tier models like Gemini-2.5-Pro, GPT-5, and Claude-Opus-4.1.
This success is largely driven by the model's profound spatial understanding, which stems from its training on high-resolution visual data with fine-grained pointing, relative-position annotations, and QA pairs. This capability is clearly validated by its strong results on EmbSpatial, RefSpatial, and RoboSpatialHome, where Qwen3-VL-235B-A22 achieves scores of 84.3, 69.9, and 73.9, respectively.
Moreover, its embodied intelligence is significantly enhanced through the integration of pointing, grounding, and spatio-temporal perception data during training, leading to top-tier scores of 52.5 on ERQA~\citep{team2025gemini} and 60.0 on VSIBench~\citep{yang2025thinking} for Qwen3-VL-235B-A22B.

\subsection{Video Understanding}

Benefiting from the scaling of training data and key architectural enhancements, Qwen3-VL demonstrates substantially improved video understanding capabilities. In particular, the integration of interleaved MRoPE, the insertion of textual timestamps, and scaling temporally dense video captions collectively enable the Qwen3-VL 8B variant to achieve performance competitive with the significantly larger Qwen2.5-VL 72B model.

We conduct a comprehensive evaluation across a diverse set of video understanding tasks, encompassing general video understanding (VideoMME~\citep{fu2024video}, MVBench~\citep{li2024mvbench}), temporal video grounding (Charades-STA~\citep{gao2017tall}), video reasoning (VideoMMMU~\citep{hu2025video}, MMVU~\citep{zhao2025mmvu}), and long-form video understanding (LVBench~\citep{wang2024lvbench}, MLVU~\citep{zhou2024mlvu}). In comparison with state-of-the-art proprietary models — including Gemini 2.5 Pro, GPT-5, and Claude Opus 4.1, Qwen3-VL demonstrates competitive and, in several cases, superior performance.
In particular, our flagship model, Qwen3-VL-235B-A22B-Instruct, achieves performance on par with leading models such as Gemini 2.5 Pro (with a thinking budget of 128) and GPT-5 minimal on standard video understanding benchmarks. By extending the context window to 256K tokens, it further attains or even surpasses Gemini-2.5-Pro on long-video evaluation tasks, most notably on MLVU. 

Regarding evaluation details, we imposed a cap of 2,048 frames per video for all benchmarks, ensuring that the total number of video tokens did not exceed 224K. The maximum number of tokens per frame was set to 768 for VideoMMMU and MMVU, and to 640 for all other benchmarks. Additionally, videos from Charades-STA were sampled at 4 frames per second (fps), while a rate of 2 fps was used for all other benchmarks. For VideoMMMU, we employed a model-based judge for evaluation, as rule-based scoring proved insufficiently accurate. It is worth noting that our comparison cannot guarantee full fairness due to resource and API limitations, which constrained the number of input frames used during evaluation: 512 for Gemini 2.5 Pro, 256 for GPT-5, and 100 for Claude Opus 4.1.

\subsection{Agent}
We evaluate UI perception with GUI-grounding tasks (ScreenSpot~\citep{cheng2024seeclick}, ScreenSpot Pro~\citep{screenspotpro}, OSWorldG\citep{xie2025scalingcomputerusegroundinguser}) and assess decision-making abilities through online environment evaluations (AndroidWorld~\citep{rawles2024androidworld}, OSWorld~\citep{xie2025osworld,osworld_verified}). For GUI grounding, Qwen3-VL-235B-A22B achieves state-of-the-art performance across multiple tasks, covering interactive interfaces on desktop, mobile, and PC, and demonstrating exceptionally strong UI perception capabilities. For online evaluations, Qwen3-VL 32B scores 41 on OSWorld and 63.7 on AndroidWorld, which surpasses the current foundation VLMs. Qwen3-VL demonstrates exceptionally strong planning, decision-making, and reflection abilities as a GUI agent. Furthermore, smaller Qwen3-VL models have demonstrated highly competitive performance on these benchmarks.

\subsection{Text-Centric Tasks}
To comprehensively evaluate the text-centric performance of Qwen3-VL, we adopt automatic benchmarks
to assess model performance on both instruct and thinking models. These benchmarks can be categorized into the following key types: (1) \textbf{Knowledge:} MMLU-Pro~\citep{mmlupro}, MMLU-Redux~\citep{DBLP:journals/corr/abs-2406-04127}, GPQA~\citep{DBLP:journals/corr/abs-2311-12022}, SuperGPQA~\citep{DBLP:journals/corr/abs-2502-14739}, (2) \textbf{Reasoning:} AIME-25~\citep{aime25}, HMMT-25~\citep{hmmt25}, LiveBench (2024-11-25)~\citep{DBLP:journals/corr/abs-2406-19314}, (3) \textbf{Code:} LiveCodeBench v6~\citep{DBLP:journals/corr/abs-2403-07974}, CFEval, OJBench~\citep{DBLP:journals/corr/abs-2506-16395}, (4) \textbf{Alignment Tasks:} IFEval~\citep{DBLP:journals/corr/abs-2311-07911}, Arena-Hard v2~\citep{DBLP:journals/corr/abs-2406-11939} \footnote{For reproducibility of Arena-Hard v2, we report the win rates evaluated by GPT-4.1.}, Creative Writing v3~\citep{DBLP:journals/corr/abs-2312-06281} \footnote{For reproducibility of Creative Writing v3, we report the scores evaluated by Claude 3.7 Sonnet.}, WritingBench~\citep{DBLP:journals/corr/abs-2503-05244}, (5) \textbf{Agent:} BFCL-v3~\citep{patil2025bfcl}, TAU2-Retail, TAU2-Airline, TAU2-Telecom, (6) \textbf{Multilingual:} MultiIF~\citep{DBLP:journals/corr/abs-2410-15553}, MMLU-ProX, INCLUDE~\citep{DBLP:conf/iclr/RomanouFSNSMACH25}, PolyMATH~\citep{DBLP:journals/corr/abs-2504-18428}.

\paragraph{Evaluation Settings} For Qwen3-VL instruct models including 235B-A22B, 32B and 30B-A3B, we configure the sampling hyperparameters with temperature = 0.7, top-p = 0.8, top-k = 20, and presence penalty = 1.5. As for the small instruct models including 8B, 4B and 2B, we set the temperature = 1.0, top-p = 1.0, top-k = 40, and presence penalty = 2.0. We set the max output length to 32,768 tokens.

For Qwen3-VL thinking models with Mixture-of-Experts (MoE) architecture, we set the sampling temperature to 0.6, top‑p to 0.95, and top‑k to 20. For the dense thinking models, we set temperature = 1.0, top‑p = 0.95, top‑k = 20, and additionally apply a presence penalty of 1.5 to encourage greater output diversity. We set the max output length to 32,768 tokens, except AIME-25, HMMT-25 and LiveCodeBench v6 where we extend the length to 81,920 tokens to provide sufficient thinking space.

The detailed results are as follows.

\paragraph{Qwen3-VL-235B-A22B}
We compare our flagship model Qwen3-VL-235B-A22B with the leading instruct and thinking models. For the Qwen3-VL-235B-A22B-Instruct, we take Qwen3-235B-A22B-Instruct-2507, DeepSeek V3 0324, and Claude-Opus-4 (without thinking) as the baselines. For the Qwen3-VL-235B-A22B-Thinking, we take Qwen3-235B-A22B-Thinking-2507, OpenAI o3 (medium), Claude-Opus-4 (with thinking) as baselines. We present the evaluation results in Table \ref{tab: exp_qwen3_vl_235b_a22b_nothinking} and Table \ref{tab: exp_qwen3_vl_235b_a22b_thinking}. 

\begin{itemize}[itemindent=0pt, labelsep=4pt, leftmargin=*]
    \item From Table \ref{tab: exp_qwen3_vl_235b_a22b_nothinking}, Qwen3-VL-235B-A22B-Instruct achieves competitive results, comparable to or even surpassing the other leading models, including DeepSeek V3 0324, Claude-Opus-4 (without thinking), and our previous flagship model Qwen3-235B-A22B-Instruct-2507. Particularly, Qwen3-VL-235B-A22B-Instruct exceeds other models on reasoning-demand tasks (e.g., mathematics and coding). It is worth noting that DeepSeek V3 0324 and Qwen3-235B-A22B-Instruct-2507 are Large Language Models, while Qwen3-VL-235B-A22B-Instruct is a Vision Language model which can process visual and textual tasks. This means that Qwen3-VL-235B-Instruct has achieved the integration of visual and textual capabilities.
    \item From Table \ref{tab: exp_qwen3_vl_235b_a22b_thinking}, Qwen3-VL-235B-A22B-Thinking also achieves competitive results compared with other leading thinking models. Qwen3-VL-235B-A22B-Thinking exceeds OpenAI o3 (medium) and Claude-Opus-4 (with thinking) on AIME-25 and LiveCodeBench v6, which means Qwen3-VL-235B-A22B-Thinking has better reasoning ability.
\end{itemize}

\begin{table}[t]
\centering
\caption{\textbf{Comparison among Qwen3-VL-235B-A22B (Instruct) and other baselines. The highest and second-best scores are shown in bold and \underline{underlined} respectively.}}
\setlength{\tabcolsep}{3.0pt}
\begin{tabular}{lllllll}
\toprule
    &  \textbf{Benchmark }  & \begin{tabular}[c]{@{}l@{}}\textbf{Qwen3-VL}\\\textbf{235B-A22B}\\{\scriptsize Instruct}\end{tabular}
    & \begin{tabular}[c]{@{}l@{}}\textbf{Qwen3}\\\textbf{235B-A22B}\\{\scriptsize Instruct-2507}\end{tabular} 
    & \begin{tabular}[c]{@{}l@{}}\textbf{Deepseek V3}\\\textbf{0324}\\{\scriptsize }\end{tabular}
    & \begin{tabular}[c]{@{}l@{}}\textbf{Claude-Opus-4}\\\textbf{(Without thinking)}\\{\scriptsize }\end{tabular} \\
\midrule
\multirow{4}{*}{Knowledge}             
                                       & MMLU-Pro               & 81.8                       & \underline{83.0}                     & 81.2               & \textbf{86.6}                           \\
                                       & MMLU-Redux             & 92.2                       & \underline{93.1}                     & 90.4               & \textbf{94.2}                        \\
                                       & GPQA                   & 74.3                       & \textbf{77.5}                     & 68.4         & \underline{74.9}                        \\
                                       & SuperGPQA              & \underline{60.4}                       & \textbf{62.6}                     & 57.3                  & 56.5                           \\
\midrule
                                       & AIME-25                 & \textbf{74.7}                       & \underline{70.3}                     & 46.6         & 33.9                        \\

Reasoning                              & HMMT-25                 & \textbf{57.4}                       & \underline{55.4}                     & 27.5               & 15.9                        \\
                                       & LiveBench {\tiny 2024-11-25}          & \underline{74.8}                       & \textbf{75.4}                     & 66.9               & 74.6                        \\

\midrule
\multirow{4}{*}{\begin{tabular}[c]{@{}l@{}}Alignment\\Tasks\end{tabular}}
                                       & IFEval                 & \underline{87.8}                       & \textbf{88.7}                     & 82.3               & 87.4                        \\
                                       & Arena-Hard V2 {\scriptsize (winrate)} & \underline{77.4}                       & \textbf{79.2}                     & 45.6               & 51.5                        \\
                                       & Creative Writing v3    & \underline{86.5}                       & \textbf{87.5}                     & 81.6               & 83.8                        \\
                                       & WritingBench           & \textbf{85.5}                       & \underline{85.2}                     & 74.5               & 79.2                        \\
\midrule

\multirow{2}{*}{Coding   \& Agent}                                 & LiveCodeBench v6     & \textbf{54.3}                       & \underline{51.8}                     & 45.2               & 44.6                           \\
                                        & BFCL-v3                  & \underline{67.7}                       & \textbf{70.9}                     & 64.7               & 60.1                        \\
                                     

\midrule
\multirow{4}{*}{Multilingualism}       & MultiIF                & \underline{76.3}                       & \textbf{77.5}                     & 66.5               & -                           \\
                                       & MMLU-ProX              & \underline{77.8}                       & \textbf{79.4}                     & 75.8               & -                           \\
                                       & INCLUDE                & \underline{80.0}                       & 79.5                     & \textbf{80.1}               & -                           \\
                                       & PolyMATH               & \underline{45.1}          & \textbf{50.2}  & 32.2               & 30.0            \\              
\bottomrule
\end{tabular}
\label{tab: exp_qwen3_vl_235b_a22b_nothinking}
\end{table}

\begin{table}[t]
\centering
\caption{\textbf{Comparison among Qwen3-VL-235B-A22B (Thinking) and other reasoning baselines. The highest and second-best scores are shown in bold and \underline{underlined} respectively.}}
\setlength{\tabcolsep}{4.2pt}
\begin{tabular}{lllllll}
\toprule

    &  \textbf{Benchmark }  & \begin{tabular}[c]{@{}l@{}}\textbf{Qwen3-VL}\\\textbf{235B-A22B}\\{\scriptsize Thinking}\end{tabular}
    & \begin{tabular}[c]{@{}l@{}}\textbf{Qwen3}\\\textbf{235B-A22B}\\{\scriptsize Thinking-2507}\end{tabular} 
    & \begin{tabular}[c]{@{}l@{}}\textbf{OpenAI o3}\\\textbf{(medium)}\\{\scriptsize }\end{tabular}
    & \begin{tabular}[c]{@{}l@{}}\textbf{Claude-Opus-4}\\\textbf{(With thinking)}\\{\scriptsize }\end{tabular} \\
\midrule
\multirow{4}{*}{Knowledge}            
                                       & MMLU-Pro               & 83.8                       & \underline{84.4}                     & \textbf{85.9}               & -                           \\
                                       & MMLU-Redux             & 93.7                       & 93.8                     & \textbf{94.9}               & \underline{94.6}                        \\
                                       & GPQA                   & 77.1                       & \underline{81.1}                     & \textbf{83.3(high)}         & 79.6                        \\
                                       & SuperGPQA              & \underline{64.3}                       & \textbf{64.9}                     & -                  & -                           \\
\midrule
                                       & AIME-25                 & \underline{89.7}                       & \textbf{92.3}                     & 88.9(high)         & 75.5                        \\

Reasoning                              & HMMT-25                 & 77.4                       & \textbf{83.9}                     & \underline{77.5}               & 58.3                        \\
                                       & LiveBench {\tiny 2024-11-25}          & \textbf{79.6}                       & \underline{78.4}                     & 78.3               & 78.2                        \\

\midrule
                                       & LiveCodeBench v6                 & \underline{70.1}                       & \textbf{74.1}                     & 58.6              & 48.9                        \\

Coding                                 & CFEval                 & 1964                       & \textbf{2134}                     & \underline{2043}               & -                           \\
                                       & OJBench                & \underline{27.5}                       & \textbf{32.5}                     & 25.4               & -                           \\
                                     
\midrule
\multirow{4}{*}{\begin{tabular}[c]{@{}l@{}}Alignment\\Tasks\end{tabular}}
                                       & IFEval                 & 88.2                       & 87.8                     & \textbf{92.1}               & \underline{89.7}                        \\
& Arena-Hard V2 {\scriptsize (winrate)} & 74.8                       & \underline{79.7}                     & \textbf{80.8}               & 59.1                        \\
                                       & Creative Writing v3    & 85.7                       & \underline{86.1}                     & \textbf{87.7}               & 83.8                        \\
                                       & WritingBench           & \underline{86.7}                       & \textbf{88.3}                     & 85.3               & 79.1                        \\
\midrule
\multirow{4}{*}{Agent}                 & BFCL-v3                  & 71.8                       & \underline{71.9}                     & \textbf{72.4}               & 61.8                        \\
                                       & TAU2-Retail            & 67.0                       & \underline{71.9}                     & \textbf{76.3}               & -                           \\
                                       & TAU2-Airline           & \underline{62.0}                       & 58.0                     & \textbf{70.0}               & -                           \\
                                       & TAU2-Telecom           & 44.7                       & \underline{45.6}                     & \textbf{60.5}               & -                           \\
\midrule
\multirow{4}{*}{Multilingualism}       & MultiIF                & 79.1                       & \textbf{80.6}                     & \underline{80.3}               & -                           \\
                                       & MMLU-ProX              & 80.6                       & \underline{81.0}                     & \textbf{83.3}               & -                           \\
                                       & INCLUDE                & 80.0                       & \underline{81.0}                     & \textbf{86.6}               & -                           \\
                                       & PolyMATH               & \underline{57.8}          & \textbf{60.1}  & 49.7               & -            \\              
\bottomrule
\end{tabular}
\label{tab: exp_qwen3_vl_235b_a22b_thinking}
\end{table}

\paragraph{Qwen3-VL-32B / 30B-A3B}
We compare our Qwen3-VL-32B and Qwen3-VL-30B-A3B models with their corresponding text-only counterparts, namely Qwen3-32B, Qwen3-30B-A3B, and Qwen3-30B-A3B-2507. We present the evaluation results in Table \ref{tab: exp_qwen3_vl_30b_a3b_nothinking} and Table \ref{tab: exp_qwen3_vl_30b_a3b_thinking}.

\begin{itemize}[itemindent=0pt, labelsep=4pt, leftmargin=*]
    \item From Table \ref{tab: exp_qwen3_vl_30b_a3b_nothinking}, for instruct models, Qwen3-VL-32B and Qwen3-VL-30B-A3B show significant performance improvement compared with Qwen3-32B and Qwen3-30B-A3B on all the benchmarks. Qwen3-VL-30B-A3B achieves comparable or even better results compared with Qwen3-30B-A3B-2507, particularly AIME-25 and HMMT-25.
    \item From Table \ref{tab: exp_qwen3_vl_30b_a3b_thinking}, for thinking models, Qwen3-VL-32B and Qwen3-VL-30B-A3B surpass the baselines in most of the benchmarks. Qwen3-VL-30B-A3B also shows comparable performance compared with Qwen3-30B-A3B-2507.
\end{itemize}  

\begin{table}[t]
\centering
\caption{\textbf{Comparison among Qwen3-VL-32B-Instruct, Qwen3-VL-30B-A3B-Instruct, and corresponding baselines.}}
\setlength{\tabcolsep}{4.0pt}
\begin{tabular}{lllllll}
\toprule
    &  \textbf{Benchmark }  & \begin{tabular}[c]{@{}l@{}}\textbf{Qwen3-VL}\\\textbf{32B}\\{\scriptsize Instruct}\end{tabular}
    & \begin{tabular}[c]{@{}l@{}}\textbf{Qwen3}\\\textbf{32B}\\{\scriptsize Instruct}\end{tabular} 
    & \begin{tabular}[c]{@{}l@{}}\textbf{Qwen3-VL}\\\textbf{30B-A3B}\\{\scriptsize Instruct}\end{tabular}
    & \begin{tabular}[c]{@{}l@{}}\textbf{Qwen3}\\\textbf{30B-A3B}\\{\scriptsize Instruct}\end{tabular}
    & \begin{tabular}[c]{@{}l@{}}\textbf{Qwen3}\\\textbf{30B-A3B}\\{\scriptsize Instruct-2507}\end{tabular} \\
\midrule
\multirow{4}{*}{Knowledge}           
                                       & MMLU-Pro               & 78.6                       & 71.9                     & 77.8      & 69.1         & 78.4                           \\
                                       & MMLU-Redux             & 89.8                       & 85.7                     & 88.4           & 84.1    & 89.3                        \\
                                       & GPQA                   & 68.9                       & 54.6                     & 70.4        & 54.8 & 70.4                        \\
                                       & SuperGPQA              & 54.6                       & 43.2                     & 53.1           & 42.2       & 53.4                           \\
\midrule
\multirow{4}{*}{Reasoning}     
                                       & AIME-25                 & 66.2                       & 20.2                     & 69.3         & 21.6 & 61.3                        \\

                            & HMMT-25                 & 46.1                       & 10.9                     & 50.6       &   12.0     & 43.0                        \\
                                       & LiveBench {\tiny 2024-11-25}          & 72.2                       & 31.3                     & 65.4      &  59.4       & 69.0                        \\

\midrule
\multirow{4}{*}{\begin{tabular}[c]{@{}l@{}}Alignment\\ Tasks\end{tabular}}
                                       & IFEval                 & 84.7                       & 83.2                     & 85.8           &83.7    & 84.7 \\
                                       & Arena-Hard V2 {\scriptsize (winrate)} & 64.7                       & 37.4                     & 58.5          &   24.8  & 69.0                        \\
                                       & Creative Writing v3    & 85.6                       & 80.6                     & 84.6     &68.1          & 86.0                        \\
                                       & WritingBench           & 82.9                       & 81.3                     & 82.6      &72.2         & 85.5                        \\
\midrule
\multirow{2}{*}{\begin{tabular}[c]{@{}l@{}}Coding \& Agent \end{tabular}}
                                & LiveCodeBench v6     & 43.8                       & 29.1                     & 42.6               & 29.0 & 43.2                           \\
                                       & BFCL-v3                  & 70.2                       & 63.0                     & 66.3      &58.6         & 65.1                        \\
                                     

\midrule
\multirow{4}{*}{Multilingualism}       & MultiIF                & 72.0                       & 70.7                     & 66.1       & 70.8        & 67.9                           \\
                                       & MMLU-ProX              & 73.4                       & 69.3                     & 70.9        & 65.1       & 72.0                           \\
                                       & INCLUDE                & 74.0                       & 69.6                     & 71.6      & 67.8         & 71.9                           \\
                                       & PolyMATH               & 40.5          & 22.5  & 44.3        & 23.3       & 43.1            \\              
\bottomrule
\end{tabular}
\label{tab: exp_qwen3_vl_30b_a3b_nothinking}
\end{table}

\begin{table}[t]
\centering
\caption{\textbf{Comparison among Qwen3-VL-32B (Thinking), Qwen3-VL-30B-A3B (Thinking), and corresponding baselines.}}
\setlength{\tabcolsep}{4.0pt}
\begin{tabular}{lllllll}
\toprule
    &  \textbf{Benchmark } & \begin{tabular}[c]{@{}l@{}}\textbf{Qwen3-VL}\\\textbf{32B}\\{\scriptsize Thinking}\end{tabular} 
    & \begin{tabular}[c]{@{}l@{}}\textbf{Qwen3}\\\textbf{32B}\\{\scriptsize Thinking}\end{tabular} 
    & \begin{tabular}[c]{@{}l@{}}\textbf{Qwen3-VL}\\\textbf{30B-A3B}\\{\scriptsize Thinking}\end{tabular}
    & \begin{tabular}[c]{@{}l@{}}\textbf{Qwen3}\\\textbf{30B-A3B}\\{\scriptsize Thinking}\end{tabular} 
    & \begin{tabular}[c]{@{}l@{}}\textbf{Qwen3}\\\textbf{30B-A3B}\\{\scriptsize Thinking-2507}\end{tabular} \\
\midrule
\multirow{4}{*}{Knowledge}       & MMLU-Pro               & 82.1                  & 79.1               & 80.5                   & 78.5                   & 80.9                        \\
                                 & MMLU-Redux             & 91.9                  & 90.9               & 90.9                   & 89.5                   & 91.4                        \\
                                 & GPQA                   & 73.1                  & 68.4               & 74.4                   & 65.8                   & 73.4                        \\
                                 & SuperGPQA              & 59.0                  & 54.1               & 56.4                   & 51.8                   & 56.8                        \\
\midrule
\multirow{3}{*}{Reasoning}       & AIME-25                 & 83.7                  & 72.9               & 83.1                   & 70.9                   & 85.0                        \\
                                 & HMMT-25                 & 64.6                  & 51.8               & 67.6                   & 49.8                   & 71.4                        \\
& LiveBench {\tiny 2024-11-25}     & 74.7                  & 65.7               & 72.1                   & 74.3                   & 76.8                        \\

\midrule
\multirow{3}{*}{Coding}          & LiveCodeBench v6     & 65.6                  & 60.6               & 64.2                   & 57.4                   & 66.0                        \\
                                 & CFEval                 & 1842                  & 1986               & 1894                   & 1940                   & 2044                        \\
                                 & OJBench                & 20.0                  & 24.1               & 23.4                   & 20.7                   & 25.1                        \\
\midrule
\multirow{4}{*}{\begin{tabular}[c]{@{}l@{}}Alignment\\ Tasks\end{tabular}} & IFEval                 & 87.8                  & 85.0               & 81.7                   & 86.5                   & 88.9                        \\
                                 & Arena-Hard V2 {\scriptsize (winrate)} & 60.5                  & 50.3               & 56.7                   & 36.3                   & 56.0                        \\
                                 & Creative Writing v3    & 83.3                  & 84.4               & 82.5                   & 79.1                   & 84.4                        \\
                                 & WritingBench           & 86.2                  & 78.4               & 85.2                   & 77.0                   & 85.0                        \\
\midrule
\multirow{4}{*}{Agent}           & BFCL-v3                & 71.7                  & 70.3               & 68.6                   & 69.1                   & 72.4                        \\
                                 & TAU2-Retail            & 59.4                  & 59.6               & 64.0                   & 34.2                   & 58.8                        \\
                                 & TAU2-Airline           & 52.5                  & 38.0               & 48.0                   & 36.0                   & 58.0                        \\
                                 & TAU2-Telecom           & 46.9                  & 26.3               & 27.2                   & 22.8                   & 26.3                        \\
\midrule
\multirow{4}{*}{Multilingualism} & MultiIF                & 78.0                  & 73.0               & 73.0                   & 72.2                   & 76.4                        \\
                                 & MMLU-ProX              & 77.2                  & 74.6               & 76.1                   & 73.1                   & 76.4                        \\
                                 & INCLUDE                & 76.3                  & 73.7               & 74.5                   & 71.9                   & 74.4                        \\
                                 & PolyMATH               & 52.0                  & 47.4               & 51.7                   & 46.1                   & 52.6   \\
\bottomrule
\end{tabular}
\label{tab: exp_qwen3_vl_30b_a3b_thinking}
\end{table}

\paragraph{Qwen3-VL-8B / 4B / 2B}
We present the evaluation results of Qwen3-VL-2B, Qwen3-VL-4B, and Qwen3-VL-8B in Table \ref{tab: exp_qwen3_vl_2b_4b_8b_nothinking} and Table \ref{tab: exp_qwen3_vl_2b_4b_8b_thinking}. For Qwen3-VL-2B and Qwen3-VL-8B, we compare them with Qwen3-1.7B and Qwen3-8B. For Qwen3-VL-4B, we compare it with Qwen3-4B and Qwen3-4B-2507. Overall, these edge-side models exhibit impressive performance and outperform baselines. These results demonstrate the efficacy of our Strong-to-Weak Distillation approach, making it possible for us to build the lightweight models with remarkably reduced costs and efforts.

\begin{table}[t]
\caption{\textbf{Comparison among Qwen3-VL-2B (Instruct), Qwen3-VL-4B (Instruct), Qwen3-VL-8B (Instruct) and corresponding baselines. }}
\small
\setlength{\tabcolsep}{2.2pt}
\begin{tabular}{lllllllll}
\toprule
    &  \textbf{Benchmark }  & \begin{tabular}[c]{@{}l@{}}\textbf{Qwen3-VL}\\\textbf{2B}\\{\scriptsize Instruct}\end{tabular}
    & \begin{tabular}[c]{@{}l@{}}\textbf{Qwen3-VL}\\\textbf{4B}\\{\scriptsize Instruct}\end{tabular} 
    & \begin{tabular}[c]{@{}l@{}}\textbf{Qwen3-VL}\\\textbf{8B}\\{\scriptsize Instruct}\end{tabular}
    & \begin{tabular}[c]{@{}l@{}}\textbf{Qwen3}\\\textbf{1.7B}\\{\scriptsize Instruct}\end{tabular} 
    & \begin{tabular}[c]{@{}l@{}}\textbf{Qwen3}\\\textbf{4B}\\{\scriptsize Instruct}\end{tabular} 
    & \begin{tabular}[c]{@{}l@{}}\textbf{Qwen3}\\\textbf{8B}\\{\scriptsize Instruct}\end{tabular} 
    & \begin{tabular}[c]{@{}l@{}}\textbf{Qwen3}\\\textbf{4B}\\{\scriptsize Instruct-2507}\end{tabular} \\
\midrule
\multirow{4}{*}{Knowledge}       
                                 & MMLU-Pro                         & 49.0                 & 67.1                 & 71.6                 & 42.3                & 58.0              & 63.4              & 69.6                   \\
                                 & MMLU-Redux                       & 66.5                 & 81.5                 & 84.9                 & 63.6                & 77.3              & 79.5              & 84.2                   \\
                                 & GPQA                             & 42.0                 & 55.9                 & 61.9                 & 34.7                & 41.7              & 39.3              & 62.0                   \\
                                 & SuperGPQA                        & 24.3                 & 40.3                 & 44.5                 & 22.8                & 32.0              & 35.8              & 42.8                   \\
\midrule
\multirow{3}{*}{Reasoning}       
                                 & AIME-25                           & 22.2                 & 46.6                 & 45.9                 & 10.6                & 19.1              & 20.9              & 47.4                   \\
                                 & HMMT-25                           & 10.9                 & 30.7                 & 32.5                 & 6.2                & 12.1              & 11.8              & 31.0                   \\
                                 & LiveBench {\tiny 2024-11-25}                    & 39.5                 & 60.9                 & 62.0                 & 35.6                & 48.4              & 53.5              & 63.0                   \\
\midrule
\multirow{4}{*}{\begin{tabular}[c]{@{}l@{}}Alignment\\Tasks\end{tabular}} & IFEval                           & 68.2                 & 82.3                 & 83.7                 & 67.1                & 81.2              & 83.0              & 83.4                   \\
                                 & Arena-Hard V2 {\scriptsize (winrate)} & 6.4                 & 30.4                 & 46.3                 & 4.1                 & 9.5              & 15.5              & 43.4                   \\
                                 & Creative   Writing v3            & 48.6                 & 72.3                 & 77.0                 & 49.1                & 53.6              & 69.0              & 83.5                   \\
                                 & WritingBench                     & 73.0                 & 82.5                 & 83.1                 & 65.1                & 68.5              & 71.4              & 83.4                   \\
\midrule
\multirow{2}{*}{Coding   \& Agent} & LiveCodeBench v6 & 20.3                 & 37.9                 & 39.3                 & 16.1                & 26.4              & 25.5              & 35.1                   \\
                                 & BFCL-v3                          & 55.4                 & 63.3                 & 66.3                 & 52.2                & 57.6              & 60.2              & 61.9                   \\
\midrule
\multirow{4}{*}{Multilingualism} & MultiIF                          & 43.2                 & 61.5                 & 66.8                 & 43.2                & 61.3              & 69.2              & 69.0                   \\
                                 & MMLU-ProX                        & 38.8                 & 59.4                 & 65.4                 & 33.5                & 49.6              & 58.0              & 61.6                   \\
                                 & INCLUDE                          & 45.8                 & 61.4                 & 67.0                 & 42.6                & 53.8              & 62.5              & 60.1                   \\
                                 & PolyMATH                         & 14.9                 & 28.8                 & 30.4                 & 10.3                & 16.6              & 18.8              & 31.1    \\
\bottomrule
\end{tabular}
\label{tab: exp_qwen3_vl_2b_4b_8b_nothinking}
\end{table}

\begin{table}[t]
\caption{\textbf{Comparison among Qwen3-VL-2B (Thinking), Qwen3-VL-4B (Thinking), Qwen3-VL-8B (Thinking) and corresponding baselines.}}
\small
\setlength{\tabcolsep}{2.2pt}
\begin{tabular}{lllllllll}
\toprule
    &  \textbf{Benchmark }  & \begin{tabular}[c]{@{}l@{}}\textbf{Qwen3-VL}\\\textbf{2B}\\{\scriptsize Thinking}\end{tabular}
    & \begin{tabular}[c]{@{}l@{}}\textbf{Qwen3-VL}\\\textbf{4B}\\{\scriptsize Thinking}\end{tabular} 
    & \begin{tabular}[c]{@{}l@{}}\textbf{Qwen3-VL}\\\textbf{8B}\\{\scriptsize Thinking}\end{tabular}
    & \begin{tabular}[c]{@{}l@{}}\textbf{Qwen3}\\\textbf{1.7B}\\{\scriptsize Thinking}\end{tabular} 
    & \begin{tabular}[c]{@{}l@{}}\textbf{Qwen3}\\\textbf{4B}\\{\scriptsize Thinking}\end{tabular} 
    & \begin{tabular}[c]{@{}l@{}}\textbf{Qwen3}\\\textbf{8B}\\{\scriptsize Thinking}\end{tabular} 
    & \begin{tabular}[c]{@{}l@{}}\textbf{Qwen3}\\\textbf{4B}\\{\scriptsize Thinking-2507}\end{tabular} \\
\midrule
\multirow{4}{*}{Knowledge}      
                                 & MMLU-Pro                         & 62.3                 & 73.6                 & 77.3                 & 58.1                & 70.4              & 74.6              & 74.0                   \\
                                 & MMLU-Redux                       & 76.9                 & 86.0                 & 88.8                 & 73.9                & 83.7              & 87.5              & 86.1                   \\
                                 & GPQA                             & 49.5                 & 64.1                 & 69.9                 & 27.9                & 55.9              & 62.0              & 65.8                   \\
                                 & SuperGPQA                        & 34.6                 & 46.8                 & 51.2                 & 31.2                & 42.7              & 47.6              & 47.8                   \\
\midrule
\multirow{4}{*}{Reasoning}       
                                 & AIME-25                           & 39.0                 & 74.5                 & 80.3                 & 36.8                & 65.6              & 67.3              & 81.3                   \\
                                 & HMMT-25                           & 22.8                 & 53.1                 & 60.6                 & 24.3                & 42.1              & 43.2              & 55.5                   \\
                                 & LiveBench {\tiny 2024-11-25}                    & 50.1                 & 68.4                 & 69.8                 & 51.1                & 63.6              & 67.1              & 71.8                   \\
\midrule
\multirow{4}{*}{\begin{tabular}[c]{@{}l@{}}Alignment\\Tasks\end{tabular}} & IFEval                           & 75.1                 & 82.6                 & 83.2                 & 72.5                & 81.9              & 85.0              & 87.4                   \\
                                 & Arena-Hard V2 {\scriptsize (winrate)} & 12.0                 & 36.8                 & 51.1                 & 4.7                 & 13.7              & 29.1              & 34.9                   \\
                                 & Creative   Writing v3            & 55.6                 & 76.1                 & 82.4                 & 50.6                & 61.1              & 78.5              & 75.6                   \\
                                 & WritingBench                     & 77.9                 & 84.0                 & 85.5                 & 68.9                & 73.5              & 75.0              & 83.3                   \\
\midrule
\multirow{2}{*}{Coding   \& Agent} & LiveCodeBench v6 & 29.3                 & 51.3                 & 58.6                 & 31.3                & 48.4              & 51.0              & 55.2                   \\
                                 & BFCL-v3                          & 57.2                 & 67.3                 & 63.0                 & 56.6                & 65.9              & 68.1              & 71.2                   \\
\midrule
\multirow{4}{*}{Multilingualism} & MultiIF                          & 58.9                 & 73.6                 & 75.1                 & 51.2                & 66.3              & 71.2              & 77.3                   \\
                                 & MMLU-ProX                        & 55.1                 & 65.0                 & 70.7                 & 50.4                & 61.0              & 68.1              & 64.2                   \\
                                 & INCLUDE                          & 53.3                 & 64.6                 & 69.5                 & 51.8                & 61.8              & 67.8              & 64.4                   \\
                                 & PolyMATH                         & 28.0                 & 44.6                 & 47.5                 & 25.2                & 40.0              & 42.7              & 46.2    \\
\bottomrule
\end{tabular}
\label{tab: exp_qwen3_vl_2b_4b_8b_thinking}
\end{table}

\subsection{Ablation Study}
\subsubsection{Vision Encoder}
 We conduct comparative experiments against the original SigLIP-2. As shown in~\cref{tab:Qwen3_ViT_results}, in zero-shot evaluation at the CLIP pretraining stage, Qwen3-ViT maintains competitive performance on standard benchmarks while achieving substantial gains on OmniBench, our in-house holistic evaluation suite designed to assess world knowledge integration under diverse and challenging conditions. Furthermore,  when integrated with the same 1.7B Qwen3 language model and trained for 1.5T tokens, Qwen3-ViT consistently outperforms the SigLIP-2-based baseline across multiple key tasks and remains significantly ahead on OmniBench, demonstrating its superiority and effectiveness as a stronger visual backbone.

\begin{table}[h]
\centering
\caption{\textbf{Ablation on Qwen3-ViT.} We compare the performance metrics of Qwen3-ViT and SigLIP-2 during the CLIP pre-training stage, and further evaluate their downstream performance in the vision-language modeling (VLM) stage when paired with the same 1.7B Qwen3 language model.}
\label{tab:Qwen3_ViT_results}
\setlength{\tabcolsep}{3.0pt}
\resizebox{\linewidth}{!}{
\begin{NiceTabular}{l|ccccccc|ccccc}
\toprule
\multirow{2}{*}{ViT} & \multicolumn{7}{c|}{Clip Bench}                                                                                                                                                                                                  & \multicolumn{5}{c}{VLM Bench}                                                                           \\
                     & \multicolumn{1}{l}{ImageNet-1K} & \multicolumn{1}{l}{ImageNet-V2} & \multicolumn{1}{l}{ImageNet-A} & \multicolumn{1}{l}{ImageNet-R} & \multicolumn{1}{l}{ImageNet-S} & \multicolumn{1}{l}{ObjectNet} & \multicolumn{1}{l}{Omni} & \multicolumn{1}{l}{OCRB} & \multicolumn{1}{l}{AI2D} & \multicolumn{1}{l}{RLWDQA} & \multicolumn{1}{l}{InfoVQA} & \multicolumn{1}{l}{Omni}  \\ 
\midrule
SigLIP-2              & 84.2                            & 78.6                            & 87.0                           & 96.1                           & 76.2                           & 79.9                          & 36.9                      & 77.2                     & 74.1                     & 58.7                       & 65.3                        & 50.1                      \\
Qwen3-ViT            & 84.6                            & 78.8                            & 87.1                           & 95.7                           & 74.5                           & 81.0                          & 45.5                      & 78.7                     & 76.2                     & 66.1                       & 67.0                        & 53.0                      \\
\bottomrule
\end{NiceTabular}
}
\end{table}

\subsubsection{DeepStack}
We conduct an ablation study to verify the effectiveness of the DeepStack mechanism. As demonstrated in~\cref{tab:deepstack_results}, the model equipped with DeepStack achieved an overall performance gain across various benchmarks, strongly affirming its effectiveness. This gain is attributed to DeepStack's ability to integrate rich visual information, which effectively boosts the capability in fine-grained visual understanding, such as on the InfoVQA and DocVQA benchmarks.

\begin{table}[!h]
\centering
\caption{\textbf{Ablation on DeepStack.} We conduct the ablation study on the DeepStack using an internal 15B-A2B LLM, with all experiments pretrained on 200 billion tokens. We directly evaluate these pretrained models on the validation sets, without any post-training.}
\label{tab:deepstack_results}
\setlength{\tabcolsep}{3.0pt}
\resizebox{\linewidth}{!}{
\begin{tabular}{@{}lcccccccccccc@{}}
\toprule
\textbf{Method} & AVG & AI2D & OCRB & TVQA & InfoVQA & ChartQA & DocVQA & MMMU & MMStar & RLWDQA & MMB$_{EN}$ & MMB$_{CN}$\\
\midrule
Baseline & 74.7 & 81.8 & 81.0 & 80.6 & 71.9 & 81.5 & 89.5 & 52.9 & 55.5 & 67.7 & 81.0 & 78.1\\
DeepStack & 76.0 & 83.2 & 83.6 & 80.5 & 74.2 & 83.3 & 91.1 & 54.1 & 57.7 & 68.1 & 81.2 & 78.5\\

\bottomrule
\end{tabular}
}
\end{table}

\begin{figure}
    \centering
    \includegraphics[width=1.0\linewidth]{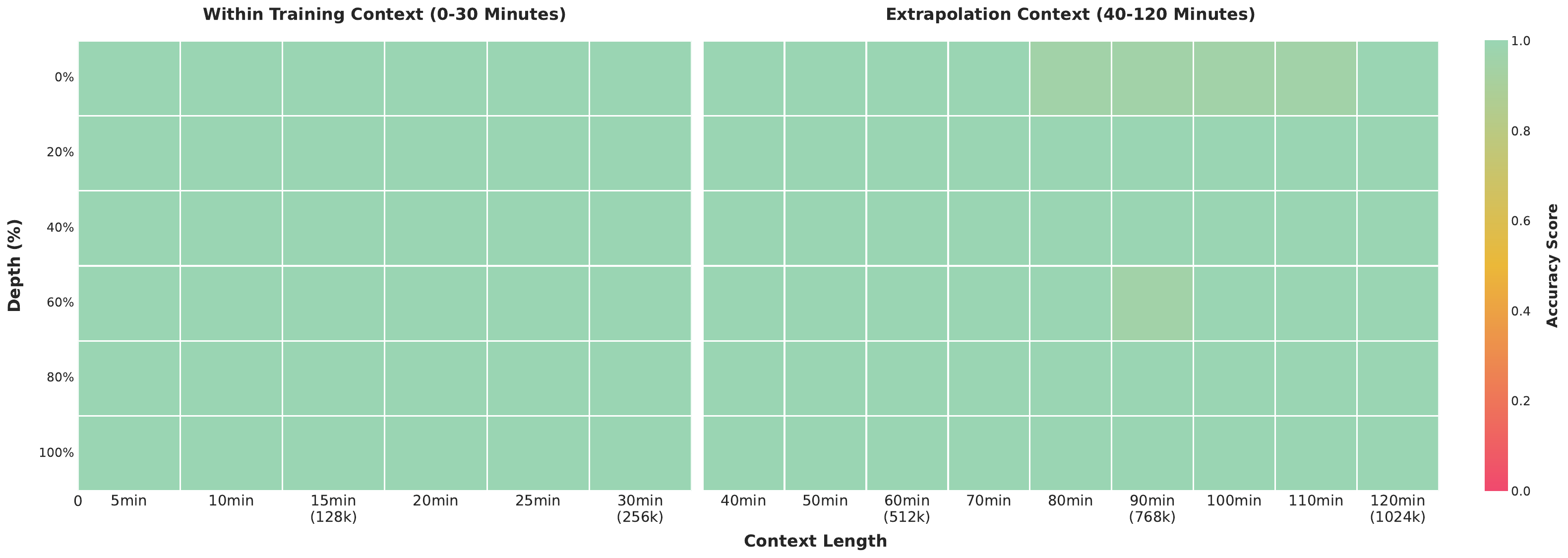}
    \caption{Needle-in-a-Haystack performance heatmap for Qwen3-VL-235B-A22B-Instruct across varying video durations and needle positions. Each cell shows accuracy (\%) for locating and answering questions about the inserted ``needle'' frame.}
    \label{fig:niah}
\end{figure}

\subsubsection{Needle-in-a-Haystack}
To evaluate the model's capability in processing long-context inputs, we construct a video ``Needle-in-a-Haystack'' evaluation on Qwen3-VL-235B-A22B-Instruct. In this task, a semantically salient ``needle'' frame—containing critical visual evidence—is inserted at varying temporal positions within a long video. The model is then tasked with accurately locating the target frame from the long video and answering the corresponding question. During evaluation, videos are uniformly sampled at 1 FPS, and frame resolution is dynamically adjusted to maintain a constant visual token budget.

As shown in~\cref{fig:niah}, the model achieves a perfect 100\% accuracy on videos up to 30 minutes in duration—corresponding to a context length of 256K tokens. Remarkably, even when extrapolating to sequences of up to 1M tokens (approximately 2 hours of video) via YaRN-based positional extension, the model retains a high accuracy of 99.5\%. These results strongly demonstrate the model's powerful long-sequence modeling capabilities.

\section{Conclusion}
In this work, we present Qwen3-VL, a state-of-the-art series of vision–language foundation models that advances the frontier of multimodal understanding and generation. By integrating high-quality multimodal data iteration and architectural innovations—such as enhanced interleaved-MRoPE, DeepStack vision-language alignment, and text-based temporal grounding—Qwen3-VL achieves unprecedented performance across a broad spectrum of multimodal benchmarks while maintaining strong pure-text capabilities. Its native support for 256K-token interleaved sequences enables robust reasoning over long, complex documents, image sequences, and videos, making it uniquely suited for real-world applications demanding high-fidelity cross-modal comprehension. The availability of both dense and Mixture-of-Experts variants ensures flexible deployment across diverse latency and quality requirements, and our post-training strategy—including non-thinking and thinking modes.

Looking forward, we envision Qwen3-VL as a foundational engine for embodied AI agents capable of seamlessly bridging the digital and physical worlds. Such agents will not only perceive and reason over rich multimodal inputs but also execute decisive, context-aware actions in dynamic environments—interacting with users, manipulating digital interfaces, and guiding robotic systems through grounded, multimodal decision-making. Future work will focus on extending Qwen3-VL’s capabilities toward interactive perception, tool-augmented reasoning, and real-time multimodal control, with the ultimate goal of enabling AI systems that learn, adapt, and collaborate alongside humans in both virtual and physical domains. Additionally, we are actively exploring unified understanding-generation architectures, leveraging visual generation capabilities to elevate overall intelligence further. By openly releasing the entire model family under the Apache 2.0 license, we aim to catalyze community-driven innovation toward the vision of truly integrated, multimodal AI agents.

\section{Contributions and Acknowledgments}

All contributors of Qwen3-VL are listed in alphabetical order by their last names.

\textbf{Core Contributors:} Shuai Bai, Yuxuan Cai, Ruizhe Chen, Keqin Chen, Xionghui Chen, Zesen Cheng, Lianghao Deng, Wei Ding, Chang Gao, Chunjiang Ge, Wenbin Ge, Zhifang Guo, Qidong Huang, Jie Huang, Fei Huang, Binyuan Hui, Shutong Jiang, Zhaohai Li, Mingsheng Li, Mei Li, Kaixin Li, Zicheng Lin, Junyang Lin, Xuejing Liu, Jiawei Liu, Chenglong Liu, Yang Liu, Dayiheng Liu, Shixuan Liu, Dunjie Lu, Ruilin Luo, Chenxu Lv, Rui Men, Lingchen Meng, Xuancheng Ren, Xingzhang Ren, Sibo Song, Yuchong Sun, Jun Tang, Jianhong Tu, Jianqiang Wan, Peng Wang, Pengfei Wang, Qiuyue Wang, Yuxuan Wang, Tianbao Xie, Yiheng Xu, Haiyang Xu, Jin Xu, Zhibo Yang, Mingkun Yang, Jianxin Yang, An Yang, Bowen Yu, Fei Zhang, Hang Zhang, Xi Zhang, Bo Zheng, Humen Zhong, Jingren Zhou, Fan Zhou, Jing Zhou, Yuanzhi Zhu, Ke Zhu

\textbf{Contributors:} Yizhong Cao, Bei Chen, Chen Cheng, Yunfei Chu, Zeyu Cui, Kai Dang, Xiaodong Deng, Yang Fan, Rongyao Fang, Tongkun Guan, Jinzheng He, Miao Hong, Songtao Jiang, Zheng Li, Xiaochuan Li, Junrong Lin, Yuqiong Liu, Yantao Liu, Na Ni, Xinyao Niu, Yatian Pang, Zihan Qiu, Tianhao Shen, Tianyi Tang, Yu Wan, Jinxi Wei, Chenfei Wu, Buxiao Wu, Xiao Xu, Mingfeng Xue, Ming Yan, Yuhuan Yang, Jiaxi Yang, Kexin Yang, Le Yu, Hao Yu, Jianke Zhang, Jianwei Zhang, Yichang Zhang, Zhenru Zhang, Siqi Zhang, Peiyang Zhang, Beichen Zhang, Hongbo Zhao, Xianwei Zhuang

\textbf{Acknowledgments:} We gratefully acknowledge the unwavering support provided by the teams led by Zulong Chen, Bing Deng, Feiyu Gao, Guanjun Jiang, Yue Liu, Hangdi Xing and Daijun Yu.
\bibliography{colm2024_conference}

@article{sapo,
title={Soft Adaptive Policy Optimization},
author={Gao, Chang and Zheng, Chujie and Chen, Xiong-Hui and Dang, Kai and Liu, Shixuan and Yu, Bowen and Yang, An and Bai, Shuai and Zhou, Jingren and Lin, Junyang},
journal={arXiv preprint arXiv:2511.20347},
year={2025}
}

@article{ma2024mmlongbench,
  title={Mmlongbench-doc: Benchmarking long-context document understanding with visualizations},
  author={Ma, Yubo and Zang, Yuhang and Chen, Liangyu and Chen, Meiqi and Jiao, Yizhu and Li, Xinze and Lu, Xinyuan and Liu, Ziyu and Ma, Yan and Dong, Xiaoyi and others},
  journal={Advances in Neural Information Processing Systems},
  volume={37},
  pages={95963--96010},
  year={2024}
}

@article{fu2024video,
  title={Video-MME: The First-Ever Comprehensive Evaluation Benchmark of Multi-modal LLMs in Video Analysis},
  author={Fu, Chaoyou and Dai, Yuhan and Luo, Yondong and Li, Lei and Ren, Shuhuai and Zhang, Renrui and Wang, Zihan and Zhou, Chenyu and Shen, Yunhang and Zhang, Mengdan and others},
  journal={arXiv:2405.21075},
  year={2024}
}

@inproceedings{li2024mvbench,
  title={Mvbench: A comprehensive multi-modal video understanding benchmark},
  author={Li, Kunchang and Wang, Yali and He, Yinan and Li, Yizhuo and Wang, Yi and Liu, Yi and Wang, Zun and Xu, Jilan and Chen, Guo and Luo, Ping and others},
  booktitle={CVPR},
  year={2024}
}

@article{MMBench,
    author  = {Yuan Liu and Haodong Duan and Yuanhan Zhang, Bo Li and Songyang Zhang and Wangbo Zhao and Yike Yuan and Jiaqi Wang and Conghui He and Ziwei Liu and Kai Chen and Dahua Lin},
    journal = {arXiv:2307.06281},
    title   = {MMBench: Is Your Multi-modal Model an All-around Player?},
    year    = {2023},
}

@article{chen2024we,
  title={Are We on the Right Way for Evaluating Large Vision-Language Models?},
  author={Chen, Lin and Li, Jinsong and Dong, Xiaoyi and Zhang, Pan and Zang, Yuhang and Chen, Zehui and Duan, Haodong and Wang, Jiaqi and Qiao, Yu and Lin, Dahua and others},
  journal={arXiv:2403.20330},
  year={2024}
}

@article{li2024omnicorpus,
  title={OmniCorpus: An Unified Multimodal Corpus of 10 Billion-Level Images Interleaved with Text},
  author={Li, Qingyun and Chen, Zhe and Wang, Weiyun and Wang, Wenhai and Ye, Shenglong and Jin, Zhenjiang and Chen, Guanzhou and He, Yinan and Gao, Zhangwei and Cui, Erfei and others},
  journal={arXiv preprint arXiv:2406.08418},
  year={2024}
}

@article{comanici2025gemini,
  title={Gemini 2.5: Pushing the frontier with advanced reasoning, multimodality, long context, and next generation agentic capabilities},
  author={Comanici, Gheorghe and Bieber, Eric and Schaekermann, Mike and Pasupat, Ice and Sachdeva, Noveen and Dhillon, Inderjit and Blistein, Marcel and Ram, Ori and Zhang, Dan and Rosen, Evan and others},
  journal={arXiv preprint arXiv:2507.06261},
  year={2025}
}

@inproceedings{lin2014microsoft,
  title={Microsoft coco: Common objects in context},
  author={Lin, Tsung-Yi and Maire, Michael and Belongie, Serge and Hays, James and Perona, Pietro and Ramanan, Deva and Doll{\'a}r, Piotr and Zitnick, C Lawrence},
  booktitle={ECCV},
  year={2014},
}

@inproceedings{vit,
  title={An Image is Worth 16x16 Words: Transformers for Image Recognition at Scale},
  author={Dosovitskiy, Alexey and Beyer, Lucas and Kolesnikov, Alexander and Weissenborn, Dirk and Zhai, Xiaohua and Unterthiner, Thomas and  Dehghani, Mostafa and Minderer, Matthias and Heigold, Georg and Gelly, Sylvain and Uszkoreit, Jakob and Houlsby, Neil},
  booktitle={ICLR},
  year={2021}
}

@inproceedings{docvqa,
  title={Docvqa: A dataset for vqa on document images},
  author={Mathew, Minesh and Karatzas, Dimosthenis and Jawahar, CV},
  booktitle={WACV},
  year={2021}
}

@inproceedings{refcoco,
  title={Referitgame: Referring to objects in photographs of natural scenes},
  author={Kazemzadeh, Sahar and Ordonez, Vicente and Matten, Mark and Berg, Tamara},
  booktitle={EMNLP},
  year={2014}
}

@inproceedings{refcocog,
  title={Generation and comprehension of unambiguous object descriptions},
  author={Mao, Junhua and Huang, Jonathan and Toshev, Alexander and Camburu, Oana and Yuille, Alan L and Murphy, Kevin},
  booktitle={CVPR},
  year={2016}
}

@article{masry2022chartqa,
  title={ChartQA: A benchmark for question answering about charts with visual and logical reasoning},
  author={Masry, Ahmed and Long, Do Xuan and Tan, Jia Qing and Joty, Shafiq and Hoque, Enamul},
  journal={arXiv:2203.10244},
  year={2022}
}

@article{Qwen2-VL,
  title={Qwen2-VL: Enhancing Vision-Language Model's Perception of the World at Any Resolution},
  author={Wang, Peng and Bai, Shuai and Tan, Sinan and Wang, Shijie and Fan, Zhihao and Bai, Jinze and Chen, Keqin and Liu, Xuejing and Wang, Jialin and Ge, Wenbin and Fan, Yang and Dang, Kai and Du, Mengfei and Ren, Xuancheng and Men, Rui and Liu, Dayiheng and Zhou, Chang and Zhou, Jingren and Lin, Junyang},
  journal={arXiv:2409.12191},
  year={2024}
}

@misc{qwen2.5vl,
      title={Qwen2.5-VL Technical Report}, 
      author={Shuai Bai and Keqin Chen and Xuejing Liu and Jialin Wang and Wenbin Ge and Sibo Song and Kai Dang and Peng Wang and Shijie Wang and Jun Tang and Humen Zhong and Yuanzhi Zhu and Mingkun Yang and Zhaohai Li and Jianqiang Wan and Pengfei Wang and Wei Ding and Zheren Fu and Yiheng Xu and Jiabo Ye and Xi Zhang and Tianbao Xie and Zesen Cheng and Hang Zhang and Zhibo Yang and Haiyang Xu and Junyang Lin},
      journal={arXiv:2502.13923},
      year={2025},
}

@misc{interleave-mrope,
      title={Revisiting Multimodal Positional Encoding in Vision-Language Models}, 
      author={Jie Huang and Xuejing Liu and Sibo Song and Ruibing Hou and Hong Chang and Junyang Lin and Shuai Bai},
      journal={arXiv:2510.23095},
      year={2025},
}

@misc{gpt5,
    title = {GPT-5 System Card},
    url = {https://cdn.openai.com/gpt-5-system-card.pdf},
    author = {OpenAI.},
    year = {2025}
}

@misc{opus4_1,
    title = {Claude Opus 4.1},
    url = {https://www.anthropic.com/news/claude-opus-4-1},
    author = {Anthropic.},
    year = {2025}
}

@article{gemini,
  title={Gemini: A family of highly capable multimodal models},
  author={Team, Gemini and Anil, Rohan and Borgeaud, Sebastian and Wu, Yonghui and Alayrac, Jean-Baptiste and Yu, Jiahui and Soricut, Radu and Schalkwyk, Johan and Dai, Andrew M and Hauth, Anja and others},
  journal={arXiv:2312.11805},
  year={2023}
}

@article{grounding_dino,
  title={Grounding DINO: Marrying DINO with Grounded Pre-Training for Open-Set Object Detection},
  author={Shilong Liu and Zhaoyang Zeng and Tianhe Ren and Feng Li and Hao Zhang and Jie Yang and Chun-yue Li and Jianwei Yang and Hang Su and Jun-Juan Zhu and Lei Zhang},
  journal={arXiv:2303.05499},
  year={2023},
}

@misc{qwen3,
      title={Qwen3 Technical Report}, 
      author={An Yang and Anfeng Li and Baosong Yang and Beichen Zhang and Binyuan Hui and Bo Zheng and Bowen Yu and others},
      journal = {arXiv:2505.09388},
      year={2025},
}

@article{mathvision,
  title={Measuring Multimodal Mathematical Reasoning with MATH-Vision Dataset}, 
  author={Ke Wang and Junting Pan and Weikang Shi and Zimu Lu and Mingjie Zhan and Hongsheng Li},
  journal={arXiv:2402.14804},
  year={2024}
}

@article{wang2024muirbench,
  title={MuirBench: A Comprehensive Benchmark for Robust Multi-image Understanding},
  author={Wang, Fei and Fu, Xingyu and Huang, James Y and Li, Zekun and Liu, Qin and Liu, Xiaogeng and Ma, Mingyu Derek and Xu, Nan and Zhou, Wenxuan and Zhang, Kai and others},
  journal={arXiv preprint arXiv:2406.09411},
  year={2024}
}

@inproceedings{fu2024blink,
  title={Blink: Multimodal large language models can see but not perceive},
  author={Fu, Xingyu and Hu, Yushi and Li, Bangzheng and Feng, Yu and Wang, Haoyu and Lin, Xudong and Roth, Dan and Smith, Noah A and Ma, Wei-Chiu and Krishna, Ranjay},
  booktitle={European Conference on Computer Vision},
  pages={148--166},
  year={2024},
  organization={Springer}
}

@article{rawles2024androidworld,
  title={AndroidWorld: A dynamic benchmarking environment for autonomous agents},
  author={Rawles, Christopher and Clinckemaillie, Sarah and Chang, Yifan and Waltz, Jonathan and Lau, Gabrielle and Fair, Marybeth and Li, Alice and Bishop, William and Li, Wei and Campbell-Ajala, Folawiyo and others},
  journal={arXiv:2405.14573},
  year={2024}
}

@article{deitke2024molmo,
  title={Molmo and pixmo: Open weights and open data for state-of-the-art multimodal models},
  author={Deitke, Matt and Clark, Christopher and Lee, Sangho and Tripathi, Rohun and Yang, Yue and Park, Jae Sung and Salehi, Mohammadreza and Muennighoff, Niklas and Lo, Kyle and Soldaini, Luca and others},
  journal={arXiv preprint arXiv:2409.17146},
  year={2024}
}

@article{hu2025video,
  title={Video-MMMU: Evaluating Knowledge Acquisition from Multi-Discipline Professional Videos},
  author={Hu, Kairui and Wu, Penghao and Pu, Fanyi and Xiao, Wang and Zhang, Yuanhan and Yue, Xiang and Li, Bo and Liu, Ziwei},
  journal={arXiv preprint arXiv:2501.13826},
  year={2025}
}

@article{zhou2024mlvu,
  title={MLVU: A Comprehensive Benchmark for Multi-Task Long Video Understanding},
  author={Zhou, Junjie and Shu, Yan and Zhao, Bo and Wu, Boya and Xiao, Shitao and Yang, Xi and Xiong, Yongping and Zhang, Bo and Huang, Tiejun and Liu, Zheng},
  journal={arXiv preprint arXiv:2406.04264},
  year={2024}
}

@article{wang2024lvbench,
  title={Lvbench: An extreme long video understanding benchmark},
  author={Wang, Weihan and He, Zehai and Hong, Wenyi and Cheng, Yean and Zhang, Xiaohan and Qi, Ji and Gu, Xiaotao and Huang, Shiyu and Xu, Bin and Dong, Yuxiao and others},
  journal={arXiv preprint arXiv:2406.08035},
  year={2024}
}

@misc{zhao2025mmvu,
      title={MMVU: Measuring Expert-Level Multi-Discipline Video Understanding}, 
      author={Yilun Zhao and Lujing Xie and Haowei Zhang and Guo Gan and Yitao Long and Zhiyuan Hu and Tongyan Hu and Weiyuan Chen and Chuhan Li and Junyang Song and Zhijian Xu and Chengye Wang and others},
      year={2025},
      eprint={2501.12380},
      archivePrefix={arXiv},
      primaryClass={cs.CV},
      url={https://arxiv.org/abs/2501.12380}, 
}

@inproceedings{gao2017tall,
  title={Tall: Temporal activity localization via language query},
  author={Gao, Jiyang and Sun, Chen and Yang, Zhenheng and Nevatia, Ram},
  booktitle={Proceedings of the IEEE international conference on computer vision},
  pages={5267--5275},
  year={2017}
}

@article{Kembhavi2016AI2D,
  title={A Diagram is Worth a Dozen Images},
  author={Aniruddha Kembhavi and Michael Salvato and Eric Kolve and Minjoon Seo and Hannaneh Hajishirzi and Ali Farhadi},
  journal={ArXiv},
  year={2016},
  volume={abs/1603.07396}
}

@article{Mathew2021InfographicVQA,
  title={InfographicVQA},
  author={Minesh Mathew and Viraj Bagal and Rub{\`e}n P{\'e}rez Tito and Dimosthenis Karatzas and Ernest Valveny and C.V. Jawahar},
  journal={2022 IEEE/CVF Winter Conference on Applications of Computer Vision (WACV)},
  year={2021},
  pages={2582-2591}
}

@article{Liu_2024_OCRBench,
    title={OCRBench: on the hidden mystery of OCR in large multimodal models},
    volume={67},
    ISSN={1869-1919},
    url={http://dx.doi.org/10.1007/s11432-024-4235-6},
    DOI={10.1007/s11432-024-4235-6},
    number={12},
    journal={Science China Information Sciences},
    publisher={Springer Science and Business Media LLC},
    author={Liu, Yuliang and Li, Zhang and Huang, Mingxin and Yang, Biao and Yu, Wenwen and Li, Chunyuan and Yin, Xu-Cheng and Liu, Cheng-Lin and Jin, Lianwen and Bai, Xiang},
    year={2024},
    month=dec }

@misc{fu2024ocrbenchv2improvedbenchmark,
    title={OCRBench v2: An Improved Benchmark for Evaluating Large Multimodal Models on Visual Text Localization and Reasoning}, 
    author={Ling Fu and Biao Yang and Zhebin Kuang and Jiajun Song and Yuzhe Li and Linghao Zhu and Qidi Luo and Xinyu Wang and Hao Lu and Mingxin Huang and Zhang Li and Guozhi Tang and Bin Shan and Chunhui Lin and Qi Liu and Binghong Wu and Hao Feng and Hao Liu and Can Huang and Jingqun Tang and Wei Chen and Lianwen Jin and Yuliang Liu and Xiang Bai},
    year={2024},
    eprint={2501.00321},
    archivePrefix={arXiv},
    primaryClass={cs.CV},
    url={https://arxiv.org/abs/2501.00321}, 
}

@misc{yang2024ccocrcomprehensivechallengingocr,
      title={CC-OCR: A Comprehensive and Challenging OCR Benchmark for Evaluating Large Multimodal Models in Literacy}, 
      author={Zhibo Yang and Jun Tang and Zhaohai Li and Pengfei Wang and Jianqiang Wan and Humen Zhong and Xuejing Liu and Mingkun Yang and Peng Wang and Shuai Bai and LianWen Jin and Junyang Lin},
      year={2024},
      eprint={2412.02210},
      archivePrefix={arXiv},
      primaryClass={cs.CV},
      url={https://arxiv.org/abs/2412.02210}, 
}

@misc{ouyang2024omnidocbenchbenchmarkingdiversepdf,
      title={OmniDocBench: Benchmarking Diverse PDF Document Parsing with Comprehensive Annotations}, 
      author={Linke Ouyang and Yuan Qu and Hongbin Zhou and Jiawei Zhu and Rui Zhang and Qunshu Lin and Bin Wang and Zhiyuan Zhao and Man Jiang and Xiaomeng Zhao and Jin Shi and Fan Wu and Pei Chu and Minghao Liu and Zhenxiang Li and Chao Xu and Bo Zhang and Botian Shi and Zhongying Tu and Conghui He},
      year={2024},
      eprint={2412.07626},
      archivePrefix={arXiv},
      primaryClass={cs.CV},
      url={https://arxiv.org/abs/2412.07626}, 
}

@article{wang2024charxiv,
  title={CharXiv: Charting Gaps in Realistic Chart Understanding in Multimodal LLMs},
  author={Wang, Zirui and Xia, Mengzhou and He, Luxi and Chen, Howard and Liu, Yitao and Zhu, Richard and Liang, Kaiqu and Wu, Xindi and Liu, Haotian and Malladi, Sadhika and Chevalier, Alexis and Arora, Sanjeev and Chen, Danqi},
  journal={arXiv preprint arXiv:2406.18521},
  year={2024}
}

@inproceedings{li2022grounded,
  title={Grounded language-image pre-training},
  author={Li, Liunian Harold and Zhang, Pengchuan and Zhang, Haotian and Yang, Jianwei and Li, Chunyuan and Zhong, Yiwu and Wang, Lijuan and Yuan, Lu and Zhang, Lei and Hwang, Jenq-Neng and others},
  booktitle={Proceedings of the IEEE/CVF Conference on Computer Vision and Pattern Recognition},
  pages={10965--10975},
  year={2022}
}

@inproceedings{paiss2023teaching,
  title={Teaching clip to count to ten},
  author={Paiss, Roni and Ephrat, Ariel and Tov, Omer and Zada, Shiran and Mosseri, Inbar and Irani, Michal and Dekel, Tali},
  booktitle={Proceedings of the IEEE/CVF International Conference on Computer Vision},
  pages={3170--3180},
  year={2023}
}

@article{cheng2024seeclick,
  title={Seeclick: Harnessing gui grounding for advanced visual gui agents},
  author={Cheng, Kanzhi and Sun, Qiushi and Chu, Yougang and Xu, Fangzhi and Li, Yantao and Zhang, Jianbing and Wu, Zhiyong},
  journal={arXiv preprint arXiv:2401.10935},
  year={2024}
}

@misc{screenspotpro,
  author    = {Kaixin Li and Ziyang Meng and Hongzhan Lin and Ziyang Luo and Yuchen Tian and Jing Ma and Zhiyong Huang and Tat-Seng Chua},
  title     = {ScreenSpot-Pro: GUI Grounding for Professional High-Resolution Computer Use},
  year      = {2025},
  note      = {Preprint},
  url       = {https://likaixin2000.github.io/papers/ScreenSpot_Pro.pdf},
}

@article{xie2025osworld,
  title={Osworld: Benchmarking multimodal agents for open-ended tasks in real computer environments},
  author={Xie, Tianbao and Zhang, Danyang and Chen, Jixuan and Li, Xiaochuan and Zhao, Siheng and Cao, Ruisheng and others},
  journal={Advances in Neural Information Processing Systems},
  volume={37},
  pages={52040--52094},
  year={2025}
}

@misc{realworldqa2024,
  title        = {RealWorldQA: A Benchmark for Real-World Spatial Understanding},
  author       = {{xAI}},
  year         = {2024},
  howpublished = {\url{https://huggingface.co/datasets/xai-org/RealworldQA}},
  note         = {Accessed: 2025-04-26}
}

@misc{claude,
  title = {Introducing {Claude}},
  author = {Anthropic},
  institution = {Anthropic},
  url = {https://www.anthropic.com/index/introducing-claude},
  year={2023}
}

@inproceedings{math,
  author       = {Dan Hendrycks and
                  Collin Burns and
                  Saurav Kadavath and
                  Akul Arora and
                  Steven Basart and
                  Eric Tang and
                  Dawn Song and
                  Jacob Steinhardt},
  title        = {Measuring Mathematical Problem Solving With the {MATH} Dataset},
  booktitle    = {NeurIPS Datasets and Benchmarks},
  year         = {2021}
}

@article{mmlupro,
  author       = {Yubo Wang and
                  Xueguang Ma and
                  Ge Zhang and
                  Yuansheng Ni and
                  Abhranil Chandra and
                  Shiguang Guo and
                  Weiming Ren and
                  Aaran Arulraj and
                  Xuan He and
                  Ziyan Jiang and
                  Tianle Li and
                  others},
  title        = {{MMLU-Pro}: {A} More Robust and Challenging Multi-Task Language Understanding
                  Benchmark},
  journal      = {CoRR},
  volume       = {abs/2406.01574},
  year         = {2024}
}

@inproceedings{meng2024deepstack,
  title={Deepstack: Deeply stacking visual tokens is surprisingly simple and effective for lmms},
  author={Meng, Lingchen and Yang, Jianwei and Tian, Rui and Dai, Xiyang and Wu, Zuxuan and Gao, Jianfeng and Jiang, Yu-Gang},
  booktitle={Advances in Neural Information Processing Systems},
  volume={37},
  pages={23464--23487},
  year={2024}
}

@article{tschannen2025siglip,
  title={Siglip 2: Multilingual vision-language encoders with improved semantic understanding, localization, and dense features},
  author={Tschannen, Michael and Gritsenko, Alexey and Wang, Xiao and Naeem, Muhammad Ferjad and Alabdulmohsin, Ibrahim and Parthasarathy, Nikhil and Evans, Talfan and Beyer, Lucas and Xia, Ye and Mustafa, Basil and others},
  journal={arXiv preprint arXiv:2502.14786},
  year={2025}
}

@article{chen2025comp,
  title={Comp: Continual multimodal pre-training for vision foundation models},
  author={Chen, Yitong and Meng, Lingchen and Peng, Wujian and Wu, Zuxuan and Jiang, Yu-Gang},
  journal={arXiv preprint arXiv:2503.18931},
  year={2025}
}

@article{mm_mt_bench,
  title={Pixtral 12B},
  author={Agrawal, Pravesh and Antoniak, Szymon and Hanna, Emma Bou and Bout, Baptiste and Chaplot, Devendra and Chudnovsky, Jessica and Costa, Diogo and De Monicault, Baudouin and Garg, Saurabh and Gervet, Theophile and others},
  journal={arXiv preprint arXiv:2410.07073},
  year={2024}
}

@misc{hallusion_bench,
      title={HallusionBench: An Advanced Diagnostic Suite for Entangled Language Hallucination \& Visual Illusion in Large Vision-Language Models}, 
      author={Tianrui Guan and Fuxiao Liu and Xiyang Wu and Ruiqi Xian and Zongxia Li and Xiaoyu Liu and Xijun Wang and Lichang Chen and Furong Huang and Yaser Yacoob and Dinesh Manocha and Tianyi Zhou},
      year={2023},
      eprint={2310.14566},
      archivePrefix={arXiv},
      primaryClass={cs.CV}
}

@article{mia-bench,
  title={Mia-bench: Towards better instruction following evaluation of multimodal llms},
  author={Qian, Yusu and Ye, Hanrong and Fauconnier, Jean-Philippe and Grasch, Peter and Yang, Yinfei and Gan, Zhe},
  journal={arXiv preprint arXiv:2407.01509},
  year={2024}
}

@article{zheng2025deepeyes,
  title={DeepEyes: Incentivizing" Thinking with Images" via Reinforcement Learning},
  author={Zheng, Ziwei and Yang, Michael and Hong, Jack and Zhao, Chenxiao and Xu, Guohai and Yang, Le and Shen, Chao and Yu, Xing},
  journal={arXiv preprint arXiv:2505.14362},
  year={2025}
}

@article{wu2025mmsearch,
  title={MMSearch-R1: Incentivizing LMMs to Search},
  author={Wu, Jinming and Deng, Zihao and Li, Wei and Liu, Yiding and You, Bo and Li, Bo and Ma, Zejun and Liu, Ziwei},
  journal={arXiv preprint arXiv:2506.20670},
  year={2025}
}

@article{jin2025search,
  title={Search-r1: Training llms to reason and leverage search engines with reinforcement learning},
  author={Jin, Bowen and Zeng, Hansi and Yue, Zhenrui and Yoon, Jinsung and Arik, Sercan and Wang, Dong and Zamani, Hamed and Han, Jiawei},
  journal={arXiv preprint arXiv:2503.09516},
  year={2025}
}

@article{lai2025mini,
  title={Mini-o3: Scaling Up Reasoning Patterns and Interaction Turns for Visual Search},
  author={Lai, Xin and Li, Junyi and Li, Wei and Liu, Tao and Li, Tianjian and Zhao, Hengshuang},
  journal={arXiv preprint arXiv:2509.07969},
  year={2025}
}

@InProceedings{Wu_2024_CVPR,
    author    = {Wu, Penghao and Xie, Saining},
    title     = {V*: Guided Visual Search as a Core Mechanism in Multimodal LLMs},
    booktitle = {Proceedings of the IEEE/CVF Conference on Computer Vision and Pattern Recognition (CVPR)},
    month     = {June},
    year      = {2024},
    pages     = {13084-13094}
}

@article{hrbench,
      title={Divide, Conquer and Combine: A Training-Free Framework for High-Resolution Image Perception in Multimodal Large Language Models}, 
      author={Wenbin Wang and Liang Ding and Minyan Zeng and Xiabin Zhou and Li Shen and Yong Luo and Dacheng Tao},
      year={2024},
      journal={arXiv preprint},
      url={https://arxiv.org/abs/2408.15556}, 
}

@inproceedings{yue2024mmmu,
  title={Mmmu: A massive multi-discipline multimodal understanding and reasoning benchmark for expert agi},
  author={Yue, Xiang and Ni, Yuansheng and Zhang, Kai and Zheng, Tianyu and Liu, Ruoqi and Zhang, Ge and others},
  booktitle={Proceedings of the IEEE/CVF Conference on Computer Vision and Pattern Recognition},
  pages={9556--9567},
  year={2024}
}

@article{lu2023mathvista,
  title={Mathvista: Evaluating mathematical reasoning of foundation models in visual contexts},
  author={Lu, Pan and Bansal, Hritik and Xia, Tony and Liu, Jiacheng and Li, Chunyuan and Hajishirzi, Hannaneh and Cheng, Hao and Chang, Kai-Wei and Galley, Michel and Gao, Jianfeng},
  journal={arXiv preprint arXiv:2310.02255},
  year={2023}
}

@article{yue2024mmmupro,
  title={Mmmu-pro: A more robust multi-discipline multimodal understanding benchmark},
  author={Yue, Xiang and Zheng, Tianyu and Ni, Yuansheng and Wang, Yubo and Zhang, Kai and Tong, Shengbang and Sun, Yuxuan and Yu, Botao and Zhang, Ge and Sun, Huan and others},
  journal={arXiv preprint arXiv:2409.02813},
  year={2024}
}

@article{wang2024measuring,
  title={Measuring multimodal mathematical reasoning with math-vision dataset},
  author={Wang, Ke and Pan, Junting and Shi, Weikang and Lu, Zimu and Ren, Houxing and Zhou, Aojun and Zhan, Mingjie and Li, Hongsheng},
  journal={Advances in Neural Information Processing Systems},
  volume={37},
  pages={95095--95169},
  year={2024}
}

@misc{rahmanzadehgervi2025visionlanguagemodelsblind,
      title={Vision language models are blind: Failing to translate detailed visual features into words}, 
      author={Pooyan Rahmanzadehgervi and Logan Bolton and Mohammad Reza Taesiri and Anh Totti Nguyen},
      year={2025},
      eprint={2407.06581},
      archivePrefix={arXiv},
      primaryClass={cs.AI},
      url={https://arxiv.org/abs/2407.06581}, 
}

@misc{roberts2025zerobenchimpossiblevisualbenchmark,
      title={ZeroBench: An Impossible Visual Benchmark for Contemporary Large Multimodal Models}, 
      author={Jonathan Roberts and Mohammad Reza Taesiri and Ansh Sharma and Akash Gupta and Samuel Roberts and Ioana Croitoru and Simion-Vlad Bogolin and Jialu Tang and Florian Langer and others},
      year={2025},
      eprint={2502.09696},
      archivePrefix={arXiv},
      primaryClass={cs.CV},
      url={https://arxiv.org/abs/2502.09696}, 
}

@misc{xu2025visulogicbenchmarkevaluatingvisual,
      title={VisuLogic: A Benchmark for Evaluating Visual Reasoning in Multi-modal Large Language Models}, 
      author={Weiye Xu and Jiahao Wang and Weiyun Wang and Zhe Chen and Wengang Zhou and Aijun Yang and Lewei Lu and Houqiang Li and Xiaohua Wang and Xizhou Zhu and others},
      year={2025},
      eprint={2504.15279},
      archivePrefix={arXiv},
      primaryClass={cs.CV},
      url={https://arxiv.org/abs/2504.15279}, 
}

@article{zou2024dynamath,
  title={Dynamath: A dynamic visual benchmark for evaluating mathematical reasoning robustness of vision language models},
  author={Zou, Chengke and Guo, Xingang and Yang, Rui and Zhang, Junyu and Hu, Bin and Zhang, Huan},
  journal={arXiv preprint arXiv:2411.00836},
  year={2024}
}

@article{xiao2024logicvista,
  title={Logicvista: Multimodal llm logical reasoning benchmark in visual contexts},
  author={Xiao, Yijia and Sun, Edward and Liu, Tianyu and Wang, Wei},
  journal={arXiv preprint arXiv:2407.04973},
  year={2024}
}

@article{qiao2024we,
  title={We-math: Does your large multimodal model achieve human-like mathematical reasoning?},
  author={Qiao, Runqi and Tan, Qiuna and Dong, Guanting and Wu, Minhui and Sun, Chong and Song, Xiaoshuai and GongQue, Zhuoma and Lei, Shanglin and Wei, Zhe and Zhang, Miaoxuan and others},
  journal={arXiv preprint arXiv:2407.01284},
  year={2024}
}

@article{song2025visualpuzzles,
  title         = {VisualPuzzles: Decoupling Multimodal Reasoning Evaluation from Domain Knowledge},
  author        = {Song, Yueqi and Ou, Tianyue and Kong, Yibo and Li, Zecheng and Neubig, Graham and Yue, Xiang},
  year          = {2025},
  journal       = {arXiv preprint arXiv:2504.10342},
  url           = {https://arxiv.org/abs/2504.10342}
}

@inproceedings{zhang2024mathverse,
  title={Mathverse: Does your multi-modal llm truly see the diagrams in visual math problems?},
  author={Zhang, Renrui and Jiang, Dongzhi and Zhang, Yichi and Lin, Haokun and Guo, Ziyu and Qiu, Pengshuo and Zhou, Aojun and Lu, Pan and Chang, Kai-Wei and Qiao, Yu and others},
  booktitle={European Conference on Computer Vision},
  pages={169--186},
  year={2024},
  organization={Springer}
}

@article{ye2025mobile,
  title={Mobile-agent-v3: Fundamental agents for gui automation},
  author={Ye, Jiabo and Zhang, Xi and Xu, Haiyang and Liu, Haowei and Wang, Junyang and Zhu, Zhaoqing and Zheng, Ziwei and others},
  journal={arXiv preprint arXiv:2508.15144},
  year={2025}
}

@article{wang2025opencua,
  title={Opencua: Open foundations for computer-use agents},
  author={Wang, Xinyuan and Wang, Bowen and Lu, Dunjie and Yang, Junlin and Xie, Tianbao and Wang, Junli and Deng, Jiaqi and Guo, Xiaole and Xu, Yiheng and Wu, Chen Henry and others},
  journal={arXiv preprint arXiv:2508.09123},
  year={2025}
}

@inproceedings{cheng2025simplevqa,
  title={Simplevqa: Multimodal factuality evaluation for multimodal large language models},
  author={Cheng, Xianfu and Zhang, Wei and Zhang, Shiwei and Yang, Jian and Guan, Xiangyuan and Wu, Xianjie and Li, Xiang and Zhang, Ge and Liu, Jiaheng and Mai, Yuying and others},
  booktitle={Proceedings of the IEEE/CVF International Conference on Computer Vision},
  pages={4637--4646},
  year={2025}
}

@inproceedings{li2025unisvg,
  title={Unisvg: A unified dataset for vector graphic understanding and generation with multimodal large language models},
  author={Li, Jinke and Yu, Jiarui and Wei, Chenxing and Dong, Hande and Lin, Qiang and Yang, Liangjing and Wang, Zhicai and Hao, Yanbin},
  booktitle={Proceedings of the 33rd ACM International Conference on Multimedia},
  pages={13156--13163},
  year={2025}
}

@inproceedings{si2025design2code,
  title={Design2Code: Benchmarking Multimodal Code Generation for Automated Front-End Engineering},
  author={Si, Chenglei and Zhang, Yanzhe and Li, Ryan and Yang, Zhengyuan and Liu, Ruibo and Yang, Diyi},
  booktitle={Proceedings of the 2025 Conference of the Nations of the Americas Chapter of the Association for Computational Linguistics: Human Language Technologies (Volume 1: Long Papers)},
  pages={3956--3974},
  year={2025}
}

@article{yang2024chartmimic,
  title={Chartmimic: Evaluating lmm's cross-modal reasoning capability via chart-to-code generation},
  author={Yang, Cheng and Shi, Chufan and Liu, Yaxin and Shui, Bo and Wang, Junjie and Jing, Mohan and Xu, Linran and Zhu, Xinyu and Li, Siheng and Zhang, Yuxiang and others},
  journal={arXiv preprint arXiv:2406.09961},
  year={2024}
}

@misc{likaixin2025iconstack,
  author       = {Li, Kaixin and others},
  title        = {IconStack},
  year         = {2025},
  publisher    = {Hugging Face},
  url = {https://huggingface.co/datasets/likaixin/IconStack-48M-Rendered-Train}
}

@inproceedings{li2024mmcode,
  title={MMCode: Benchmarking Multimodal Large Language Models for Code Generation with Visually Rich Programming Problems},
  author={Li, Kaixin and Tian, Yuchen and Hu, Qisheng and Luo, Ziyang and Huang, Zhiyong and Ma, Jing},
  booktitle={Findings of the Association for Computational Linguistics: EMNLP 2024},
  pages={736--783},
  year={2024}
}

@article{DBLP:journals/corr/abs-2406-04127,
  author       = {Aryo Pradipta Gema and
                  Joshua Ong Jun Leang and
                  Giwon Hong and
                  Alessio Devoto and
                  Alberto Carlo Maria Mancino and
                  Rohit Saxena and
                  Xuanli He and
                  Yu Zhao and
                  Xiaotang Du and
                  Mohammad Reza Ghasemi Madani and
                  Claire Barale and
                  Robert McHardy and
                  Joshua Harris and
                  Jean Kaddour and
                  Emile van Krieken and
                  Pasquale Minervini},
  title        = {Are We Done with MMLU?},
  journal      = {CoRR},
  volume       = {abs/2406.04127},
  year         = {2024},
  url          = {https://doi.org/10.48550/arXiv.2406.04127},
  doi          = {10.48550/ARXIV.2406.04127},
  eprinttype    = {arXiv},
  eprint       = {2406.04127},
  bibsource    = {dblp computer science bibliography, https://dblp.org}
}

@article{DBLP:journals/corr/abs-2311-12022,
  author       = {David Rein and
                  Betty Li Hou and
                  Asa Cooper Stickland and
                  Jackson Petty and
                  Richard Yuanzhe Pang and
                  Julien Dirani and
                  Julian Michael and
                  Samuel R. Bowman},
  title        = {{GPQA:} {A} Graduate-Level Google-Proof Q{\&}A Benchmark},
  journal      = {CoRR},
  volume       = {abs/2311.12022},
  year         = {2023},
  url          = {https://doi.org/10.48550/arXiv.2311.12022},
  doi          = {10.48550/ARXIV.2311.12022},
  eprinttype    = {arXiv},
  eprint       = {2311.12022},
  bibsource    = {dblp computer science bibliography, https://dblp.org}
}

@article{DBLP:journals/corr/abs-2502-14739,
  author       = {M{-}A{-}P Team},
  title        = {SuperGPQA: Scaling {LLM} Evaluation across 285 Graduate Disciplines},
  journal      = {CoRR},
  volume       = {abs/2502.14739},
  year         = {2025},
  url          = {https://doi.org/10.48550/arXiv.2502.14739},
  doi          = {10.48550/ARXIV.2502.14739},
  eprinttype    = {arXiv},
  eprint       = {2502.14739},
  bibsource    = {dblp computer science bibliography, https://dblp.org}
}

@article{DBLP:journals/corr/abs-2406-19314,
  author       = {Colin White and
                  Samuel Dooley and
                  Manley Roberts and
                  Arka Pal and
                  Benjamin Feuer and
                  Siddhartha Jain and
                  Ravid Shwartz{-}Ziv and
                  Neel Jain and
                  others},
  title        = {LiveBench: {A} Challenging, Contamination-Free {LLM} Benchmark},
  journal      = {CoRR},
  volume       = {abs/2406.19314},
  year         = {2024},
  url          = {https://doi.org/10.48550/arXiv.2406.19314},
  doi          = {10.48550/ARXIV.2406.19314},
  eprinttype    = {arXiv},
  eprint       = {2406.19314},
  bibsource    = {dblp computer science bibliography, https://dblp.org}
}

@article{DBLP:journals/corr/abs-2403-07974,
  author       = {Naman Jain and
                  King Han and
                  Alex Gu and
                  Wen{-}Ding Li and
                  Fanjia Yan and
                  Tianjun Zhang and
                  Sida Wang and
                  Armando Solar{-}Lezama and
                  Koushik Sen and
                  Ion Stoica},
  title        = {LiveCodeBench: Holistic and Contamination Free Evaluation of Large
                  Language Models for Code},
  journal      = {CoRR},
  volume       = {abs/2403.07974},
  year         = {2024},
  url          = {https://doi.org/10.48550/arXiv.2403.07974},
  doi          = {10.48550/ARXIV.2403.07974},
  eprinttype    = {arXiv},
  eprint       = {2403.07974},
  bibsource    = {dblp computer science bibliography, https://dblp.org}
}

@article{DBLP:journals/corr/abs-2506-16395,
  author       = {Zhexu Wang and
                  Yiping Liu and
                  Yejie Wang and
                  Wenyang He and
                  Bofei Gao and
                  Muxi Diao and
                  Yanxu Chen and
                  Kelin Fu and
                  Flood Sung and
                  Zhilin Yang and
                  Tianyu Liu and
                  Weiran Xu},
  title        = {OJBench: {A} Competition Level Code Benchmark For Large Language Models},
  journal      = {CoRR},
  volume       = {abs/2506.16395},
  year         = {2025},
  url          = {https://doi.org/10.48550/arXiv.2506.16395},
  doi          = {10.48550/ARXIV.2506.16395},
  eprinttype    = {arXiv},
  eprint       = {2506.16395},
  bibsource    = {dblp computer science bibliography, https://dblp.org}
}

@article{DBLP:journals/corr/abs-2311-07911,
  author       = {Jeffrey Zhou and
                  Tianjian Lu and
                  Swaroop Mishra and
                  Siddhartha Brahma and
                  Sujoy Basu and
                  Yi Luan and
                  Denny Zhou and
                  Le Hou},
  title        = {Instruction-Following Evaluation for Large Language Models},
  journal      = {CoRR},
  volume       = {abs/2311.07911},
  year         = {2023},
  url          = {https://doi.org/10.48550/arXiv.2311.07911},
  doi          = {10.48550/ARXIV.2311.07911},
  eprinttype    = {arXiv},
  eprint       = {2311.07911},
  bibsource    = {dblp computer science bibliography, https://dblp.org}
}

@article{DBLP:journals/corr/abs-2406-11939,
  author       = {Tianle Li and
                  Wei{-}Lin Chiang and
                  Evan Frick and
                  Lisa Dunlap and
                  Tianhao Wu and
                  Banghua Zhu and
                  Joseph E. Gonzalez and
                  Ion Stoica},
  title        = {From Crowdsourced Data to High-Quality Benchmarks: Arena-Hard and
                  BenchBuilder Pipeline},
  journal      = {CoRR},
  volume       = {abs/2406.11939},
  year         = {2024},
  url          = {https://doi.org/10.48550/arXiv.2406.11939},
  doi          = {10.48550/ARXIV.2406.11939},
  eprinttype    = {arXiv},
  eprint       = {2406.11939},
  bibsource    = {dblp computer science bibliography, https://dblp.org}
}

@article{DBLP:journals/corr/abs-2503-05244,
  author       = {Yuning Wu and
                  Jiahao Mei and
                  Ming Yan and
                  Chenliang Li and
                  Shaopeng Lai and
                  Yuran Ren and
                  Zijia Wang and
                  Ji Zhang and
                  Mengyue Wu and
                  Qin Jin and
                  Fei Huang},
  title        = {WritingBench: {A} Comprehensive Benchmark for Generative Writing},
  journal      = {CoRR},
  volume       = {abs/2503.05244},
  year         = {2025},
  url          = {https://doi.org/10.48550/arXiv.2503.05244},
  doi          = {10.48550/ARXIV.2503.05244},
  eprinttype    = {arXiv},
  eprint       = {2503.05244},
  bibsource    = {dblp computer science bibliography, https://dblp.org}
}

@article{DBLP:journals/corr/abs-2410-15553,
  author       = {Yun He and
                  Di Jin and
                  Chaoqi Wang and
                  Chloe Bi and
                  Karishma Mandyam and
                  Hejia Zhang and
                  Chen Zhu and
                  Ning Li and
                  Tengyu Xu and
                  Hongjiang Lv and
                  Shruti Bhosale and
                  Chenguang Zhu and
                  Karthik Abinav Sankararaman and
                  Eryk Helenowski and
                  Melanie Kambadur and
                  Aditya Tayade and
                  Hao Ma and
                  Han Fang and
                  Sinong Wang},
  title        = {Multi-IF: Benchmarking LLMs on Multi-Turn and Multilingual Instructions
                  Following},
  journal      = {CoRR},
  volume       = {abs/2410.15553},
  year         = {2024},
  url          = {https://doi.org/10.48550/arXiv.2410.15553},
  doi          = {10.48550/ARXIV.2410.15553},
  eprinttype    = {arXiv},
  eprint       = {2410.15553},
  bibsource    = {dblp computer science bibliography, https://dblp.org}
}

@inproceedings{DBLP:conf/iclr/RomanouFSNSMACH25,
  author       = {Angelika Romanou and
                  Negar Foroutan and
                  Anna Sotnikova and
                  Zeming Chen and
                  Sree Harsha Nelaturu and
                  Shivalika Singh and
                  Rishabh Maheshwary and
                  Micol Altomare and
                  Mohamed A. Haggag and
                  Imanol Schlag and
                  others},
  title        = {{INCLUDE:} Evaluating Multilingual Language Understanding with Regional
                  Knowledge},
  booktitle    = {The Thirteenth International Conference on Learning Representations,
                  {ICLR} 2025, Singapore, April 24-28, 2025},
  publisher    = {OpenReview.net},
  year         = {2025},
  bibsource    = {dblp computer science bibliography, https://dblp.org}
}

@article{DBLP:journals/corr/abs-2504-18428,
  author       = {Yiming Wang and
                  Pei Zhang and
                  Jialong Tang and
                  Haoran Wei and
                  Baosong Yang and
                  Rui Wang and
                  Chenshu Sun and
                  Feitong Sun and
                  Jiran Zhang and
                  Junxuan Wu and
                  Qiqian Cang and
                  Yichang Zhang and
                  Fei Huang and
                  Junyang Lin and
                  others},
  title        = {PolyMath: Evaluating Mathematical Reasoning in Multilingual Contexts},
  journal      = {CoRR},
  volume       = {abs/2504.18428},
  year         = {2025},
  url          = {https://doi.org/10.48550/arXiv.2504.18428},
  doi          = {10.48550/ARXIV.2504.18428},
  eprinttype    = {arXiv},
  eprint       = {2504.18428},
  bibsource    = {dblp computer science bibliography, https://dblp.org}
}

@inproceedings{patil2025bfcl,
  title={The Berkeley Function Calling Leaderboard (BFCL): From Tool Use to Agentic Evaluation of Large Language Models},
  author={Patil, Shishir G. and Mao, Huanzhi and Cheng-Jie Ji, Charlie and Yan, Fanjia and Suresh, Vishnu and Stoica, Ion and E. Gonzalez, Joseph},
  year={2024},
  booktitle = {Advances in Neural Information Processing Systems},
}

@article{DBLP:journals/corr/abs-2312-06281,
  author       = {Samuel J. Paech},
  title        = {EQ-Bench: An Emotional Intelligence Benchmark for Large Language Models},
  journal      = {CoRR},
  volume       = {abs/2312.06281},
  year         = {2023},
  url          = {https://doi.org/10.48550/arXiv.2312.06281},
  doi          = {10.48550/ARXIV.2312.06281},
  eprinttype    = {arXiv},
  eprint       = {2312.06281},
  bibsource    = {dblp computer science bibliography, https://dblp.org}
}

@misc{hmmt25,
  author = {{HMMT}},
  title = { Hmmt 2025},
  year = {2025},
  howpublished = {\url{ https://www.hmmt.org}},
}

@misc{aime25,
  author = {{AIME}},
  title = {AIME problems and solutions},
  year = {2025},
  url = {https://artofproblemsolving.com/wiki/index.php/AIMEProblemsandSo
lutions},
}

@article{chen2024timemarker,
  title={Timemarker: A versatile video-llm for long and short video understanding with superior temporal localization ability},
  author={Chen, Shimin and Lan, Xiaohan and Yuan, Yitian and Jie, Zequn and Ma, Lin},
  journal={arXiv preprint arXiv:2411.18211},
  year={2024}
}

@article{team2025gemini,
  title={Gemini robotics: Bringing ai into the physical world},
  author={Team, Gemini Robotics and Abeyruwan, Saminda and Ainslie, Joshua and Alayrac, Jean-Baptiste and Arenas, Montserrat Gonzalez and Armstrong, Travis and Balakrishna, Ashwin and Baruch, Robert and Bauza, Maria and Blokzijl, Michiel and others},
  journal={arXiv preprint arXiv:2503.20020},
  year={2025}
}

@inproceedings{yang2025thinking,
  title={Thinking in space: How multimodal large language models see, remember, and recall spaces},
  author={Yang, Jihan and Yang, Shusheng and Gupta, Anjali W and Han, Rilyn and Fei-Fei, Li and Xie, Saining},
  booktitle={Proceedings of the Computer Vision and Pattern Recognition Conference},
  pages={10632--10643},
  year={2025}
}

@article{du2024embspatial,
  title={Embspatial-bench: Benchmarking spatial understanding for embodied tasks with large vision-language models},
  author={Du, Mengfei and Wu, Binhao and Li, Zejun and Huang, Xuanjing and Wei, Zhongyu},
  journal={arXiv preprint arXiv:2406.05756},
  year={2024}
}

@article{zhou2025roborefer,
  title={RoboRefer: Towards Spatial Referring with Reasoning in Vision-Language Models for Robotics},
  author={Zhou, Enshen and An, Jingkun and Chi, Cheng and Han, Yi and Rong, Shanyu and Zhang, Chi and Wang, Pengwei and Wang, Zhongyuan and Huang, Tiejun and Sheng, Lu and others},
  journal={arXiv preprint arXiv:2506.04308},
  year={2025}
}

@inproceedings{song2025robospatial,
  title={Robospatial: Teaching spatial understanding to 2d and 3d vision-language models for robotics},
  author={Song, Chan Hee and Blukis, Valts and Tremblay, Jonathan and Tyree, Stephen and Su, Yu and Birchfield, Stan},
  booktitle={Proceedings of the Computer Vision and Pattern Recognition Conference},
  pages={15768--15780},
  year={2025}
}

@article{duan2025codeplot,
  title={CodePlot-CoT: Mathematical Visual Reasoning by Thinking with Code-Driven Images},
  author={Duan, Chengqi and Sun, Kaiyue and Fang, Rongyao and Zhang, Manyuan and Feng, Yan and Luo, Ying and Liu, Yufang and Wang, Ke and Pei, Peng and Cai, Xunliang and others},
  journal={arXiv preprint arXiv:2510.11718},
  year={2025}
}

@misc{lu2025videoagenttrekcomputerusepretraining,
      title={VideoAgentTrek: Computer Use Pretraining from Unlabeled Videos}, 
      author={Dunjie Lu and Yiheng Xu and Junli Wang and Haoyuan Wu and Xinyuan Wang and Zekun Wang and Junlin Yang and Hongjin Su and Jixuan Chen and Junda Chen and Yuchen Mao and Jingren Zhou and Junyang Lin and Binyuan Hui and Tao Yu},
      year={2025},
      eprint={2510.19488},
      archivePrefix={arXiv},
      primaryClass={cs.CL},
      url={https://arxiv.org/abs/2510.19488}, 
}

@misc{xie2025scalingcomputerusegroundinguser,
      title={Scaling Computer-Use Grounding via User Interface Decomposition and Synthesis}, 
      author={Tianbao Xie and Jiaqi Deng and Xiaochuan Li and Junlin Yang and Haoyuan Wu and Jixuan Chen and Wenjing Hu and Xinyuan Wang and Yuhui Xu and Zekun Wang and Yiheng Xu and Junli Wang and Doyen Sahoo and Tao Yu and Caiming Xiong},
      year={2025},
      eprint={2505.13227},
      archivePrefix={arXiv},
      primaryClass={cs.AI},
      url={https://arxiv.org/abs/2505.13227}, 
}

@article{osworld_verified,
  title = {Introducing OSWorld-Verified},
  author = {Tianbao Xie and Mengqi Yuan and Danyang Zhang and Xinzhuang Xiong and Zhennan Shen and Zilong Zhou and Xinyuan Wang and Yanxu Chen and Jiaqi Deng and Junda Chen and Bowen Wang and Haoyuan Wu and Jixuan Chen and Junli Wang and Dunjie Lu and Hao Hu and Tao Yu},
  journal = {xlang.ai},
  year = {2025},
  month = {July},
  url = "https://xlang.ai/blog/osworld-verified"
}

@article{diao2025climb,
  title={Climb: Clustering-based iterative data mixture bootstrapping for language model pre-training},
  author={Diao, Shizhe and Yang, Yu and Fu, Yonggan and Dong, Xin and Su, Dan and Kliegl, Markus and Chen, Zijia and Belcak, Peter and Suhara, Yoshi and Yin, Hongxu and others},
  journal={arXiv preprint arXiv:2504.13161},
  year={2025}
}

@article{wettig2025organize,
  title={Organize the Web: Constructing Domains Enhances Pre-Training Data Curation},
  author={Wettig, Alexander and Lo, Kyle and Min, Sewon and Hajishirzi, Hannaneh and Chen, Danqi and Soldaini, Luca},
  journal={arXiv preprint arXiv:2502.10341},
  year={2025}
}

@article{zhu2023multimodal,
  title={Multimodal c4: An open, billion-scale corpus of images interleaved with text},
  author={Zhu, Wanrong and Hessel, Jack and Awadalla, Anas and Gadre, Samir Yitzhak and Dodge, Jesse and Fang, Alex and Yu, Youngjae and Schmidt, Ludwig and Wang, William Yang and Choi, Yejin},
  journal={Advances in Neural Information Processing Systems},
  volume={36},
  pages={8958--8974},
  year={2023}
}

@article{laurenccon2023obelics,
  title={Obelics: An open web-scale filtered dataset of interleaved image-text documents},
  author={Lauren{\c{c}}on, Hugo and Saulnier, Lucile and Tronchon, L{\'e}o and Bekman, Stas and Singh, Amanpreet and Lozhkov, Anton and Wang, Thomas and Karamcheti, Siddharth and Rush, Alexander and Kiela, Douwe and others},
  journal={Advances in Neural Information Processing Systems},
  volume={36},
  pages={71683--71702},
  year={2023}
}

@article{johnson2019billion,
  title={Billion-scale similarity search with {GPUs}},
  author={Johnson, Jeff and Douze, Matthijs and J{\'e}gou, Herv{\'e}},
  journal={IEEE Transactions on Big Data},
  volume={7},
  number={3},
  pages={535--547},
  year={2019},
  publisher={IEEE}
}

@article{douze2024faiss,
  title={The Faiss library},
  author={Matthijs Douze and Alexandr Guzhva and Chengqi Deng and Jeff Johnson and Gergely Szilvasy and Pierre-Emmanuel Mazaré and Maria Lomeli and Lucas Hosseini and Hervé Jégou},
  year={2024},
  eprint={2401.08281},
  archivePrefix={arXiv},
  primaryClass={cs.LG}
}

@inproceedings{shao2019objects365,
  title={Objects365: A large-scale, high-quality dataset for object detection},
  author={Shao, Shuai and Li, Zeming and Zhang, Tianyuan and Peng, Chao and Yu, Gang and Zhang, Xiangyu and Li, Jing and Sun, Jian},
  booktitle={Proceedings of the IEEE/CVF international conference on computer vision},
  pages={8430--8439},
  year={2019}
}

@article{kuznetsova2020open,
  title={The open images dataset v4: Unified image classification, object detection, and visual relationship detection at scale},
  author={Kuznetsova, Alina and Rom, Hassan and Alldrin, Neil and Uijlings, Jasper and Krasin, Ivan and Pont-Tuset, Jordi and Kamali, Shahab and Popov, Stefan and Malloci, Matteo and Kolesnikov, Alexander and others},
  journal={International journal of computer vision},
  pages={1956--1981},
  year={2020},
}

@inproceedings{brazil2023omni3d,
  title={Omni3d: A large benchmark and model for 3d object detection in the wild},
  author={Brazil, Garrick and Kumar, Abhinav and Straub, Julian and Ravi, Nikhila and Johnson, Justin and Gkioxari, Georgia},
  booktitle={Proceedings of the IEEE/CVF conference on computer vision and pattern recognition},
  pages={13154--13164},
  year={2023}
}

@inproceedings{song2015sun,
  title={Sun rgb-d: A rgb-d scene understanding benchmark suite},
  author={Song, Shuran and Lichtenberg, Samuel P and Xiao, Jianxiong},
  booktitle={Proceedings of the IEEE conference on computer vision and pattern recognition},
  pages={567--576},
  year={2015}
}

@inproceedings{roberts2021hypersim,
  title={Hypersim: A photorealistic synthetic dataset for holistic indoor scene understanding},
  author={Roberts, Mike and Ramapuram, Jason and Ranjan, Anurag and Kumar, Atulit and Bautista, Miguel Angel and Paczan, Nathan and Webb, Russ and Susskind, Joshua M},
  booktitle={Proceedings of the IEEE/CVF international conference on computer vision},
  pages={10912--10922},
  year={2021}
}

@article{baruch2021arkitscenes,
  title={Arkitscenes: A diverse real-world dataset for 3d indoor scene understanding using mobile rgb-d data},
  author={Baruch, Gilad and Chen, Zhuoyuan and Dehghan, Afshin and Dimry, Tal and Feigin, Yuri and Fu, Peter and Gebauer, Thomas and Joffe, Brandon and Kurz, Daniel and Schwartz, Arik and others},
  journal={arXiv preprint arXiv:2111.08897},
  year={2021}
}
\bibliographystyle{colm2024_conference}

\appendix
\clearpage
\section{Benchmarks}
\label{sec:benchmarks}
We evaluate \qwenvl on a wide range of public benchmarks across distinct capabilities: multimodal reasoning, general visual question answering, subjective experience \& instruction following, document understanding (including OCR), 2D/3D visual grounding and counting, spatial reasoning,  video understanding, GUI agent, and Text-Centric tasks. Below, we provide a detailed list of all the benchmarks used.
\begin{itemize}[itemindent=0pt, labelsep=4pt, leftmargin=*]
\item \textbf{Multimodal Reasoning:} We evaluate the models on 12 benchmarks spanning a diverse range of domains—from mathematics and STEM to visual reasoning and puzzle-solving tasks: MMMU~\citep{yue2024mmmu}, MMMU-Pro~\citep{yue2024mmmupro}, MathVision~\citep{wang2024measuring}, MathVision-Wild\textsubscript{\scriptsize photo}, MathVista~\citep{lu2023mathvista}, We-Math~\citep{qiao2024we}, MathVerse~\citep{zhang2024mathverse}, DynaMath~\citep{zou2024dynamath}, Math-VR~\citep{duan2025codeplot}, LogicVista~\citep{xiao2024logicvista}, VisualPuzzles~\citep{song2025visualpuzzles},  VLM are Blind~\citep{rahmanzadehgervi2025visionlanguagemodelsblind}, ZeroBench~(Main/Subtasks)~\citep{roberts2025zerobenchimpossiblevisualbenchmark}, and
VisuLogic~\citep{xu2025visulogicbenchmarkevaluatingvisual}.
\item \textbf{General Visual Question Answering:} We evaluate the models on 4 General VQA benchmarks: MMBench-V1.1~\citep{MMBench}, RealWorldQA~\citep{realworldqa2024}, MMStar~\citep{chen2024we}, and SimpleVQA~\cite{cheng2025simplevqa}.
\item \textbf{Subjective Experience and Instruction Following:} We evaluate the model on 3 benchmarks, across subject experience and complex instruction following: HallusionBench~\citep{hallusion_bench}, MM-MT-Bench~\citep{mm_mt_bench}, and MIA-Bench~\citep{mia-bench}.
\item \textbf{Document Understanding:} We perform comprehensive evaluation on OCR and document understanding ability of Qwen3-VL series across a diverse range OCR related benchmarks: DocVQA~\citep{docvqa}, InfoVQA~\citep{Mathew2021InfographicVQA}, AI2D~\citep{Kembhavi2016AI2D}, ChartQA~\citep{masry2022chartqa}, OCRBench~\citep{Liu_2024_OCRBench}, OCRBench\_v2~\citep{fu2024ocrbenchv2improvedbenchmark}, CC-OCR~\citep{yang2024ccocrcomprehensivechallengingocr}, OmniDocBench~\citep{ouyang2024omnidocbenchbenchmarkingdiversepdf}, CharXiv~\citep{wang2024charxiv}, and MMLongBench-Doc~\citep{ma2024mmlongbench}.

\item \textbf{2D/3D Grounding and Spatial Understanding:} We evaluate the models on 11 benchmarks include 2D grounding, 3D grounding and spatial understanding: RefCOCO/+/g~\citep{refcoco, refcocog}, ODinW-13~\citep{li2022grounded}, CountBench~\citep{paiss2023teaching}, ARKitScenes~\citep{baruch2021arkitscenes}, Hypersim~\citep{roberts2021hypersim}, SUN RGB-D~\citep{song2015sun}, ERQA~\citep{team2025gemini}, VSIBench~\citep{yang2025thinking}, EmbSpatial~\citep{du2024embspatial},RefSpatial~\citep{zhou2025roborefer}, and RoboSpatialHome~\citep{song2025robospatial}.
\item \textbf{Video Understanding:} We use seven benchmarks to evaluate the model's video understanding capabilities: VideoMME~\citep{fu2024video}, MVBench~\citep{li2024mvbench}, VideoMMMU~\citep{hu2025video}, MMVU~\citep{zhao2025mmvu}, LVBench~\citep{wang2024lvbench}, MLVU~\citep{zhou2024mlvu}, Charades-STA~\citep{gao2017tall}.

\item \textbf{Coding:} We evaluate the model's multi-modal coding capabilities, particularly in front-end reconstruction and SVG generation, using the Design2Code~\citep{si2025design2code}, ChartMimic~\citep{yang2024chartmimic}, and UniSVG~\citep{li2025unisvg} benchmarks.

\item \textbf{GUI Agent:} We evaluate GUI agent capabilities using benchmarks that test both perception and decision-making. For perception, we use ScreenSpot~\citep{cheng2024seeclick}, ScreenSpot Pro~\citep{screenspotpro}, and OSWorldG~\citep{xie2025scalingcomputerusegroundinguser} to measure GUI grounding and understanding of interface layouts across devices. For decision-making, we use AndroidWorld~\citep{rawles2024androidworld} and OSWorld~\citep{xie2025osworld,osworld_verified} to evaluate interactive control, planning, and execution within real or simulated operating environments.

\item \textbf{Text-Centric Tasks:} We evaluate the models on a wide range of text-centric datasets. (1) \textbf{Knowledge:} MMLU-Pro~\citep{mmlupro}, MMLU-Redux~\citep{DBLP:journals/corr/abs-2406-04127}, GPQA~\citep{DBLP:journals/corr/abs-2311-12022}, SuperGPQA~\citep{DBLP:journals/corr/abs-2502-14739}, (2) \textbf{Reasoning:} AIME-25~\citep{aime25}, HMMT-25~\citep{hmmt25}, LiveBench (2024-11-25)~\citep{DBLP:journals/corr/abs-2406-19314}, (3) \textbf{Code:} LiveCodeBench v6~\citep{DBLP:journals/corr/abs-2403-07974}, CFEval, OJBench~\citep{DBLP:journals/corr/abs-2506-16395}, (4) \textbf{Alignment Tasks:} IFEval~\citep{DBLP:journals/corr/abs-2311-07911}, Arena-Hard v2~\citep{DBLP:journals/corr/abs-2406-11939} , Creative Writing v3~\citep{DBLP:journals/corr/abs-2312-06281}, WritingBench~\citep{DBLP:journals/corr/abs-2503-05244}, (5) \textbf{Agent:} BFCL-v3~\citep{patil2025bfcl}, TAU2-Retail, TAU2-Airline, TAU2-Telecom, (6) \textbf{Multilingual:} MultiIF~\citep{DBLP:journals/corr/abs-2410-15553}, MMLU-ProX, INCLUDE~\citep{DBLP:conf/iclr/RomanouFSNSMACH25}, PolyMATH~\citep{DBLP:journals/corr/abs-2504-18428}.

\end{itemize}

\clearpage

\section{Evaluation Prompts}
\label{sec:eval_prompts}

To ensure reproducibility and facilitate future research, we provide here the complete set of prompts used to evaluate our model across all benchmarks. These prompts were consistently applied during inference to maintain fairness and comparability.

\subsection{STEM \& Puzzle}
\label{sec:stem}

\begin{tcolorbox}[
  title=MMMU,
  fonttitle=\bfseries,
  colframe=black,
  colback=white,
  toptitle=1mm,
  bottomtitle=1mm,
  top=2mm,
  verbatim,
]
<image>

Question: \{question\}

Options:

\{options\}

Please select the correct answer from the options above.
\end{tcolorbox}

\begin{tcolorbox}[
  title=MMMUPro\_Standard,
  fonttitle=\bfseries,
  colframe=black,
  colback=white,
  toptitle=1mm,
  bottomtitle=1mm,
  top=2mm,
  verbatim,
]
<image>

\{question\}

\{options\}

Please select the correct answer from the options.
\end{tcolorbox}

\begin{tcolorbox}[
  title=MMMUPro\_Vision,
  fonttitle=\bfseries,
  colframe=black,
  colback=white,
  toptitle=1mm,
  bottomtitle=1mm,
  top=2mm,
  verbatim,
]
<image>

Identify the problem and solve it. Think step by step before answering.
\end{tcolorbox}

\begin{tcolorbox}[
  title=MathVista,
  title={MathVista | MathVision | MathVerse | LogicVista},
  fonttitle=\bfseries,
  colframe=black,
  colback=white,
  toptitle=1mm,
  bottomtitle=1mm,
  top=2mm,
  verbatim,
]
<image>

\{question\}
\end{tcolorbox}

\begin{tcolorbox}[
  title=We-Math,
  fonttitle=\bfseries,
  colframe=black,
  colback=white,
  toptitle=1mm,
  bottomtitle=1mm,
  top=2mm,
  verbatim,
]
<image>

Now, we require you to solve a multiple-choice math question. Please briefly describe your thought process and provide the final answer(option).

Question: \{question\}

Option: \{options\}

Regarding the format, please answer following the template below, and be sure to include two <> symbols:

<Thought process>: <<your thought process>> <Answer>: <<your option>>
\end{tcolorbox}

\begin{tcolorbox}[
  title=ZeroBench,
  fonttitle=\bfseries,
  colframe=black,
  colback=white,
  toptitle=1mm,
  bottomtitle=1mm,
  top=2mm,
  verbatim,
]
<image>

\{question\}

Let's think step by step and give the final answer in curly braces, like this: \{final answer\}
\end{tcolorbox}

\begin{tcolorbox}[
  title=DynaMath,
  fonttitle=\bfseries,
  colframe=black,
  colback=white,
  toptitle=1mm,
  bottomtitle=1mm,
  top=2mm,
  verbatim,
]
<image>

\#\# Question

\{question\}

\#\# Answer Instruction: Please provide an answer to the question outlined above. Your response should adhere to the following JSON format, which includes two keys: 'solution' and 'short answer'. The 'solution' key can contain detailed steps needed to solve the question, and the 'short answer' key should provide a concise response.

Example of expected JSON response format:

\{

    "solution": "[Detailed step-by-step explanation]",
    
    "short answer": "[Concise Answer]"
    
\}
\end{tcolorbox}

\begin{tcolorbox}[
  title=VLMBlind,
  fonttitle=\bfseries,
  colframe=black,
  colback=white,
  toptitle=1mm,
  bottomtitle=1mm,
  top=2mm,
  verbatim,
]
<image>

Question: \{question\}
\end{tcolorbox}

\begin{tcolorbox}[
  title=VisuLogic,
  fonttitle=\bfseries,
  colframe=black,
  colback=white,
  toptitle=1mm,
  bottomtitle=1mm,
  top=2mm,
  verbatim,
]
<image>

\{question\}

Solve the complex visual logical reasoning problem through step-by-step reasoning. Think about the reasoning process first and answer the question following this format: Answer:{/}/boxed\{\$LETTER\}
\end{tcolorbox}

\begin{tcolorbox}[
  title=VisualPuzzles-Direct,
  fonttitle=\bfseries,
  colframe=black,
  colback=white,
  toptitle=1mm,
  bottomtitle=1mm,
  top=2mm,
  verbatim,
]
<image>

Question: \{question\}

Options:

\{options\}

Answer the question with the option's letter from the given choices directly.
\end{tcolorbox}

\begin{tcolorbox}[
  title=VisualPuzzles-CoT,
  fonttitle=\bfseries,
  colframe=black,
  colback=white,
  toptitle=1mm,
  bottomtitle=1mm,
  top=2mm,
  verbatim,
]
<image>

Question: \{question\}

Options:

\{options\}

Solve the multiple-choice question and then answer with the option letter from the given choices. The last line of your response should be of the following format: 'Answer: \$LETTER' (without quotes), where LETTER is one of the options. Think step by step before answering.
\end{tcolorbox}

\subsection{GeneralVQA}
\label{sec:vqa}

\begin{tcolorbox}[
  title={MMBench | RealWorldQA | MMStar},
  fonttitle=\bfseries,
  colframe=black,
  colback=white,
  toptitle=1mm,
  bottomtitle=1mm,
  top=2mm,
  verbatim,
]
<image>

Question: \{question\}

Options:

\{options\}

Please select the correct answer from the options above.
\end{tcolorbox}

\begin{tcolorbox}[
  title=SimpleVQA,
  fonttitle=\bfseries,
  colframe=black,
  colback=white,
  toptitle=1mm,
  bottomtitle=1mm,
  top=2mm,
  verbatim,
]
<image>

\{question\}
\end{tcolorbox}

\subsection{Alignment}
\label{sec:alignment}

\begin{tcolorbox}[
  title={HallusionBench | MM\_MT\_Bench | MIA-Bench},
  fonttitle=\bfseries,
  colframe=black,
  colback=white,
  toptitle=1mm,
  bottomtitle=1mm,
  top=2mm,
  verbatim,
]
<image>

\{question\}
\end{tcolorbox}

\subsection{Document-Understanding}
\label{sec:doc}

\begin{tcolorbox}[
  title=MMLongBench-Doc,
  fonttitle=\bfseries,
  colframe=black,
  colback=white,
  toptitle=1mm,
  bottomtitle=1mm,
  top=2mm,
  verbatim,
]
<image\_1>

<image\_2>

...

<image\_n>

\{question\}
\end{tcolorbox}

\begin{tcolorbox}[
  title={DocVQA | InfoVQA | ChartQA\_TEST},
  fonttitle=\bfseries,
  colframe=black,
  colback=white,
  toptitle=1mm,
  bottomtitle=1mm,
  top=2mm,
  verbatim,
]
<image>

\{question\}

Answer the question using a single word or phrase.
\end{tcolorbox}

\begin{tcolorbox}[
  title=AI2D,
  fonttitle=\bfseries,
  colframe=black,
  colback=white,
  toptitle=1mm,
  bottomtitle=1mm,
  top=2mm,
  verbatim,
]
<image>

Question: \{question\}

Options:

\{options\}

Please select the correct answer from the options above.
\end{tcolorbox}

\begin{tcolorbox}[
  title={OCRBench | OCRBench\_v2 | CC-OCR | CharXiv},
  fonttitle=\bfseries,
  colframe=black,
  colback=white,
  toptitle=1mm,
  bottomtitle=1mm,
  top=2mm,
  verbatim,
]
<image>

\{question\}
\end{tcolorbox}

\begin{tcolorbox}[
  title=OmniDocBench,
  fonttitle=\bfseries,
  colframe=black,
  colback=white,
  toptitle=1mm,
  bottomtitle=1mm,
  top=2mm,
  verbatim,
]
<image>

You are an AI assistant specialized in converting PDF images to Markdown format. Please follow these instructions for the conversion:

1. Text Processing:
- Accurately recognize all text content in the PDF image without guessing or inferring.
- Convert the recognized text into Markdown format.
- Maintain the original document structure, including headings, paragraphs, lists, etc.

\begin{verbatim}
2. Mathematical Formula Processing:
- Convert all mathematical formulas to LaTeX format.
- Enclose inline formulas with \( \). For example: This is an inline formula \( E = mc^2 \)
- Enclose block formulas with \[ \]. For example: \[ \frac{-b \pm \sqrt{b^2 - 4ac}}{2a} \]
\end{verbatim}

3. Table Processing:
- Convert tables to HTML format.
- Wrap the entire table with <table> and </table>.

4. Figure Handling:
- Ignore figures in the PDF image. Do not attempt to describe or convert images.

5. Output Format:
- Ensure the output Markdown document has a clear structure with appropriate line breaks between elements.
- For complex layouts, try to maintain the original document's structure and format as closely as possible.

Please strictly follow these guidelines to ensure accuracy and consistency in the conversion. Your task is to accurately convert the content of the PDF image into Markdown format without adding any extra explanations or comments.
\end{tcolorbox}

\subsection{2D/3D Grounding}
\label{sec:grounding}

\begin{tcolorbox}[
  title=RefCOCO,
  fonttitle=\bfseries,
  colframe=black,
  colback=white,
  toptitle=1mm,
  bottomtitle=1mm,
  top=2mm,
  verbatim,
]
<image>

Locate every object that matches the description "\{ref\_sentence\}" in the image. Report bbox coordinates in JSON format.
\end{tcolorbox}

\begin{tcolorbox}[
  title=CountBench,
  fonttitle=\bfseries,
  colframe=black,
  colback=white,
  toptitle=1mm,
  bottomtitle=1mm,
  top=2mm,
  verbatim,
]
<image>

Question: \{question\}

Options:

\{options\}

Please select the correct answer from the options above.
\end{tcolorbox}

\begin{tcolorbox}[
  title=ODinW-13,
  fonttitle=\bfseries,
  colframe=black,
  colback=white,
  toptitle=1mm,
  bottomtitle=1mm,
  top=2mm,
  verbatim,
]
<image>

Locate every instance that belongs to the following categories: \'\{obj\_names\}\'. Report bbox coordinates in JSON format.
\end{tcolorbox}

\begin{tcolorbox}[
  title={ARKitScenes | Hypersim  | SUNRGBD},
  fonttitle=\bfseries,
  colframe=black,
  colback=white,
  toptitle=1mm,
  bottomtitle=1mm,
  top=2mm,
  verbatim,
]
<image>

Locate the \{class\_name \} in the provided image and output their positions and dimensions using 3D bounding boxes. The results must be in the JSON format: \texttt{[{"bbox\_3d":[x\_center, y\_center, z\_center, x\_size, y\_size, z\_size, roll, pitch, yaw],"label":"category"}]}. 
\end{tcolorbox}

\subsection{Embodied/Spatial
Understanding}
\label{sec:embodied}

\begin{tcolorbox}[
  title=ERQA,
  fonttitle=\bfseries,
  colframe=black,
  colback=white,
  toptitle=1mm,
  bottomtitle=1mm,
  top=2mm,
  verbatim,
]
<image\_1>

<image\_2>

...

<image\_n>

\{question\}
\end{tcolorbox}

\begin{tcolorbox}[
  enhanced,
  title=VSI-Bench,
  fonttitle=\bfseries,
  colframe=black,
  colback=white,
  toptitle=1mm,
  bottomtitle=1mm,
  top=1mm,
  bottom=1mm,
  left=1mm,
  right=1mm,
  middle=1mm,
  segmentation style={solid, draw=black, line width=0.4pt}
]
\textbf{multiple-choice:}
\begin{Verbatim}
<video>
These are frames of a video.
{question}
Options:
{options}
Answer with the option's letter from the given choices directly.
\end{Verbatim}

\tcblower

\textbf{open-ended:}
\begin{Verbatim}
<video>
These are frames of a video.
{question}
Please answer the question using a single word or phrase.
\end{Verbatim}
\end{tcolorbox}

\begin{tcolorbox}[
  title=EmbSpatialBench,
  fonttitle=\bfseries,
  colframe=black,
  colback=white,
  toptitle=1mm,
  bottomtitle=1mm,
  top=2mm,
  verbatim,
]
<image>

\{question\}
\end{tcolorbox}

\begin{tcolorbox}[
  title=RoboSpatialHome,
  fonttitle=\bfseries,
  colframe=black,
  colback=white,
  toptitle=1mm,
  bottomtitle=1mm,
  top=2mm,
  verbatim,
]
<image>

Locate \{object\_name\} in this image. Output the point coordinates in JSON format.

For example:

[

    \{"point\_2d": [x, y], "label": "point\_1"\}
    
]
\end{tcolorbox}

\begin{tcolorbox}[
  title=RefSpatialBench,
  fonttitle=\bfseries,
  colframe=black,
  colback=white,
  toptitle=1mm,
  bottomtitle=1mm,
  top=2mm,
  verbatim,
]
<image>

\{question\} Output the point coordinates in JSON format.

For example:

[

     \{"point\_2d": [x, y], "label": "point\_1"\}
     
]
\end{tcolorbox}

\subsection{Multi-Image}
\label{sec:multi-img}

\begin{tcolorbox}[
  title=BLINK,
  fonttitle=\bfseries,
  colframe=black,
  colback=white,
  toptitle=1mm,
  bottomtitle=1mm,
  top=2mm,
  verbatim,
]
<image>

Question: \{question\}

Options:

\{options\}

Please select the correct answer from the options above.
\end{tcolorbox}

\begin{tcolorbox}[
  title=MUIRBENCH,
  fonttitle=\bfseries,
  colframe=black,
  colback=white,
  toptitle=1mm,
  bottomtitle=1mm,
  top=2mm,
  verbatim,
]
<image\_1>

<text\_1>

<image\_2>

<text\_2>

...

<image\_n>

<text\_n>

Answer with the option's letter from the given choices directly.
\end{tcolorbox}

\subsection{Video Understanding}
\label{sec:video}

\begin{tcolorbox}[
  title={MVBench | VideoMME | MLVU | LVBench - For instruct models},
  fonttitle=\bfseries,
  colframe=black,
  colback=white,
  toptitle=1mm,
  bottomtitle=1mm,
  top=2mm,
  verbatim,
]
<video>

Select the best answer to the following multiple-choice question based on the video. 

Respond with only the letter (A, B, C, or D) of the correct option.

Question: \{question\} Possible answer choices:

\{options\}

The best answer is:
\end{tcolorbox}

\begin{tcolorbox}[
  title={MVBench | VideoMME | MLVU | LVBench - For thinking models},
  fonttitle=\bfseries,
  colframe=black,
  colback=white,
  toptitle=1mm,
  bottomtitle=1mm,
  top=2mm,
  verbatim,
]
<video>

Select the best answer to the following multiple-choice question based on the video. Respond with only the letter (A, B, C, or D) of the correct option.

Question: \{question\}

\{options\}

Please reason step-by-step, identify relevant visual content, analyze key timestamps and clues, and then provide the final answer.
\end{tcolorbox}

\begin{tcolorbox}[
  title=Charades-STA,
  fonttitle=\bfseries,
  colframe=black,
  colback=white,
  toptitle=1mm,
  bottomtitle=1mm,
  top=2mm,
  verbatim,
]
<video>

Give you a textual query: \{query\_text\}

When does the described content occur in the video? 

Please return the timestamp in seconds.
\end{tcolorbox}

\begin{tcolorbox}[
  enhanced,
  title=VideoMMMU,
  fonttitle=\bfseries,
  colframe=black,
  colback=white,
  toptitle=1mm,
  bottomtitle=1mm,
  top=1mm,
  bottom=1mm,
  left=1mm,
  right=1mm,
  middle=1mm,
  segmentation style={solid, draw=black, line width=0.4pt}
]
\textbf{Perception \& Comprehension:}
\begin{Verbatim}
<video>
{question}
{options}
Please ignore the Quiz question in last frame of the video.
\end{Verbatim}

\tcblower 

\textbf{Adaptation-multiple-choice:}
\begin{Verbatim}
<video>
<image>
You should watch and learn the video content. Then apply what you learned to answer the 
following multi-choice question. The image for this question is at the end of the video.
{question}
{options}
\end{Verbatim}

\tcbline 

\textbf{Adaptation-open-ended:}
\begin{Verbatim}
<video>
<image>
You should watch and learn the video content. Then apply what you learned to answer the 
following open-ended question. The image for this question is at the end of the video.
{question}
\end{Verbatim}
\end{tcolorbox}

\begin{tcolorbox}[
  enhanced,
  title=MMVU,
  fonttitle=\bfseries,
  colframe=black,
  colback=white,
  toptitle=1mm,
  bottomtitle=1mm,
  top=1mm,
  bottom=1mm,
  left=1mm,
  right=1mm,
  middle=1mm,
  segmentation style={solid, draw=black, line width=0.4pt},
]

\textbf{multiple-choice:}

\begin{Verbatim}[breaklines=true, breakanywhere=true, breaksymbolleft={}]
<video>
{question}
{options}
Visual Information: processed video
Answer the given multiple-choice question step by step. Begin by explaining your reasoning process clearly. Conclude by stating the final answer using the following format: "Therefore, the final answer is: $LETTER" (without quotes), where $LETTER is one of the options. Think step by step before answering.
\end{Verbatim}
\tcbline

\textbf{open-ended:}
\begin{Verbatim}[breaklines=true, breakanywhere=true, breaksymbolleft={}]
<video>
{question}
Visual Information: processed video
Answer the given question step by step. Begin by explaining your reasoning process clearly. 
Conclude by stating the final answer using the following format: "Therefore, the final answer is: "Answer: $ANSWER" (without quotes), where $ANSWER is the final answer of the question. Think step by step before answering.
\end{Verbatim}

\end{tcolorbox}

\subsection{Perception with Tool}
\label{sec:perception_w_tool}

\begin{tcolorbox}[
  title=V*,
  fonttitle=\bfseries,
  colframe=black,
  colback=white,
  toptitle=1mm,
  bottomtitle=1mm,
  top=2mm,
  verbatim,
]

Your role is that of a research assistant specializing in visual information. Answer questions about images by looking at them closely and then using research tools. Please follow this structured thinking process and show your work.

\

Start an iterative loop for each question:

\

- **First, look closely:** Begin with a detailed description of the image, paying attention to the user's question. List what you can tell just by looking, and what you'll need to look up.

- **Next, find information:** Use a tool to research the things you need to find out.

- **Then, review the findings:** Carefully analyze what the tool tells you and decide on your next action.

\

Continue this loop until your research is complete.

\

To finish, bring everything together in a clear, synthesized answer that fully responds to the user's question.

\

\#Tools

\ 

You may call one or more functions to assist with the user query.

\ 

You are provided with function signatures within \texttt{\textless tools\textgreater}\texttt{\textless /tools\textgreater} XML tags:
\par\noindent\texttt{\textless tools\textgreater}

\texttt{\{}
\texttt{"type":"function",}
\texttt{"function":\ \{"name":\ "image\_zoom\_in\_tool",\ "description":\ "Zoom in on a specific region of an image by cropping it based on a bounding box (bbox) and an optional object label",\ "arguments":\ \{"type":\ "object",\ "properties":\ \{"bbox\_2d":\ \{"type":\ "array",\ "items":\ \{"type":\ "number"\},\ "minItems":\ 4,\ "maxItems":\ 4,\ "description":\ "The bounding\ box\ of\ the\ region to zoom in, as [x1, y1, x2, y2], where (x1, y1) is the top-left corner and (x2, y2) is the bottom-right corner"\}, "label": \{"type": "string", "description": "The name or label of the object in the specified bounding box"\}, "img\_idx": \{"type": "number", "description": "The index of the zoomed-in image (starting from 0)"\}\}, "required": ["bbox\_2d", "label", "img\_idx"]}\}\}\}

\texttt{\textless /tools\textgreater}

For each function call, return a JSON object with function name and arguments within \texttt{\textless tool\_call\textgreater}\texttt{\textless /tool\_call\textgreater} XML tags:
\par\noindent\texttt{\textless tool\_call\textgreater}\\
\texttt{\{\{"name": \textless function-name\textgreater , "arguments": \textless args-json-object\textgreater \}\}}\\
\texttt{\textless /tool\_call\textgreater}

<image>

\{question\}

\end{tcolorbox}

\begin{tcolorbox}[
  title={HRBench4K | HRBench8K},
  fonttitle=\bfseries,
  colframe=black,
  colback=white,
  toptitle=1mm,
  bottomtitle=1mm,
  top=2mm,
  verbatim,
]

Your role is that of a research assistant specializing in visual information. Answer questions about images by looking at them closely and then using research tools. Please follow this structured thinking process and show your work.

\

Start an iterative loop for each question:

\

- **First, look closely:** Begin with a detailed description of the image, paying attention to the user's question. List what you can tell just by looking, and what you'll need to look up.

- **Next, find information:** Use a tool to research the things you need to find out.

- **Then, review the findings:** Carefully analyze what the tool tells you and decide on your next action.

\

Continue this loop until your research is complete.

\

To finish, bring everything together in a clear, synthesized answer that fully responds to the user's question.

\

\#Tools

\ 

You may call one or more functions to assist with the user query.

\ 

You are provided with function signatures within \texttt{\textless tools\textgreater}\texttt{\textless /tools\textgreater} XML tags:
\par\noindent\texttt{\textless tools\textgreater}

\texttt{\{}
\texttt{"type":"function",}
\texttt{"function":\ \{"name":\ "image\_zoom\_in\_tool",\ "description":\ "Zoom in on a specific region of an image by cropping it based on a bounding box (bbox) and an optional object label",\ "arguments":\ \{"type":\ "object",\ "properties":\ \{"bbox\_2d":\ \{"type":\ "array",\ "items":\ \{"type":\ "number"\},\ "minItems":\ 4,\ "maxItems":\ 4,\ "description":\ "The bounding\ box\ of\ the\ region to zoom in, as [x1, y1, x2, y2], where (x1, y1) is the top-left corner and (x2, y2) is the bottom-right corner"\}, "label": \{"type": "string", "description": "The name or label of the object in the specified bounding box"\}, "img\_idx": \{"type": "number", "description": "The index of the zoomed-in image (starting from 0)"\}\}, "required": ["bbox\_2d", "label", "img\_idx"]}\}\}\}

\texttt{\textless /tools\textgreater}

For each function call, return a JSON object with function name and arguments within \texttt{\textless tool\_call\textgreater}\texttt{\textless /tool\_call\textgreater} XML tags:
\par\noindent\texttt{\textless tool\_call\textgreater}\\
\texttt{\{\{"name": \textless function-name\textgreater , "arguments": \textless args-json-object\textgreater \}\}}\\
\texttt{\textless /tool\_call\textgreater}

<image>

\{question\}

\{options\}

\end{tcolorbox}

\subsection{Coding}

\begin{tcolorbox}[
  title=Design2Code (Generation),
  fonttitle=\bfseries,
  colframe=black,
  colback=white,
  toptitle=1mm,
  bottomtitle=1mm,
  top=2mm,
  verbatim,
]
<image>

You are an expert web developer who specializes in HTML and CSS.
A user will provide you with a screenshot of a webpage.
You need to return a single HTML file that uses HTML and CSS to reproduce the given website.
Include all CSS code in the HTML file itself.
If it involves any images, use "rick.jpg" as the placeholder.
Some images on the webpage are replaced with a blue rectangle as the placeholder, and use "rick.jpg" for those as well.
Do not hallucinate any dependencies on external files. You do not need to include JavaScript scripts for dynamic interactions.
Pay attention to things like size, text, position, and color of all the elements, as well as the overall layout.
Respond with the content of the HTML+CSS file:
\end{tcolorbox}

\begin{tcolorbox}[
  title=Design2Code (GPT-o4-mini Evaluation),
  fonttitle=\bfseries,
  colframe=black,
  colback=white,
  toptitle=1mm,
  bottomtitle=1mm,
  top=2mm,
  verbatim,
]
I will give you two images. The first is the reference, and the second is generated from the first via code rendering. Please rate their similarity from 0-100, where 0 means completely different and 100 means identical. Provide the score inside a LaTeX \boxed{} and briefly explain your reasoning.

<reference\_image>

<generated\_image>
\end{tcolorbox}

\subsection{Agent}
\begin{tcolorbox}[
  title={Screenspot | Screenspot-Pro | OSWorld-G},
  fonttitle=\bfseries,
  colframe=black,
  colback=white,
  toptitle=1mm,
  bottomtitle=1mm,
  top=2mm,
  breakable
]
\textbf{Tools}

You may call one or more functions to assist with the user query.

You are provided with function signatures within \texttt{\textless tools\textgreater} \dots{} \texttt{\textless /tools\textgreater} XML tags:
\par\noindent\texttt{\textless tools\textgreater}
\texttt{\{}
\texttt{"name":"computer\_use",}
\texttt{"description": "Use a mouse to interact with a computer. The screen's resolution is \textless display\_width\_px\textgreater x \textless display\_height\_px\textgreater ."}
\texttt{"notes": "Click with the cursor tip centered on targets; avoid edges unless asked. Do not use other tools (type, key, scroll, left\_click\_drag). Only left\_click and mouse\_move are allowed. If you can't find the element, terminate and report failure.",}
\texttt{"parameters":\{}
\texttt{\ \ "type":"object",}
\texttt{\ \ "required":["action"],}
\texttt{\ \ "properties":\{}
\texttt{\ \ \ \ "action":\{}
\texttt{\ \ \ \ \ \ "type":"string",}
\texttt{\ \ \ \ \ \ "enum":["mouse\_move","left\_click"],}
\texttt{\ \ \ \ \ \ "description":"The action to perform."}
\texttt{\ \ \ \ \},}
\texttt{\ \ \ \ "coordinate":\{}
\texttt{\ \ \ \ \ \ "type":"array",}
\texttt{\ \ \ \ \ \ "description":"(x, y): pixels from left/top. Required for action=mouse\_move and action=left\_click."}
\texttt{\}}
\texttt{\}}
\texttt{\}}
\texttt{\}}

\texttt{\textless /tools\textgreater}

For each function call, return a JSON object with function name and arguments within \texttt{\textless tool\_call\textgreater} \dots{} \texttt{\textless /tool\_call\textgreater} XML tags:
\par\noindent\texttt{\textless tool\_call\textgreater}\\
\texttt{\{\{"name": \textless function-name\textgreater , "arguments": \textless args-json-object\textgreater \}\}}\\
\texttt{\textless /tool\_call\textgreater}

Additionally, if you think the task is infeasible (e.g., the task is not related to the image), return:
\par\noindent\texttt{\textless tool\_call\textgreater}\\
\texttt{\{"name": "computer\_use", "arguments": \{"action": "terminate", "status": "failure"\}\}}\\
\texttt{\textless /tool\_call\textgreater}
\end{tcolorbox}


\end{document}